%% file: main_text.tex
\documentclass{article}

\input{math_commands.tex}

\usepackage{microtype}
\usepackage{graphicx}
\usepackage{booktabs} %

\usepackage[hyphens]{url}
\usepackage{hyperref}
\hypersetup{colorlinks=true,breaklinks=true}

\usepackage[accepted]{icml2024}

\usepackage{amsmath}
\usepackage{amssymb}
\usepackage{mathtools}
\usepackage{amsthm}

\usepackage[capitalize,noabbrev]{cleveref}

\theoremstyle{plain}
\newtheorem{theorem}{Theorem}[section]

\newtheorem{lemma}[theorem]{Lemma}

\theoremstyle{definition}

\theoremstyle{remark}

\usepackage[textsize=tiny]{todonotes}

\usepackage{wrapfig}
\usepackage{floatflt}
\usepackage{subcaption}

\usepackage{algorithm}
\usepackage[noend]{algpseudocode}

\usepackage{commath}

\newcommand{\rulesep}{\unskip\ \vrule\ }

\algrenewcommand\algorithmicindent{1.0em}
\algrenewcommand\algorithmicrequire{\textbf{Input:}}
\algrenewcommand\algorithmicensure{\textbf{Output:}}
\algsetblock[Name]{Init}{6553}{0}{0cm}
\algdefaulttext[Name]{Into}
\newcommand{\Input}{\Require}

\newlength{\twosubht}
\newsavebox{\twosubbox}

\icmltitlerunning{Data-Regularised Environment Design}

\begin{document}

\twocolumn[
\icmltitle{DRED: Zero-Shot Transfer in Reinforcement Learning via \\ Data-Regularised Environment Design}

\icmlsetsymbol{equal}{*}

\begin{icmlauthorlist}
\icmlauthor{Samuel Garcin}{edi}
\icmlauthor{James Doran}{huaw}
\icmlauthor{Shangmin Guo}{edi}
\icmlauthor{Christopher G. Lucas}{edi}
\icmlauthor{Stefano V. Albrecht}{edi}
\end{icmlauthorlist}

\icmlaffiliation{edi}{School of Informatics, University of Edinburgh}
\icmlaffiliation{huaw}{Huawei}

\icmlcorrespondingauthor{Samuel Garcin}{s.garcin@ed.ac.uk}

\icmlkeywords{Reinforcement learning, Zero-shot transfer, Zero-shot generalisation, Environment design, Generative modelling}

\vskip 0.3in
]

\printAffiliationsAndNotice{}  %

\begin{abstract}
Autonomous agents trained using deep reinforcement learning (RL) often lack the ability to successfully generalise to new environments, even when these environments share characteristics with the ones they have encountered during training. In this work, we investigate how the sampling of individual environment instances, or levels, affects the zero-shot generalisation (ZSG) ability of RL agents. We discover that, for deep actor-critic architectures sharing their base layers, prioritising levels according to their value loss minimises the mutual information between the agent's internal representation and the set of training levels in the generated training data. This provides a novel theoretical justification for the regularisation achieved by certain adaptive sampling strategies. We then turn our attention to unsupervised environment design (UED) methods, which assume control over level generation. We find that existing UED methods can significantly shift the training distribution, which translates to low ZSG performance. To prevent both overfitting and distributional shift, we introduce \textit{data-regularised environment design} (DRED). DRED generates levels using a generative model trained to approximate the ground truth distribution of an initial set of level parameters. Through its grounding, DRED achieves significant improvements in ZSG over adaptive level sampling strategies and UED methods.
\end{abstract}

\section{Introduction}\label{sec:intro}
\begin{figure}[bht]
    \centering
    \begin{subfigure}{0.24\linewidth}
        \includegraphics[width=.95\linewidth]{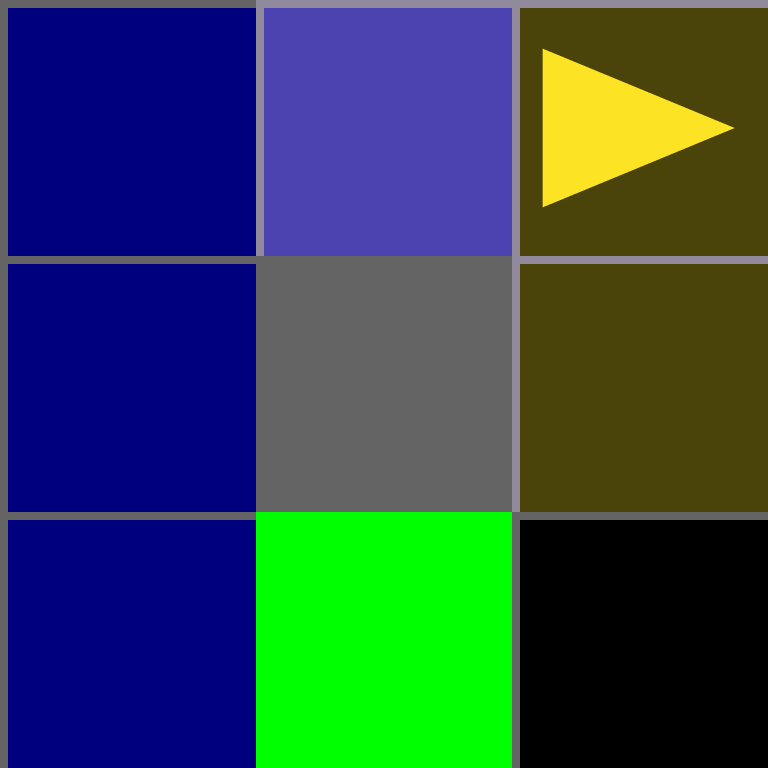}
        \caption{}
        \label{subfig:lvla}
    \end{subfigure}%
    \begin{subfigure}{0.24\linewidth}
        \includegraphics[width=.95\linewidth]{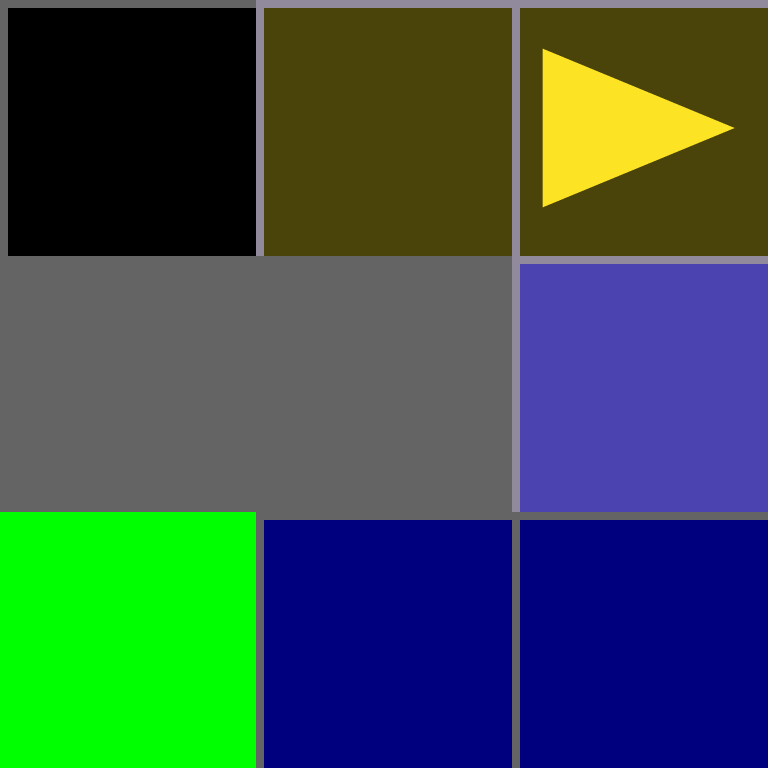}
        \caption{}
        \label{subfig:lvlb}
    \end{subfigure}%
        \begin{subfigure}{0.24\linewidth}
        \includegraphics[width=.95\linewidth]{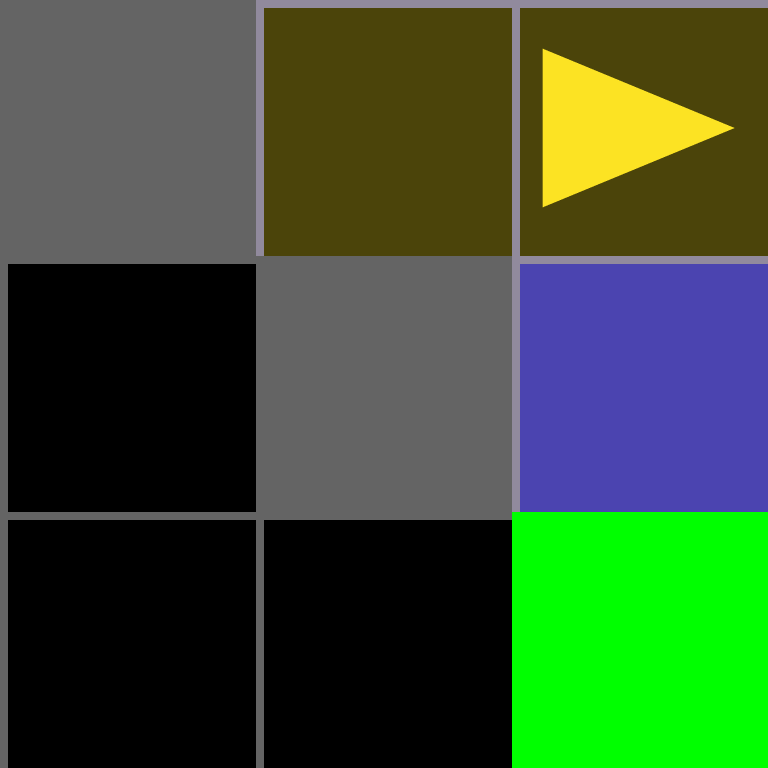}
        \caption{}
        \label{subfig:lvlc}
    \end{subfigure}%
    \hspace{-3pt}
    \rulesep
    \begin{subfigure}{0.24\linewidth}
        \includegraphics[width=0.95\linewidth]{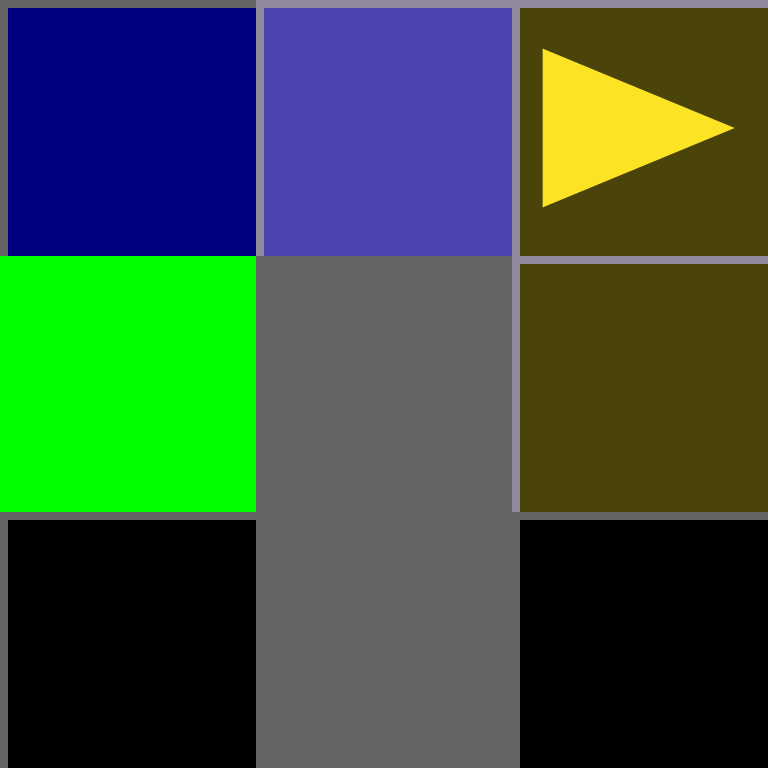}
        \caption{}
        \label{subfig:lvld}
    \end{subfigure}%
    \caption{The agent (yellow) must navigate to the goal (green) but cannot pass through walls (grey) and only observes tiles directly adjacent to itself (highlighted yellow). An agent trained over levels (a)-(c) will transfer zero-shot to level (d) if it has learnt a behavior adapted to the task semantics of following blue tiles to the goal location.
    }
    \label{fig:CMDPex}
\end{figure}
A central challenge facing modern reinforcement learning (RL) is learning policies capable of zero-shot transfer of learned behaviours to a wide range of environment settings. Prior applications of RL algorithms \citep{rubiks,CLrobloc,fastCLrobloc} indicate that strong zero-shot generalisation (ZSG) can be achieved through an adaptive sampling strategy over the set of environment instances available during training, which we refer to as the set of training \textit{levels}. However, the relationship between ZSG and the level sampling process remains poorly understood. In this work, we draw novel connections between this process and the minimisation of an upper bound on the generalisation error. This bound depends on the \textit{mutual information} (MI) between the learned policy $\pi$ and the set of training levels $L$,
\begin{equation} \label{eq:MI}
\noindent
    \mut{\pi}{L} = \entropy(\ri) + \sum_{i\in L} \int_{\bar{\pi} \in \Delta^\sA}\hspace{-1em} P(\pi{=}\bar{\pi}, \ri{=}i) \log P(\ri{=}i |\pi{=}\bar{\pi}),
\end{equation}
where $i$ refers to an individual training level, $\entropy(\ri)$ the entropy of the level distribution and $\Delta^\sA$ the output space of the policy. A model with high $\mut{\pi}{L}$ outputs action distributions that are highly correlated to the level identity. In other words, the model internally predicts the level it is currently in and learns an ensemble of level-specific policies. In general, such an agent will not transfer to new levels zero-shot, as illustrated in the minimal example in \Cref{fig:CMDPex}. In this scenario, it is possible to identify a training level from its initial observation. An agent with high $\mut{\pi}{L}$ would use this information to infer the goal location without having to observe it first. In doing so, it can ignore the task semantics shared across levels, while still maximising its returns on the training set. In fact, it can solve level (a) in fewer steps by ignoring these semantics, as there exists a ``shortcut'' unique to (a). When deployed on (d) after training, the agent will predict it is in (a), since (a) and (d) share the same initial observation. As a result the agent is likely to follow the (a)-specific policy, which will not transfer zero-shot.

As extensively demonstrated in prior work, increasing the number of training levels \citep{OverfittingDRL,procgen,Packer2018AssessingGI} or the amount of data generated in each level (e.g. performing data augmentation on observations \citep{raileanu2021UCB-DrAC,Kostrikov2021ImageAllYouNeed}) reduces the generalisation error, also called \textit{generalisation gap}, defined as the difference in episodic returns when evaluating the agent on the train set and on the full level distribution. From an information theoretic perspective, these approaches induce an implicit information bottleneck on $\mut{\pi}{L}$ by making any particular level more difficult to identify for a model with a fixed representational capacity. However, it is not clear how certain level sampling strategies achieve smaller generalisation gaps (compared to uniform sampling) without increasing the number of training levels nor augmenting the data generated \citep{PLR,Robust-PLR}. In \Cref{sec:procgen}, we show that these strategies may be understood as adaptive rejection sampling mechanisms that prevent data with high $\mut{\pi}{L}$ from being generated. This allows us to build a connection between these strategies and the minimisation of an upper bound on the generalisation gap that depends on $\mut{\pi}{L}$. We supplement our findings with an empirical evaluation of different sampling strategies in the Procgen benchmark \citep{procgen}, and observe a strong correlation between $\mut{\pi}{L}$ and the generalisation gap.

We then introduce \textit{data-regularised environment design} (DRED) in \Cref{sec:ssed}. DRED combines adaptive sampling with data augmentation, but does not perform data augmentation across observations but across \textit{levels}. It does so by learning a generative model of the full distribution of levels we would like the agent to transfer to, trained on a limited starting set of levels from that distribution. DRED then employs an adaptive sampling strategy over an augmented set consisting of the starting levels and the generated levels.

While training on an augmented set is effective in preventing overfitting, it may cause distributional shift if that set is not drawn from the same distribution as the original set. We find that existing environment design methods that rely on an unsupervised generation process cause significant distributional shift, and may lead to poor performance at test time. There is therefore a trade-off between augmenting the training set to prevent instance-overfitting (i.e. to prevent learning level-specific policies), and ensuring that this augmented set is consistent with the original distribution (i.e. to avoid distributional shift). We measure DRED's capabilities in a gridworld navigation task that was designed to highlight this trade-off. We find that DRED achieves a smaller generalisation gap than sampling methods restricted to the starting levels, and does so while maintaining a small distributional shift. This differentiates DRED from unsupervised environment design (UED) methods, which do not account for a target task or target distribution. DRED achieves significant improvements in the agent's ZSG capabilities, reaching 1.2 times the returns of the next best baseline on held-out levels. It also improves performance by two to three times on more difficult instantiations of the target task.\footnote{Our code and experimental data are available at \url{https://github.com/uoe-agents/dred}.} %

\section{Preliminaries}\label{sec:bg}
\textbf{Reinforcement learning.} We model an individual level as a Partially Observable Markov Decision Process (POMDP) $\langle \sA, \sO, \sS, \mathcal{T}, \Omega, R, p_0, \gamma \rangle$ where $\sA$ is the action space, $\sO$ is the observation space, $\sS$ is the set of states, $\mathcal{T} : \sS \times \sA \rightarrow \Delta^\sS$ and $\Omega: \sS \rightarrow \Delta^\sO$ are the transition and observation kernels (denoting $\Delta^\sS$ as the set of all possible probability distributions over $\sS$), $R:\sS\times \sA\rightarrow \sR$ is the reward function, $p_0(\rs)$ is the initial state distribution and $\gamma$ is the discount factor. We consider the episodic RL setting, in which the agent attempts to learn a policy $\pi$ maximising the expected discounted return $V^\pi (s_t) = \mathbb{E}_\pi[\sum^T_{\bar{t}=t}\gamma^{t - \bar{t}} r_t]$ over an episode terminating at timestep $T$, where $s_t$ and $r_t$ are the state and reward at step $t$. We use $V^\pi$ to refer to $V^\pi (s_0)$, the expected episodic returns taken from the first timestep of the episode. In this work, we focus on on-policy actor-critic algorithms \citep{A3C,DDPG,PPO} representing the agent policy $\pi_\vtheta$ and value estimate $\hat{V}_\vtheta$ with neural networks (we use $\vtheta$ to refer to model weights). The policy and value networks usually share an intermediate state representation $b_\vtheta(o_t)$ (or for recurrent architectures $b_\vtheta(H^o_t)$, $H^o_t = \{o_0,\cdots,o_t\}$ being the history of observations $o_i$).

\textbf{Contextual MDPs.} Following \citet{kirk2023survey}, we model the set of environment instances we aim to generalise over as a Contextual-MDP (CMDP) $\mathcal{M} = \langle \sA, \sO, \sS, \mathcal{T}, \Omega, R,  p_0(\rs|\rvx), \gamma, \sX_C, p(\rvx)\rangle$. In a CMDP, the reward function and the transition and observation kernels also depend on the context set $\sX_C$ with associated distribution $p(\rvx)$, that is we have $\mathcal{T} : \sS \times \sX_C \times \sA \rightarrow \Delta^\sS$, $\Omega: \sS \times \sX_C \rightarrow \Delta^\sO$, $R:\sS \times \sX_C \times \sA \rightarrow \sR$. Each element $\vx \in \sX_C$ is not observable by the agent and instantiates a level $\lvl_\vx$ of the CMDP with initial state distribution $p_0(\rs|\rvx)$. The CMDP is equivalent to a POMDP if we consider its state space to be $\sS \times \sX_C$, which means that the agent is always in a partially observable setting, even when the state space of individual levels is fully observable. 

We assume access to a parametrisable simulator with parameter space $\sX$, with $\sX_C \subset \sX$. While prior work expects $\sX_C$ to correspond to all solvable levels in $\sX$ (often defined as the levels in which a positive return is achievable), we will extend our analysis to the more general setting in which there may be more than one CMDP within $\sX$, whereas we aim to solve a specific target CMDP. We refer to levels with parameters $\vx \in \sX_C$ as \textit{in-context} and to levels outside of this set as \textit{out-of-context}.

\textbf{The generalisation gap.} We start training with access to a limited set of level parameters $X_{\text{train}}\subset \sX_C$ sampled from $p(\rvx)$, and evaluate generalisation using a set of held-out level parameters $X_{\text{test}}$, also sampled from $p(\rvx)$. We can estimate generalisation error using a formulation reminiscent of supervised learning,
\begin{equation} \label{eq:GenGap}
    \text{GenGap}(\pi) \coloneqq \frac{1}{|X_{\text{train}}|} \sum_{\vx \in X_{\text{train}}} V^\pi_{\lvl_\vx} - \frac{1}{|X_{\text{test}}|}\sum_{\vx \in X_{\text{test}}} V^\pi_{\lvl_\vx}.
\end{equation}

\citet{instance_invariant} extend generalisation results in the supervised setting \citep{MI_gengap_SL} to obtain an upper bound for the $\text{GenGap}$. 
\begin{theorem}\label{th:gen_gap}
  For any CMDP such that $|V^\pi_C (H_t^o)| \leq D/2,\forall H_t^o, \pi$, with $D$ being a constant, then for any set of training levels $L$, and policy $\pi$
\begin{equation}\label{eq:GenGap_bound}
   \text{GenGap}(\pi) \leq \sqrt{\frac{2D^2}{|L|} \times \mut{L}{\pi}}.
\end{equation}   
\end{theorem}
We will show that minimising this bound is an effective surrogate objective for reducing the $\text{GenGap}$. 

\textbf{Adaptive level sampling.} We study the connection between $\mut{L}{\pi}$ and adaptive sampling strategies over $L$. Prioritised Level Replay \citep[PLR,][]{PLR} introduce a scoring function $\textbf{score}(\tau_i, \pi)$ which compute level scores from trajectory rollouts $\tau_i$. Scores are used to define an adaptive sampling distribution over a level buffer $\Lambda$, with
\begin{equation}\label{eq:plr_buffer_dist}
    P_\Lambda = (1 - \rho) \cdot P_S + \rho \cdot P_R,
\end{equation}
where $P_S$ is a distribution parametrised by the level scores and $\rho$ is a coefficient mixing $P_S$ with a staleness distribution $P_R$ that promotes levels replayed less recently. \citet{PLR} experiment with different scoring functions, and empirically find that the scoring function based on the $\ell_1$-value loss $S^V_i = \textbf{score}(\tau_i, \pi) = (1 / |\tau_i|) \sum_{H_t^o\in\tau_i} |\hat{V} (H_t^o) - V^\pi_i (H_t^o) |$ incurs a significant reduction in the $\text{GenGap}$ at test time. 

In the remaining sections, we draw novel connections between the $\ell_1$-value loss prioritisation strategy and the minimisation of $\mut{L}{\pi}$. We then introduce DRED, a level generation and sampling framework training the agent over an augmented set of levels. DRED jointly minimises $\mut{L}{\pi}$ while increasing $|L|$ and as such is more effective at minimising the bound from \Cref{th:gen_gap}.
\section{How does adaptive level sampling impact generalisation in RL?} \label{sec:procgen}

\begin{figure*}[tb]
            \includegraphics[width=1\linewidth]{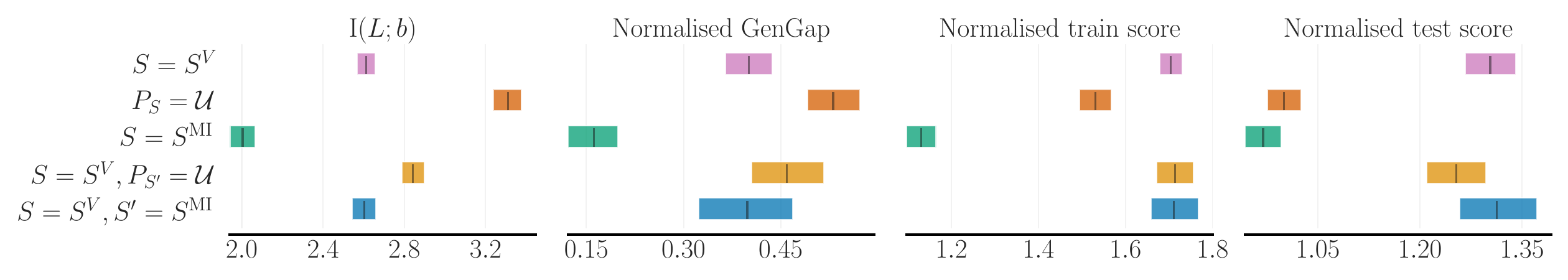}
    \caption{Aggregated $\mut{L}{b}$, $\text{GenGap}$, train and test scores of different sampling strategies over 5 seeds across all Procgen games, using the rliable library \citep{rliable}. Coloured boxes indicate the 95\% confidence interval. For each game, the train and test scores and the $\text{GenGap}$ are normalised by the mean score of $(P_S = \mathcal{U})$ over the test set. Per-game scores, mutual information and classifier accuracy are reported in \Cref{app:results_procgen}, and \Cref{app:procgen} provides an extended description of our experimental setup.}
    \label{fig:procgen_results}
\end{figure*}
In this section, we establish that adaptive sampling strategies reduce the $\text{GenGap}$ by inducing an \textit{information bottleneck} on $\mut{L}{\pi}$. We begin our analysis by deriving a method for empirically estimating an upper bound for $\mut{L}{\pi}$. We find this upper bound and the generalisation gap to be well correlated for different sampling strategies in the Procgen benchmark \citep{procgen}, a benchmark of 16 games designed to measure generalisation in RL. As in prior work \citep{PLR}, we observe that $\ell_1$-value loss prioritisation reduces the $\text{GenGap}$ while improving sample efficiency. Here we take a further step, and aim to understand \textit{why} value loss prioritisation minimises the $\text{GenGap}$.\footnote{For a discussion on the relationship of value loss and sample efficiency, we refer the reader to \citet{Schaul2015PrioritizedER}.} We do so by establishing a connection between the value loss and $\mut{L}{\pi}$. We conclude this section by highlighting the importance of maintaining a low distributional shift during training, an important feature of DRED, which we will introduce in the next section.
\subsection{Mutual information estimation}
$\mut{L}{\pi}$ is often difficult to measure from the model outputs alone. We estimate instead $\mut{L}{b}$, the mutual information at the last shared layer between the actor and critic. As shown in the following lemma, $\mut{L}{b}$ upper bounds $\mut{L}{\pi}$.
\begin{lemma} \label{lem:mi_fixed_L}
    (proof in appendix) Given a set of training levels $L$ and policy $\pi = f \circ b$, where $b(H^o_t) = h_t \in \sB$ is an intermediate representation function and $f : \sB \rightarrow \Delta^\sA$ maps to the agent's action distribution, we have
    \begin{align} \label{eq:lemma_result}
        \mut{L}{\pi} \leq \entropy(\ri) + \sum_{i\in L} \int_{\sB} p(h, i) \log p(i|h)dh,
    \end{align}
    where the right-hand side is equivalent to $\mut{L}{b}$.
\end{lemma}
This result applies to any state representation function $b$, including the non-recurrent case where $b(H^o_t) = b(o_t)$. To remain consistent with the CMDP, we must set $p(\ri)$ to $p(\rvx)$, making the entropy $\entropy(\ri)$ a constant. However, the second term in \Cref{eq:lemma_result} %
depends on the output of the learned representation $b_\vtheta$, and may be empirically estimated as
\begin{equation} \label{eq:mi_empirical_approx}
   \sum_{i\in L} \int_{\sB} p(h, i) \log p(i|h)dh \approx \frac{1}{N} \sum^{N}_{n} \log p_\vtheta(i^{(n)}|\vh^{(n)}),
\end{equation}
where $p_\vtheta$ predicts the current level from representations $\vh_t = b_\vtheta (H_t^o)$ collected during rollouts.

\subsection{Comparing adaptive sampling strategies} \label{sec:procgen_exp}
To evaluate their impact on the $\text{GenGap}$, we compare different adaptive sampling strategies in the ``easy'' setting of Procgen ($|L|=200$, $25M$ timesteps).  We use the default PPO \citep{PPO} architecture and hyperparameters proposed by \citet{procgen}, which uses a non-recurrent intermediate representation $b_\vtheta (o_t)$. We measure $\mut{L}{b}$ using \Cref{eq:mi_empirical_approx}, parametrising $p_\vtheta$ as a linear classifier.
\Cref{fig:procgen_results} compares uniform sampling $(P_S=\mathcal{U})$ with adaptive sampling strategies obtained from different level scoring functions in \Cref{eq:buffer_dist}. We compare value loss scoring $(S=S^V)$ with $(S=S^{\text{MI}})$, a scoring function that directly prioritises levels with low $\mut{L}{b}$. $(S=S^{\text{MI}})$ exploits that \Cref{eq:mi_empirical_approx} may be decomposed into level specific terms $I_i = \sum_{t}^T \log p_\vtheta(i|b_\vtheta(o_t))$ and uses scoring strategy $\textbf{score}(\tau_i, \pi) = I_i$. 

We also consider mixed strategies $(S=S^V, S^\prime=S^{\text{MI}})$ and $(S=S^V, P_{S^\prime}=\mathcal{U})$, which are obtained by introducing a secondary scoring function to \Cref{eq:plr_buffer_dist},
\begin{equation} \label{eq:buffer_dist}
    P_\Lambda = (1 - \rho) \cdot ((1-\eta) \cdot  P_S + \eta \cdot P_{S^\prime}) + \rho \cdot P_R,
\end{equation}
where $\eta$ is the mixing coefficient, which may be scheduled over the course of training. We report our main observations below, with extended analysis and results available in \Cref{app:results_procgen}.
    
    \textbf{First observation.} Minimising $\mut{L}{b}$ on the generated training data using an adaptive distribution results in agent representations with reduced $\mut{L}{b}$ under the \textit{original distribution}.
    
    \textbf{Second observation.} We find this representation to be highly informative of the level identity. Out of 200 training levels, our linear classifier predicts the current level with 49\% accuracy with $(P_S=\mathcal{U})$ and its accuracy only drops to 35\% with $(S=S^V)$ and 19\% with $(S=S^{\text{MI}})$. This indicates that observations get mapped to level-specific \textit{clusters} in the representation space, which let the agent learn level-specific policies.
    
    \textbf{Third observation.} We measure a strong positive correlation $(\rho=0.6)$ and rank correlation (Kendall $\xi=0.5$, $p<1e-50$) between $\mut{L}{b}$ and the $\text{GenGap}$, across all procgen games and sampling strategies tested. This makes $\mut{L}{b}$ a useful proxy for the $\text{GenGap}$. $\mut{L}{b}$ has lower variance, does not require normalisation across environments and, crucially, \textit{does not necessitate a held-out test set of levels} to be measured.
    
    \textbf{Fourth observation.} $(S=S^{\text{MI}})$ achieves the highest reductions in $\mut{L}{b}$ and the $\text{GenGap}$. However, it also significantly degrades performance during training, and is the worst performing strategy when considering final test scores. This result is consistent with \Cref{th:gen_gap} and \Cref{lem:mi_fixed_L}, as $\mut{L}{b}$ bounds the GenGap and not the test returns.\footnote{Excessive data regularisation is not desirable: in the most extreme case, destroying all information contained within the training data would guarantee $\mut{L}{b} = \text{GenGap} = 0$ but it would also make the performance on the train and test sets equally bad.} $(S=S^V)$ strikes a good balance between improving sample efficiency and reducing the $\text{GenGap}$. Indeed, our best performing mixing schedule for $(S=S^V, S^\prime=S^{\text{MI}})$ (reported here) only achieved a very marginal (and not statistically significant) improvement in test score and $\text{GenGap}$ over $(S=S^V)$.
    
    \textbf{Fifth observation.} Under $(S=S^V, P_{S^\prime}=\mathcal{U})$ training scores are similar to $(S=S^V)$ but both the $\text{GenGap}$ and $\mut{L}{b}$ increase. We hypothesise that the mechanisms responsible for improving sample efficiency and generalisation in $(S=S^V)$ are different, and that the latter is achieved through adaptive $\mut{L}{b}$ regularisation.
\subsection{Value loss prioritisation reduces $\mut{L}{b}$}

While it is intuitive that an adaptive strategy such as $(S=S^{\text{MI}})$ minimises $\mut{L}{b}$ by preventing the agent to train on informative levels, the relationship between $(S=S^{V})$ and $\mut{L}{b}$ is less evident. To understand this relationship it is helpful to remember that $(S=S^{V})$ \textit{rejects} levels with low value loss as much as it prioritises levels with high value loss; and to consider what achieving low value loss means in the CMDP setting.

We can estimate the value function $V^\pi_C$ of a CMDP from level specific value functions $V^\pi_{\lvl}$ by employing the unbiased value estimator lemma from \citet{instance_invariant}.

\begin{lemma}\label{lem:unbiased_value_estimator}
    Given a policy $\pi$ and a set of levels $L$ from a CMDP, we have $\forall H_t^o$  $(t < \infty)$ compatible with $L$ (meaning that the observation sequence $H_t^o$ occurs in $L$), $\mathbb{E}_{L|H_t^o}[V^\pi_{\lvl}(H_t^o)] = V^\pi_C(H_t^o)$, with $V^\pi_{\lvl}(H_t^o)$ being the expected returns under $\pi$ given observation history $H_t^o$ in a given level $\lvl$, and $V^\pi_C(H_t^o)$ being the expected returns across all possible occurrences of $H_t^o$ in the CMDP.
\end{lemma}
It follows that the value prediction loss also employs level-specific targets, with 
\begin{equation} \label{eq:Lv}
    L_{V}(\vtheta) = \frac{1}{N} \sum_{n}^N (\hat{V}_\vtheta({H_t^{o}}^{(n)}) - {V^\pi_{i}}^{(n)})^2,
\end{equation}
and \Cref{lem:unbiased_value_estimator} guaranteeing convergence to an unbiased estimator for $V^\pi_C$ when minimising \Cref{eq:Lv}. 

Equivalently, we can express each level-specific value function $V^\pi_{i}$ as a combination of the CMDP value function $V^\pi_C$ and a level-specific function $v^\pi_i$, \Cref{lem:unbiased_value_estimator} ensuring that the $v^\pi_i$ functions cancel out in expectation. Nevertheless, when $v^\pi_i$ is not zero everywhere, identifying the current level $i$ and predicting $v^\pi_i$ are both necessary to predict individual value targets $V^\pi_i$. Perfect value prediction over the training levels (i.e. overfitting and reaching zero value loss) therefore necessitates learning an intermediate representation from which the current level $i$ is identifiable, implying high $\mut{L}{b}$. It follows that, by rejecting levels with low value loss, value loss prioritisation prevents the agent from collecting data from the levels in which $b$ has begun to overfit. 

In other words, value loss prioritisation acts as a form of \textit{rejection sampling} that prevents data with high $\mut{L}{b}$ from being generated. This rejection sampling is adaptive, as $L_V(\vtheta)$ depends on the current learned weights $\vtheta$, and thus continually resists convergence to a representation overfitting to any particular level.
\begin{figure}[htb]
    \centering
    \begin{subfigure}[]{1.0\linewidth}
    \includegraphics[width=1\linewidth]{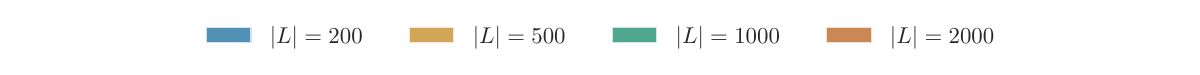}
    \vspace{-0.7cm}
    \end{subfigure}\\
    \begin{subfigure}[]{0.49\linewidth}
    \includegraphics[width=1\linewidth]{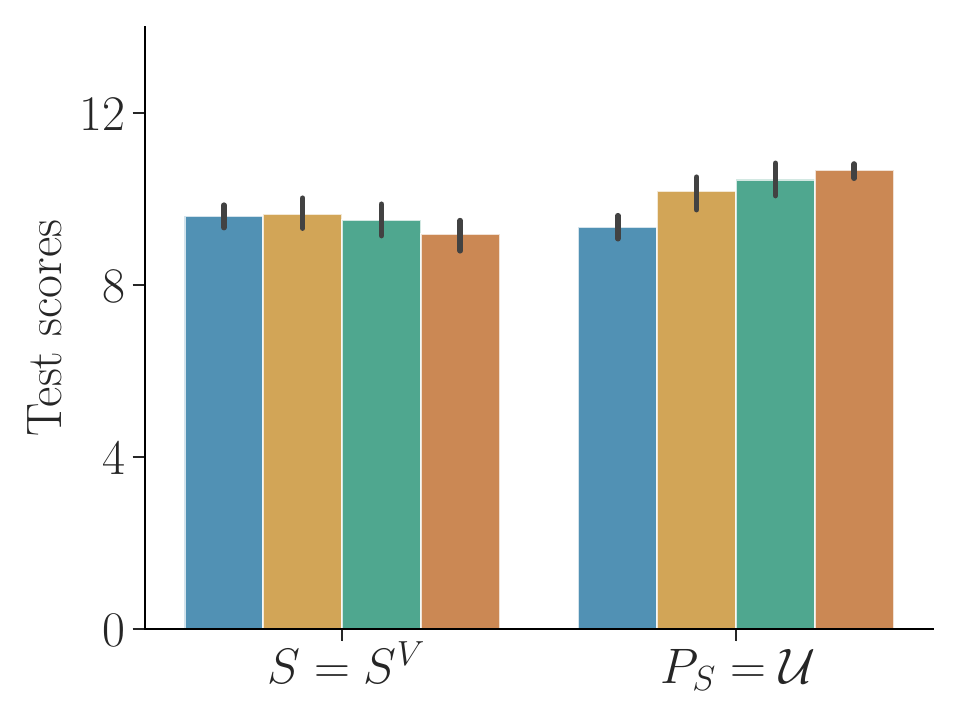}
    \end{subfigure}
    \begin{subfigure}[]{0.49\linewidth}
    \includegraphics[width=1\linewidth]{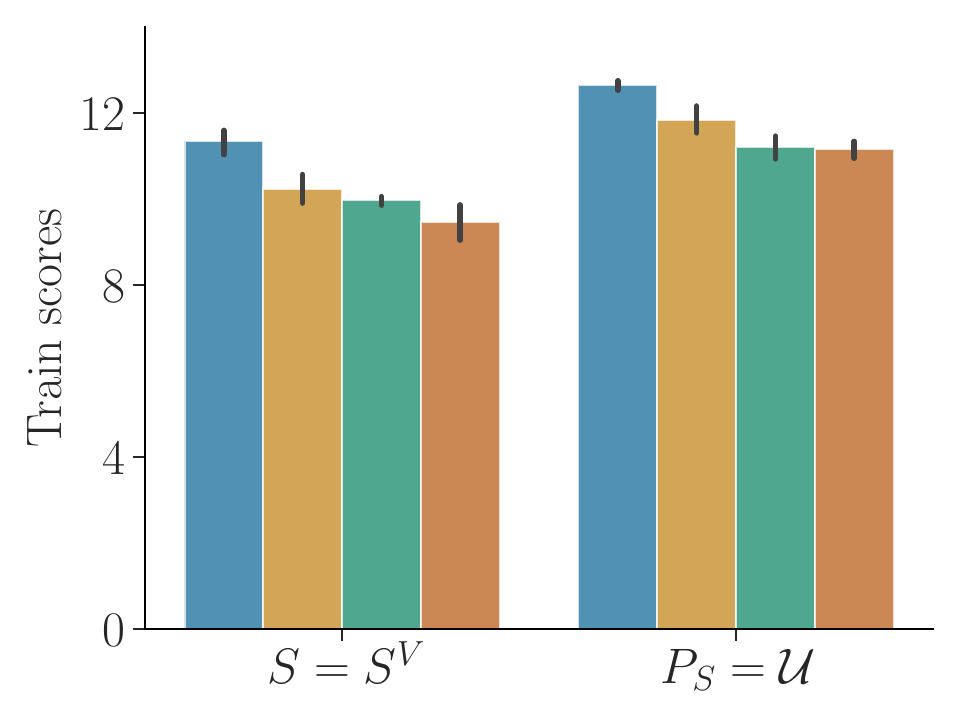}
    \end{subfigure}\\
    \begin{subfigure}[]{0.49\linewidth}
    \includegraphics[width=1\linewidth]{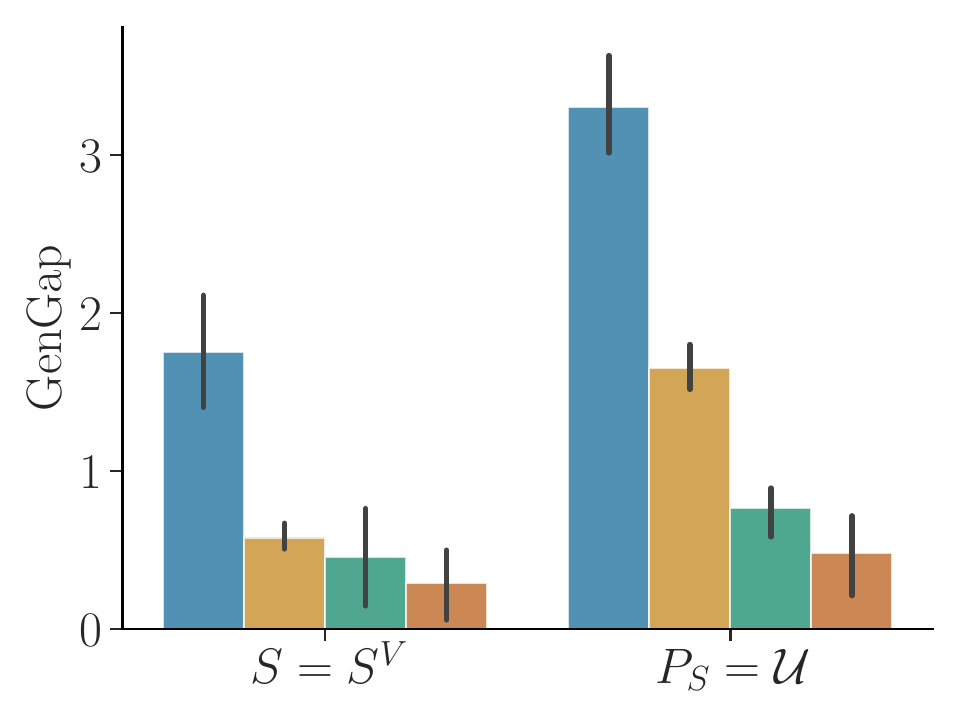}
    \end{subfigure}
    \begin{subfigure}[]{0.49\linewidth}
    \includegraphics[width=1\linewidth]{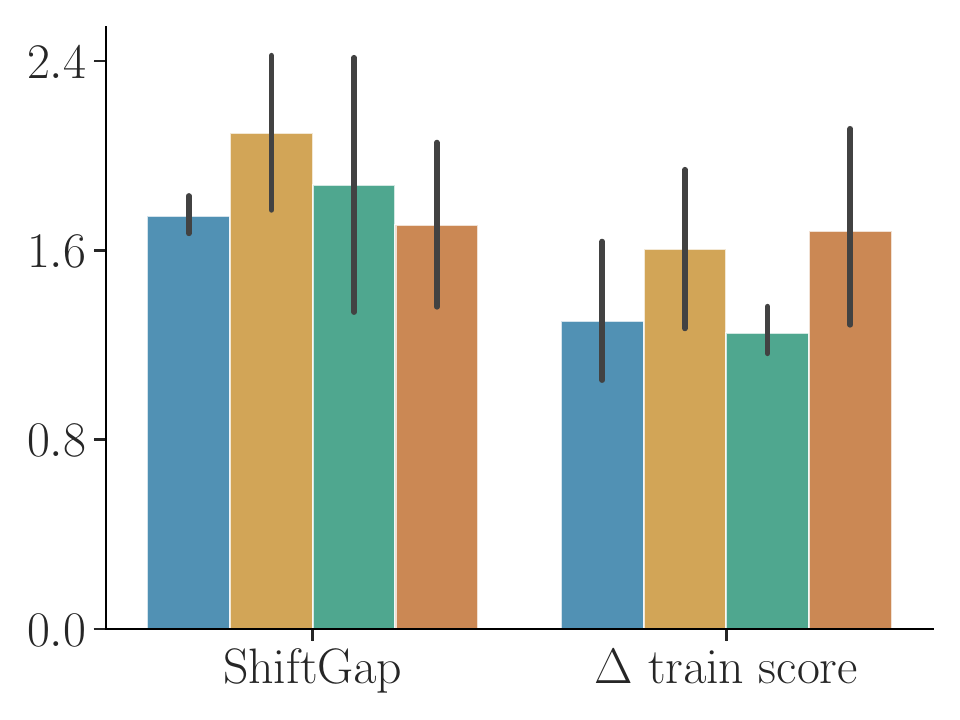}
    \end{subfigure}
    \caption{Investigating the effect of increasing $|L|$ when adaptive sampling induces distributional shift. Top row reports the test and train performance; the $\text{GenGap}$ is on the bottom left; on the bottom right is compared, for $(S=S^V)$, the $\text{ShiftGap}$ and the score differential with $(P_S=\mathcal{U})$ over the train set. Vertical bars indicate standard error over 5 seeds.}\label{fig:numlvl_exp}
\end{figure}
\subsection{Distributional shift and the effect of increasing $|L|$}
\begin{figure*}[htb]
    \centering
    \includegraphics[width=1\linewidth]{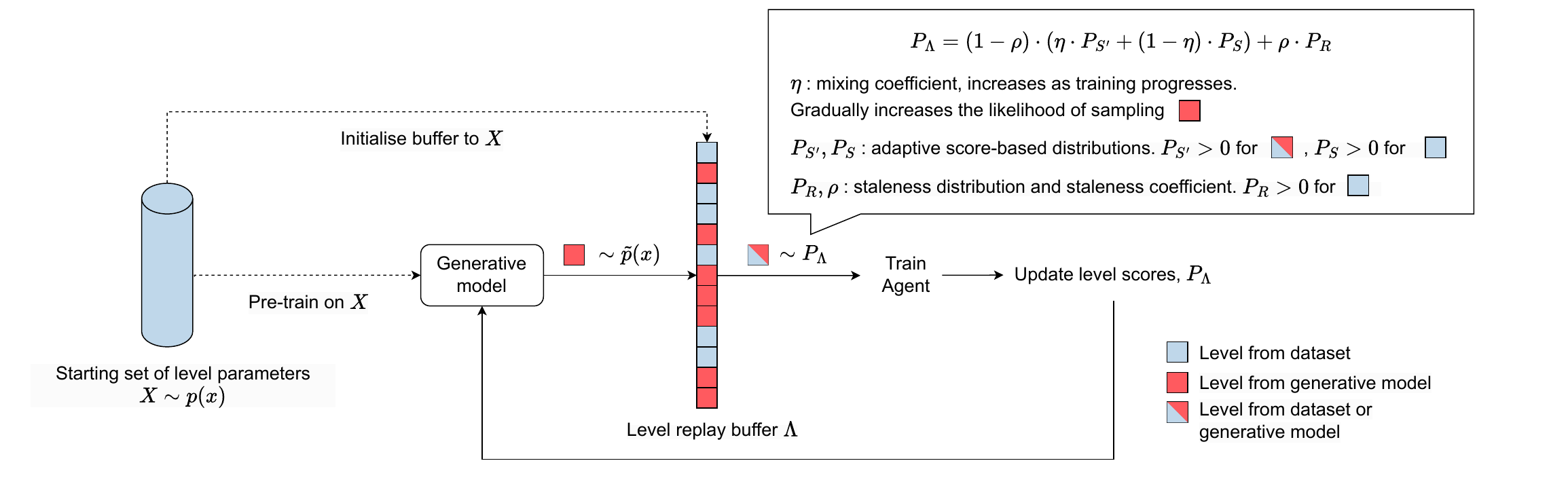}
    \caption{Overview of the data-regularised environment design framework.}
    \label{fig:DRED-diagram}
\end{figure*}
In 3 out of 16 Procgen games, $(S=S^V)$ does not improve over $(P_S=\mathcal{U})$ on the test set, and does worse on the train set (per-game results are available in \Cref{app:procgen_extended_results}). As we are interested in combining adaptive sampling with the generation of additional levels, we conduct a preliminary experiment investigating how the inclusion of additional training levels impacts ZSG in ``Miner'', one of the games in which $(S=S^V)$ underperforms. 

In \Cref{fig:numlvl_exp} we observe larger $L$ results in a reduction in the $\text{GenGap}$ and in an increase in test scores under uniform sampling, as reported in several prior studies \citep{OverfittingDRL,procgen,Packer2018AssessingGI}. For $(S=S^V)$ we also observe a reduction in $\text{GenGap}$ but it is caused by scores dropping over the train set instead of improving over the test set. This phenomenon is not optimisation related, as train scores for $(S=S^V)$ under its own adaptive distribution remain in-line with the $(P_S=\mathcal{U})$ train scores. In fact it is caused by distributional shift, a potential pitfall of adaptive distributions. In Miner, the shifted training distribution prevents the agent to improve its test scores further, even as the number of training levels is increased. 

As this phenomenon is not captured by the $\text{GenGap}$, %
we propose the \textit{shift-induced gap}, or $\text{ShiftGap}$, a complementary metric quantifying the performance reduction induced by distributional shift under non-uniform training distributions,
\begin{equation} \label{eq:ShiftGap}
    \text{ShiftGap}(\pi) \coloneqq \sum_{i \in \Lambda} P_\Lambda(i) \cdot V^\pi_{\lvl} - \frac{1}{|L|} \sum_{i \in L} V^\pi_{\lvl}.
\end{equation}
Unlike the $\text{GenGap}$, the $\text{ShiftGap}$ does not necessitates held-out test levels to be measured, and only relies on scores being normalised across levels.\footnote{Note that using $X_\text{test}$ in the second term makes $\text{ShiftGap}\equiv\text{GenGap}$ when $P_\Lambda = \mathcal{U}$, whereas it should be 0.} We find that the $\text{ShiftGap}$ for $(S=S^V)$ closely matches (within measurement error) the performance differential between $(S=S^V)$ and $(P_S=\mathcal{U})$ over the train set. In later experiments, we will show it is important for the $\text{ShiftGap}$ to remain small in order to achieve strong ZSG.
\begin{algorithm}[!htb]
\centering
\caption{Data-regularised environment design}\label{alg:DRED}
\begin{algorithmic}[1]
\Input Pre-trained encoder and decoder networks $\psi_{\vtheta_E}$, $\phi_{\vtheta_D}$, level parameters $X_\text{train}$, number of pairs $M$, number of interpolations per pair $K$
\State Initialise level buffer $\Lambda$ to level parameters in $X_\text{train}$ 
\State Initialise agent policy $\pi$
\While{\textit{not converged}}
    \State Instantiate levels from batch $X \sim P_\Lambda$ and collect rollouts, update level scores $S, S^\prime$ in $\Lambda$
    \State Update $\pi$ using collected rollouts
    \State Uniformly sample $M$ pairs from $X_\text{train}$
    \For {$(\vx_{i}, \vx_{j})$ in pairs} %
        \State Compute latent parameters using $\psi_{\vtheta_E}$ and $K$ interpolations between $(\vmu_\rvz, \vsigma_\rvz)_{i}$ and $(\vmu_\rvz, \vsigma_\rvz)_{j}$
        \For {$(\vmu_\rvz, \vsigma_\rvz)_k$ in interpolations}
            \State Sample embedding $\vz \sim \mathcal{N}(\vmu_\rvz, \vsigma_\rvz)$
            \State Instantiate $\tilde{\vx} \gets \phi_{\vtheta_D}(\vz)$ and compute $s, s^\prime$
            \State Add $\langle \tilde{\vx}, s, s^\prime \rangle$ to $\Lambda$ %
        \EndFor
    \EndFor
\EndWhile
\end{algorithmic}
\end{algorithm}
\section{Data-regularised environment design}\label{sec:ssed}
We have established that we can reduce $\mut{L}{b}$ and the $\text{GenGap}$ by increasing $|L|$ or by employing an adaptive sampling strategy. However, we have observed that increasing $|L|$ may not result in higher test set performance when there is significant distributional shift during training. As the CMDP context distribution $p(\rvx)$ is rarely known in practical applications, artificially generating extra training levels is an additional source of distributional shift. To capitalise on the benefits provided by adaptive sampling and level generation, while limiting distributional shift, we propose \textit{data-regularised environment design} (DRED), a framework that combines adaptive sampling with a level generation process approximating $p(\rvx)$.

Instead of direct knowledge of $p(\rvx)$, we assume having access to a limited set of level parameters $X_\text{train} \sim p(\rvx)$. Each $\vx \in X_\text{train}$ instantiates a level $i_\vx$ from the CMDP. We are allowed to sample from the full simulator parameter space $\sX$, which means we can \textit{augment} our set of training levels with new levels $\tilde{\rvx} \in \sX$.

DRED consists of two components: a \textit{generative phase}, in which an augmented set $\tilde{X}$ is generated from a batch $X \sim \mathcal{U}(X_\text{train})$ and is added to the buffer $\Lambda$, and a \textit{replay phase}, in which we use the adaptive distribution $P_\Lambda$ to sample levels from the buffer. We alternate between the generative and replay phases, and only perform gradient updates on the agent during the replay phase. \Cref{alg:DRED} and \Cref{fig:DRED-diagram} describe the full DRED pipeline, and we provide further details on each phase below.
\subsection{The generative phase}
We initialise the buffer $\Lambda$ to contain $X_\text{train}$ levels and gradually add generated levels $\tilde{X}$ over the course of training. DRED is not restricted to a particular approach to obtain $\tilde{X}$. In this work, we use a VAE \citep{VAE, VAE-pmlr-v32-rezende14} and refer to our method as VAE-DRED. The VAE models the underlying training data distribution $p(\rvx)$ %
as stochastic realisations of a latent distribution $p(\rvz)$ via a generative model $p(\rvx \mid \rvz)$.
The model is pre-trained on $X_\text{train}$ by maximising the variational ELBO
\begin{equation} \label{eq:ELBO}
\begin{split}
    \mathcal{L}_{\text{ELBO}} = & \mathop{\mathbb{E}}_{\vx\sim p(\rvx)} [  \mathop{\mathbb{E}}_{\vz \sim q(\rvz|\rvx; \psi_{\vtheta_E})}[\log p(\rvx \mid \rvz; \phi_{\vtheta_D})] \\
    & - \beta \KL(q(\rvz \mid \rvx;\psi_{\vtheta_E})\mid\mid p(\rvz))],
\end{split}
\end{equation}
where $q(\rvz \mid \rvx; \psi_{\vtheta_E})$ is a variational approximation of an intractable model posterior distribution $p(\rvz \mid \rvx)$ and $\KL(\cdot \mid\mid \cdot)$ denotes the Kullback--Leibler divergence, which is balanced using the coefficient $\beta$, as proposed by \citet{betaVAE}. 
The generative $p(\rvx \mid \rvz; \phi_{\vtheta_D})$ and variational $q(\rvz \mid \rvx ; \psi_{\vtheta_E})$ models are parametrised via encoder and decoder networks $\psi_{\vtheta_E}$ and $\phi_{\vtheta_D}$.%

Maximising \Cref{eq:ELBO} fits the VAE such that $p(\rvx; \phi_{\vtheta_D}) = \int p(\rvx \mid \rvz; \phi_{\vtheta_D}) p(\rvz) \dif \rvz \approx p(\rvx)$. Out-of-context levels are less likely with $\tilde{\vx} \sim p(\rvx; \phi_{\vtheta_D})$ than with $\tilde{\vx} \sim \mathcal{U}(\sX)$, and we show in \Cref{sec:minigrid} that this aspect is key in enabling DRED agents to outperform UED agents. %
We follow \citet{white2016sampling} and interpolate $\rvz$ in the latent space between the latent representations of a pair of samples $(\vx_i, \vx_j) \sim X_\text{train}$ (instead of sampling $\rvz$ from $p(\rvz)$), as this improves the quality of the generated $\tilde{\vx}$.

After generating a batch of levels parameters $\tilde{X}$ we collect rollouts (without updating agent weights) to compute their scores, adding to the buffer $\Lambda$ any level scoring higher than the lowest scoring generated level in $\Lambda$. We only consider levels solved at least once during rollouts for inclusion, to ensure that unsolvable levels do not get added in. We provide additional details on the VAE architecture, hyperparameters and pre-training process in \Cref{app:vae_imp}.

\subsection{The replay phase}
All levels in $X_\text{train}$ originate from $p(\rvx)$ and are in-context, whereas generated levels, which are obtained from an approximation of $p(\rvx)$, do not benefit from as strong of a guarantee. As training on out-of-context levels can significantly harm the agents' performance on the CMDP, we control the ratio between $X_\text{train}$ and augmented levels using the mixed scoring introduced in \Cref{eq:buffer_dist} for $P_\Lambda$. The support of $P_S$ and $P_R$ is limited to $X_\text{train}$, whereas $P_{S^\prime}$ supports the entire buffer. We set both $S$ and $S^\prime$ to score levels according to the $\ell_1$-value loss. We find out-of-context levels to be particularly harmful in the early stages of training, and limit their occurrence early on by linearly increasing $\eta$ from 0 to 1 over the course of training.
\section{DRED experiments}\label{sec:minigrid}
Our experiments seek to answer the following questions: 1)~How important is it to remain grounded to the target CMDP when generating additional levels, instead of simply maximising level diversity? 2) Is DRED successful in grounding the training distribution to the target CMDP, and does it improve transfer to held-out levels and edge-cases?
\subsection{Experimental setup}
 We choose Minigrid \citep{gym_minigrid}, a partially observable gridworld navigation domain, for our experiments. Despite its simplicity, Minigrid has a controllable level parameter space (unlike Procgen), and levels are parameterised to vectors describing the locations, starting states and appearance of the objects in the grid. 
 
When benchmarking UED methods (and RL algorithms in general) it is implicitly agreed upon that each (solvable) level instantiated belongs to the target CMDP. Yet, this is rarely true in a practical application, as in-context level parameters $\sX_C$ often correspond to a small and highly structured region of the simulator parameter space $\sX$. With this is mind, we seek to design a target CMDP with similar properties in Minigrid. We define the context space of our target CMDP as spanning the layouts where the location of green ``moss'' tiles and orange ``lava'' tiles are respectively positively and negatively correlated to their distance to the goal location. 
We employ procedural generation to obtain a set $X_\text{train}$ of 512 level parameters (we refer the reader to \Cref{fig:layouts_dataset} for a visualisation of levels from $X_\text{train}$, and to \Cref{app:cmdp_levelset} for extended details on the CMDP specification and procedural generation processes).

As the agent only observes its immediate surroundings and does not know the goal location a priori, the optimal CMDP policy is one that exploits the semantics shared by all levels in the CMDP, exploring first areas with high perceived moss density and avoiding areas with high lava density. Other CMDPs exist in the level space, and may correspond to incompatible optimal policies (for example a CMDP in which the correlation of moss and lava tiles with the goal is reversed). As such, it is important to maintain consistency with the CMDP semantics when generating new levels.

Our first set of baselines is restricted to sample from $X_\text{train}$, and consists of uniform sampling ($\mathcal{U}$) and sampling using the $\ell_1$-value loss strategy (PLR). Our second set uses level generation, removing this restriction. We consider domain randomisation (DR) \citep{DR} which generates levels by sampling uniformly between pre-determined ranges of parameters; RPLR \citep{Robust-PLR}, which combines PLR with DR used as its generator; and the current UED state-of-the-art, ACCEL \citep{ACCEL}, an extension of RPLR replacing DR by a generator making local edits to currently high scoring levels in the buffer. %
In all experiments we train the same PPO \citep{PPO} agent for 27k updates. Additional details on our implementation are provided in \Cref{app:impl_minigrid}.
\subsection{Results}
\begin{figure*}[tb]
\centering
    \begin{subfigure}[]{.08\linewidth}
    \centering
    \includegraphics[width=1\linewidth]{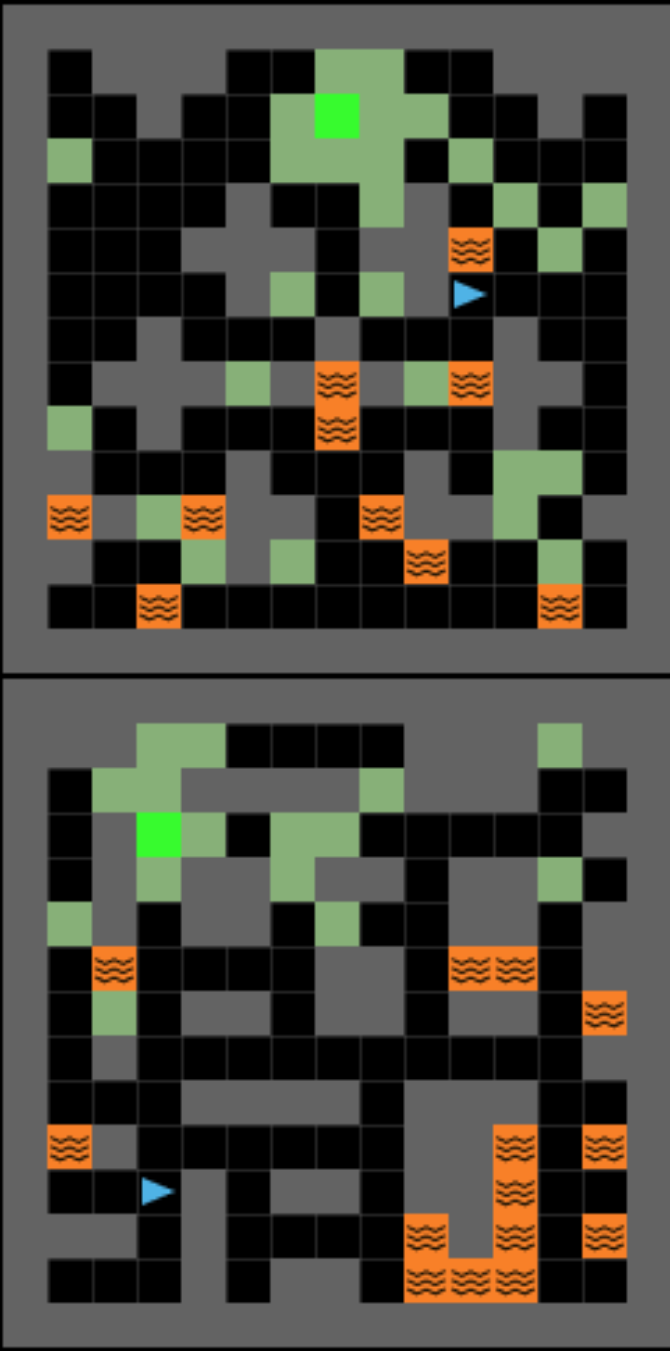}
  \end{subfigure}%
   \begin{subfigure}[]{.72\linewidth}
    \centering
    \includegraphics[width=1\linewidth]{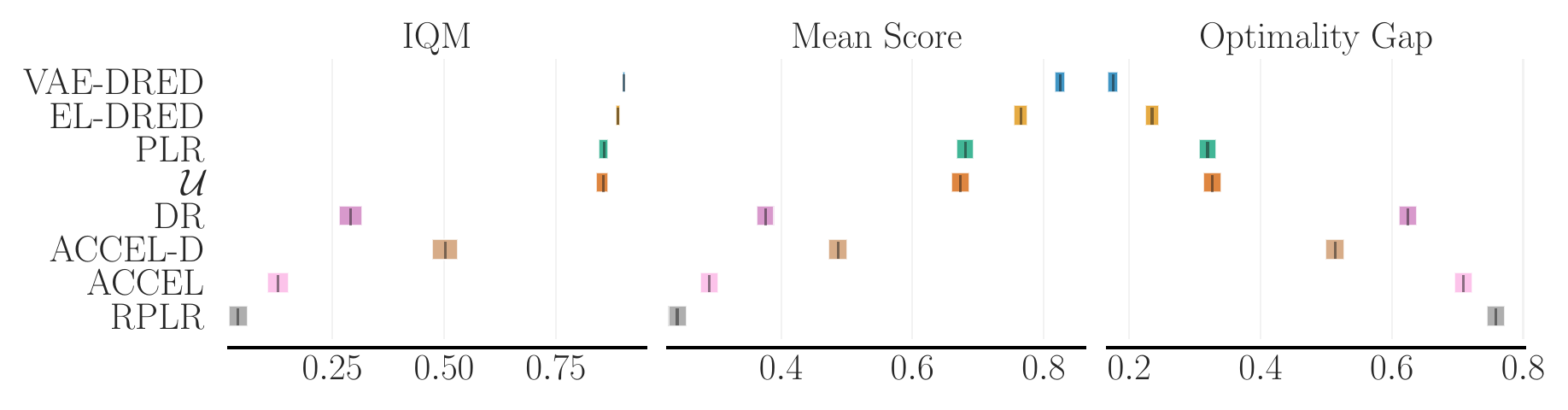}
  \end{subfigure}%
    \begin{subfigure}[]{.2\linewidth}
    \centering
    \includegraphics[width=1\linewidth]{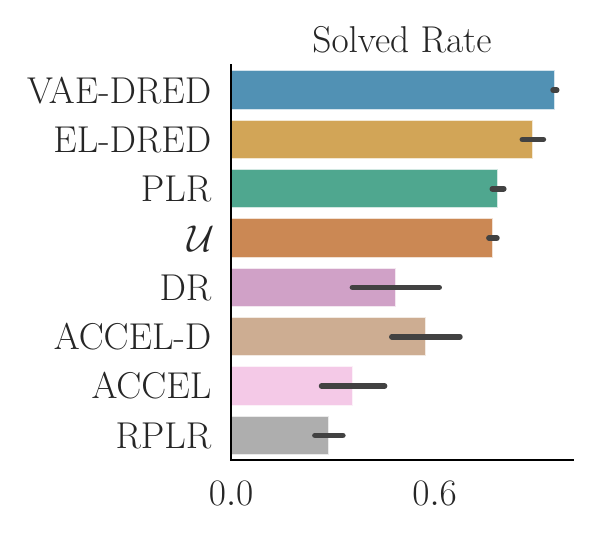}
    \end{subfigure}%
    
    \begin{subfigure}[]{.08\linewidth}
    \centering
    \includegraphics[width=1\linewidth]{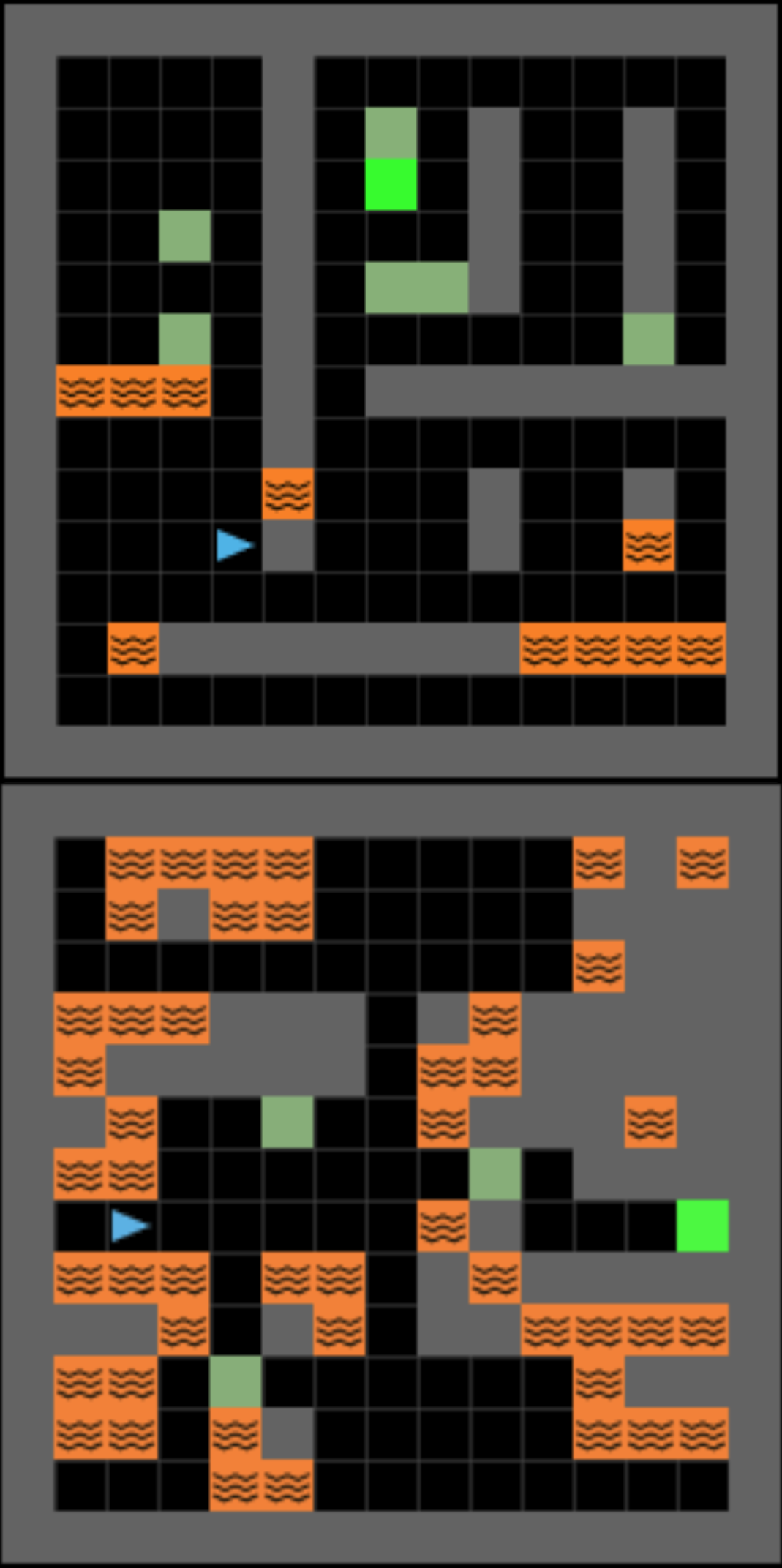}
  \end{subfigure}%
   \begin{subfigure}[]{.72\linewidth}
    \centering
    \includegraphics[width=1\linewidth]{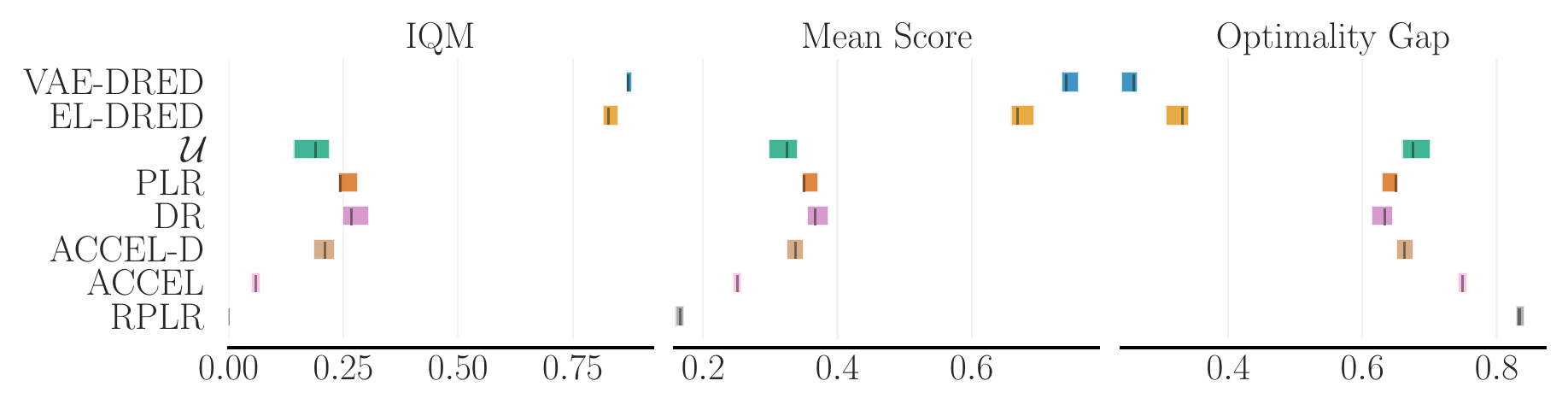}
  \end{subfigure}%
    \begin{subfigure}[]{.2\linewidth}
    \centering
    \includegraphics[width=1\linewidth]{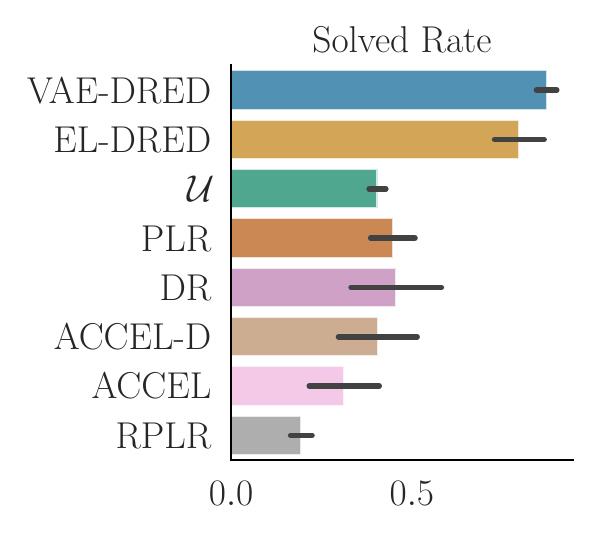}
    \end{subfigure}%
\caption{Aggregate final performance and mean solved rate on $X_\text{test}$, an evaluation set of 2048 levels sampled from $p(\rvx)$ and held-out during training (top), and on 448 in-context edge cases (bottom). Example layouts from each evaluation set are plotted on the left. The coloured boxes indicate a 99\% confidence interval and the black horizontal bars indicate standard error across 5 training seeds. We refer the reader to \Cref{app:cmdp_levelset} for additional details on our evaluation sets.}
\label{fig:zsg_test_set_and_edge_cases}
\end{figure*}
\textbf{ZSG to held-out levels.} As shown in \Cref{fig:zsg_test_set_and_edge_cases} (top), VAE-DRED achieves statistically significant improvements in its IQM (inter-quantile mean), mean score, optimality gap (compared to the level-specific optimal policy) and mean solved rate over other methods on held-out levels from the CMDP. VAE-DRED drastically increases the number of training levels available to the agent while remaining consistent with the target CMDP. VAE-DRED maintains a small distributional shift, which we quantify in our extended analysis in \Cref{app:dist-shift}, and a low $\text{ShiftGap}$ throughout training (\Cref{fig:gen_gaps}). This is thanks to its generative model effectively approximating $p(\rvx)$, and to its mixed sampling strategy ensuring levels from $X_\text{train}$ are sampled often, and even more-so early-on. Despite small $\text{GenGap}$ (\Cref{fig:gen_gaps}), UED baselines achieve low test scores, as they perform poorly on both the test set and on $X_\text{train}$.  Further analysis conducted in \Cref{app:dist-shift} confirms that, in general, UED methods incur larger distributional shift than DRED or adaptive sampling strategies, and result in larger $\text{ShiftGap}$.

\textbf{ZSG to edge cases.} We next investigate whether VAE-DRED's level generation improves robustness to in-context \textit{edge cases} with a near zero likelihood of occurring in $X_\text{train}$ (\Cref{fig:zsg_test_set_and_edge_cases}, bottom). We find VAE-DRED to be particularly dominant in this setting, achieving over three times DR's IQM, the next best method, and twice its solved rate and mean score. VAE-DRED makes the agent robust to edge cases by introducing additional diversity in the training levels. These generated levels remain consistent with the CMDP semantics, as can be observed qualitatively in renderings of generated levels in \Cref{fig:interpolated_levels} and quantitatively in the training distribution metrics reported in in \Cref{fig:buffermetrics}.

\begin{figure}[htb]
\centering
    \begin{subfigure}[]{.49\linewidth}
    \centering
    \includegraphics[width=1\linewidth]{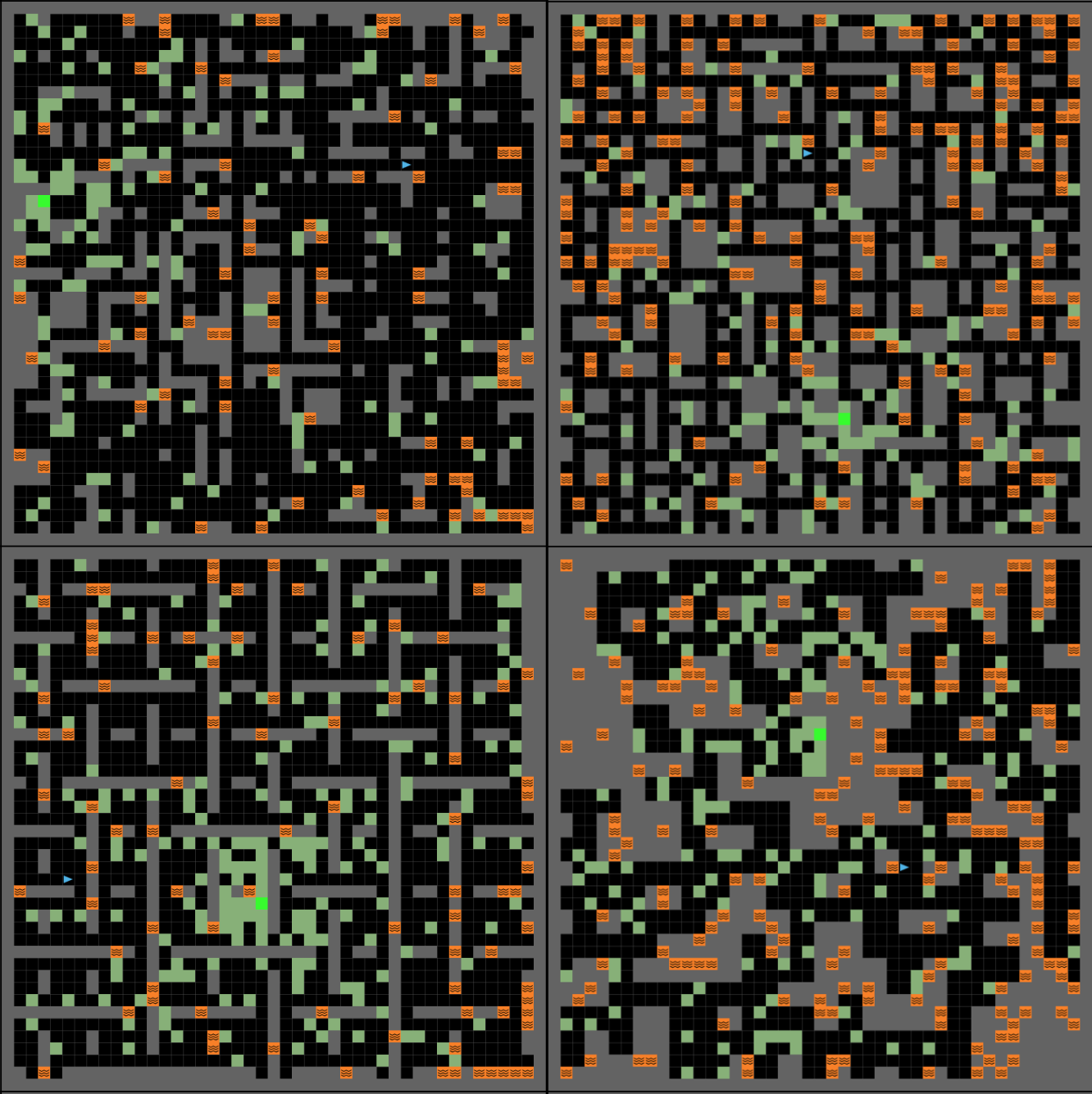}
  \end{subfigure}%
    \begin{subfigure}[]{.49\linewidth}
    \centering
    \begin{subfigure}[]{1\linewidth}
    \centering
    \includegraphics[width=.8\linewidth]{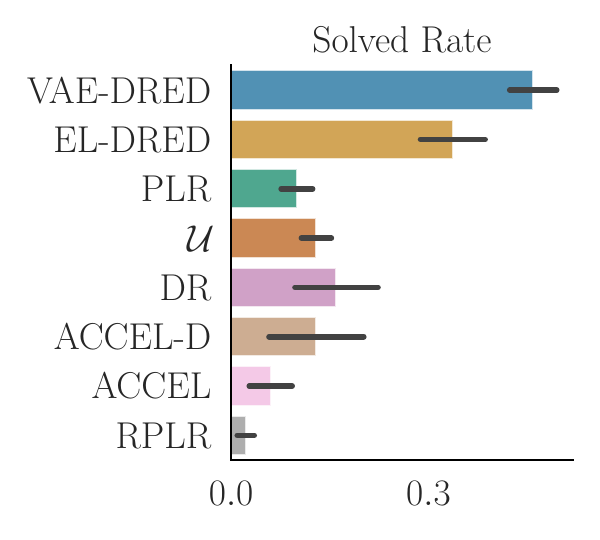}
    \end{subfigure}\\

    \begin{subfigure}[]{1\linewidth}
    \includegraphics[width=1\linewidth]{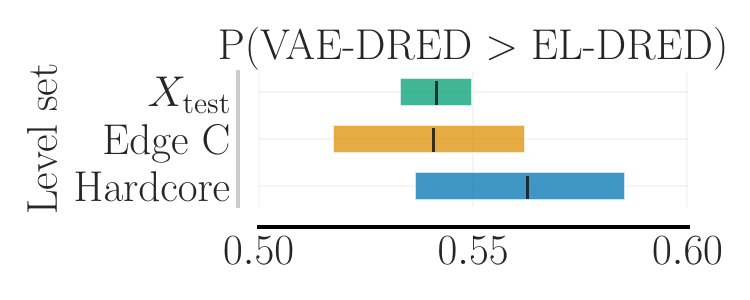}
    \end{subfigure}
    \end{subfigure}%
\caption{Top right: Solved rate achieved on 216 ``Hardcore'' levels (examples plotted on the left). Bottom right: Probability of VAE-DRED achieving higher zero-shot returns than its ablation EL-DRED, on different evaluation sets. Coloured boxes indicate the 99\% confidence interval. %
}
\label{fig:zsg_hard}
\end{figure}

\textbf{ZSG to hard levels.} In \Cref{fig:zsg_hard,fig:zsg_hard_set_and_edge_cases}, we evaluate transfer to in-context ``Hardcore'' levels. Being 9 times larger in area than training levels, Hardcore levels are significantly more
challenging to solve, even for Humans. This setting is where the performance gap between VAE-DRED and other methods is the largest, with VAE-DRED solving three times as many levels as the next best baseline.

\textbf{Ablations.} To better understand the importance of the pre-trained generative model, we introduce EL-DRED, which replaces the VAE with ACCEL's local edit strategy. EL-DRED may be viewed as a DRED variant of ACCEL augmenting $X_\text{train}$ using a non-parametric generative method, or, equivalently, as an ablation of VAE-DRED that does not approximate $p(\rvx)$, and is therefore less grounded to the CMDP. EL-DRED outperforms all other methods in each of the level sets depicted above, with the exception of VAE-DRED. In \Cref{fig:zsg_hard} (bottom right), we compare the two methods, and find that VAE-DRED remains significantly more likely to outperform EL-DRED in each level set. Finally, ACCEL-D shows that initialising the buffer to $X_\text{train}$ isn't sufficient for preventing ACCEL's editing mechanism to rapidly incur significant distributional shift. The only difference between EL-DRED and ACCEL-D is $\eta$ being set to 1 throughout training (see the text box in \Cref{fig:DRED-diagram} for a depiction of the role of $\eta$ in DRED). Yet, the gap in performance is significant, and highlights the importance of avoiding out-of-context levels early in training.%
\section{Related work}\label{sec:rw}
\textbf{Buffer-free sampling strategies.} Domain randomisation \citep[DR,][]{DR, DR1997}, is one of the earliest proposed methods for improving the generalisation ability of RL agents by augmenting the set of available training levels. It does so by sampling uniformly between manually specified ranges of environment parameters. Subsequent contributions introduce an implicit prioritisation over the generated set by inducing a minimax return \citep[robust adversarial RL,][]{Pinto2017RobustAdversarialRL} or a minimax regret game \citep[UED,][]{PAIRED} between the agent and a \textit{level generator}, which are trained concurrently. These adversarial formulations prioritise levels in which the agent is currently performing poorly to encourage robust generalisation over the sampled set, with UED achieving better Nash equilibrium theoretical guarantees. CLUTR \citep{CLUTR} removes the need for domain-specific RL environments and improves sample efficiency by having the level generator operate within a low dimensional latent space of a generative model pre-trained on randomly sampled level parameters. However, buffer-free methods remain vastly outperformed by a well calibrated DR implementation and the buffer-based sampling strategies discussed next.

\textbf{Buffer-based sampling strategies.} Prioritised sampling is often applied to off-policy algorithms, where individual interactions within the replay buffer are prioritised \citep{Schaul2015PrioritizedER} or resampled with different goals in multi-goal RL \citep{HER,zhang2020automatic}. Prioritised Level Replay \citep[PLR,][]{PLR} instead affects the sampling process of \textit{future} experiences, and is thus applicable to both on- and off-policy algorithms. PLR maintains a buffer of training levels and empirically demonstrates that prioritising levels using a scoring function proportional to high value prediction loss results in higher sample efficiency and improved ZSG performance. Robust PLR \citep[RPLR,][]{Robust-PLR} extends PLR to the UED setting, using DR as its level generation mechanism, while ACCEL \citep{ACCEL} gradually evolves new levels by performing random edits on high scoring levels in the buffer. SAMPLR \citep{SAMPLR} proposes to  eliminate the distributional shift induced by the prioritisation strategy by modifying individual interactions using a second simulator that runs in parallel. However, SAMPLR only applies to settings with direct access to the ground truth context distribution, while DRED learns to approximate this distribution.

\textbf{Mutual information minimisation in RL.} In prior work, mutual information has been minimised in order to mitigate instance-overfitting, either by learning an ensemble of policies \citep{instance_invariant,E-POMDP}, performing data augmentation on observations \citep{raileanu2021UCB-DrAC,Kostrikov2021ImageAllYouNeed}, an auxiliary objective \citep{mhairiCMID,dunion2024mvd} or introducing information bottlenecks through selective noise injection on the agent model \citep{vib_actor_critic,coinrun}. In contrast, our work is the first to draw connections between mutual information minimisation and adaptive level sampling.
\section{Conclusion}\label{sec:disc}
In this work, we investigated the impact of the level sampling process on the ZSG capabilities of RL agents. We found adaptive sampling strategies are best understood as data regularisation techniques minimising the mutual information between the agent's internal representation and the identity of training levels. In doing so, these methods minimise an upper bound on the generalisation gap, and our experiments show that this bound acts as an effective proxy for reducing this gap in practice. This theoretical framing allowed us to understand the mechanism behind the improved generalisation achieved by value loss prioritised level sampling, which had only been justified empirically in prior work. We introduced DRED, a framework combining adaptive sampling with the generation of new levels using a learned model of the context distribution. We propose VAE-DRED, an application of DRED using a VAE to learn the context distribution. Our experiments show that VAE-DRED prevents the significant distributional shift observed in other UED methods. By jointly achieving low $\text{GenGap}$ and $\text{ShiftGap}$, VAE-DRED achieves strong generalisation performance on in-distribution test levels, while also being robust to in-context edge cases.

In future work, we plan to investigate how DRED methods perform in more complex environments. Our experiments show that unsupervised environment generation can be problematic, even in gridworlds, and these issues are bound to worsen when the environment parameter space has higher complexity and dimensionality. DRED possesses the ability to leverage an existing dataset to inform its generative process, which we believe will be instrumental in scaling environment design techniques to practical applications. We are particularly interested in studying how DRED could leverage real world datasets of level parameters that have started to become available. \citet{OpenRooms} introduced a dataset of indoor environments geared towards robotics and embodied AI tasks, \citet{argoverse2} published city maps for autonomous driving while \citet{metabox,protein_meta_opt} published a dataset of protein docking problems. Level parameters remain costly to collect or prescribe manually, and thus these datasets remain much smaller in size than text or image datasets. In maximising the generalisation potential of a limited number of training environments, we hope DRED can reduce start-up costs associated with extending RL to new practical applications.

\section*{Impact statement}
This paper presents work whose goal is to advance the field of Machine Learning. There are many potential societal consequences of our work, none which we feel must be specifically highlighted here.

We believe parametrisable simulators are better suited to benchmark RL algorithms than procedural environments, as they provide a fine degree of control over the environment and are more consistent with a realistic application setting, as argued by \citet{kirk2023survey}. However, reproducibility can be challenging without access to the data generated during experiments. To assist with this, we make all of our experimental data, including model checkpoints, level datasets, logged data and the code for reproducing the figures in this paper openly available.

We open-source our code for specifying arbitrary CMDPs in Minigrid and generate their associated level sets (we describe the generation process in detail in \Cref{app:cmdp_levelset}). We also provide a dataset of 1.5M procedurally generated minigrid base layouts to facilitate level set generation.%

\section*{Acknowledgements}
This work was supported by the United Kingdom Research and Innovation EPSRC Centre for Doctoral Training in Robotics and Autonomous Systems (RAS) in Edinburgh.

\bibliography{basebibfile}
\bibliographystyle{icml2024}

\newpage
\onecolumn
\appendix

\input{supplementary}

\end{document}

%% file: math_commands.tex
\usepackage{amsmath,amsfonts,bm}

\def\eqref#1{equation~\ref{#1}}

\def\1{\bm{1}}

\def\ri{{\textnormal{i}}}

\def\rs{{\textnormal{s}}}

\def\rvx{{\mathbf{x}}}

\def\rvz{{\mathbf{z}}}

\def\vmu{{\bm{\mu}}}
\def\vsigma{{\bm{\sigma}}}

\def\vtheta{{\bm{\theta}}}

\def\vh{{\bm{h}}}

\def\vx{{\bm{x}}}

\def\vz{{\bm{z}}}

\DeclareMathAlphabet{\mathsfit}{\encodingdefault}{\sfdefault}{m}{sl}
\SetMathAlphabet{\mathsfit}{bold}{\encodingdefault}{\sfdefault}{bx}{n}

\def\sA{{\mathbb{A}}}
\def\sB{{\mathbb{B}}}

\def\sO{{\mathbb{O}}}

\def\sR{{\mathbb{R}}}
\def\sS{{\mathbb{S}}}

\def\sX{{\mathbb{X}}}

\newcommand{\KL}{D_{\mathrm{KL}}}

\newcommand{\mut}[2]{\text{I}(#1;#2)} 
\newcommand{\lvl}{i}

\newcommand{\entropy}{\mathbf{H}}

%% file: supplementary.tex
\section{Theoretical results}\label{app:proofs}

\begin{lemma}
    Given a set of training levels $L$ and an agent model $\pi = f \circ b$, where $b(H^o_t) = h_t \in \sB$ is an intermediate representation function and $f : \sB \rightarrow \Delta^\sA$ maps to the agent's action distribution, we have
    \begin{align}
        \mut{L}{\pi} \leq \entropy(\ri) + \sum_{i\in L} \int_{\sB} p(h, i) \log p(i|h)dh,
    \end{align}
    where the right-hand side is equivalent to $\mut{L}{b}$.

    proof:
    
    Since the information chain of our model follows $H_t^o \rightarrow b \rightarrow f$, we have $\mut{L}{f \circ b} \leq \mut{L}{b}$, from the data processing inequality. $\mut{L}{b}$ can then be manipulated as follows
    \begin{align}\label{eq:proof_mi_kld}
        \mut{L}{b} &= \sum_{i \in L} \int_{\sB} p(h,i) \log \frac{p(h,i)}{p(h)p(i)}dh \\
        &= - \sum_{i \in L} \int_{\sB} p(h,i) \log p(i) dh + \sum_{i \in L} \int_{\sB} p(h, i) \log p(i|h) dh \\
        &= \entropy(\ri) + \sum_{i\in L} \int_{\sB} p(h, i) \log p(i|h)dh.
    \end{align}

\end{lemma}

\section{Procgen additional experimental results}\label{app:results_procgen}

\subsection{Measuring the relationship between $\mut{L}{b}$ and the $\text{GenGap}$ across sampling methods and procgen games}\label{app:mi_gengap}

\begin{table}[!htb]
\caption{$\mut{L}{b}$ measured under different adaptive sampling strategies. We report aggregated $\mut{L}{b}$ across 5 training runs, each initialised with a different seed. To compute $\mut{L}{b}$ for each run and environment, we first fit a linear classifier $p_\vtheta$ to predict level identities from the agent's penultimate layer outputs, using rollouts from levels sampled uniformly from $L$. We then estimate $\mut{L}{b}$ using \Cref{eq:mi_empirical_approx}, using a different set of rollouts also sampled uniformly from $L$. In the last row we compute the mean $\mut{L}{b}$  across environments for each run, and we report the mean and standard deviation of that quantity across all runs. Bolded methods are not significantly different from the method with lowest mean ($p<0.05$), unless all are, in which case none are bolded.}
\label{tb:procgen_mi}
\vskip 0.15in
\begin{center}
\begin{small}
\begin{sc}
\begin{tabular}{lrrrrr}
\toprule
               Environment &                  $S=S^V$ &        $P_S=\mathcal{U}$ &      $S=S^{\mathrm{MI}}$ & $S=S^V, P_{S^\prime}=\mathcal{U}$ & $S=S^V, S^\prime=S^{\mathrm{MI}}$ \\
\midrule
                   Bigfish & \textbf{1.74 $\pm$ 0.36} &          4.33 $\pm$ 0.48 & \textbf{1.45 $\pm$ 0.36} &                   2.43 $\pm$ 0.53 &          \textbf{1.77 $\pm$ 0.70} \\
                     Heist &          4.67 $\pm$ 0.31 & \textbf{3.68 $\pm$ 0.36} & \textbf{4.08 $\pm$ 0.50} &                   4.54 $\pm$ 0.37 &                   4.76 $\pm$ 0.19 \\
                   Climber &          4.03 $\pm$ 0.23 &          4.40 $\pm$ 0.18 & \textbf{2.44 $\pm$ 0.44} &                   4.48 $\pm$ 0.15 &                   4.09 $\pm$ 0.17 \\
                 Caveflyer &          3.01 $\pm$ 0.17 &          3.94 $\pm$ 0.21 & \textbf{1.72 $\pm$ 0.20} &                   3.38 $\pm$ 0.14 &                   3.00 $\pm$ 0.18 \\
                    Jumper &          4.49 $\pm$ 0.13 & \textbf{3.87 $\pm$ 0.40} & \textbf{3.38 $\pm$ 0.38} &                   4.37 $\pm$ 0.19 &                   4.39 $\pm$ 0.07 \\
                  Fruitbot & \textbf{0.15 $\pm$ 0.09} &          2.76 $\pm$ 0.11 &          0.40 $\pm$ 0.17 &                   0.24 $\pm$ 0.09 &          \textbf{0.09 $\pm$ 0.07} \\
                   Plunder &          1.46 $\pm$ 0.22 &          2.95 $\pm$ 0.74 & \textbf{1.09 $\pm$ 0.20} &                   1.80 $\pm$ 0.41 &                   1.76 $\pm$ 0.22 \\
                   Coinrun & \textbf{1.38 $\pm$ 0.07} &          2.29 $\pm$ 0.19 & \textbf{1.30 $\pm$ 0.14} &                   1.59 $\pm$ 0.05 &          \textbf{1.36 $\pm$ 0.15} \\
                     Ninja &          2.36 $\pm$ 0.30 &          2.62 $\pm$ 0.24 & \textbf{1.42 $\pm$ 0.44} &                   3.00 $\pm$ 0.17 &                   2.45 $\pm$ 0.19 \\
                    Leaper &          1.72 $\pm$ 0.08 &          1.06 $\pm$ 0.08 & \textbf{0.79 $\pm$ 0.13} &                   1.79 $\pm$ 0.22 &                   1.65 $\pm$ 0.18 \\
                      Maze &          4.76 $\pm$ 0.09 & \textbf{4.27 $\pm$ 0.22} &          4.79 $\pm$ 0.15 &                   4.73 $\pm$ 0.09 &                   4.72 $\pm$ 0.04 \\
                     Miner & \textbf{4.36 $\pm$ 0.13} &          4.81 $\pm$ 0.02 & \textbf{4.42 $\pm$ 0.27} &          \textbf{4.53 $\pm$ 0.22} &          \textbf{4.29 $\pm$ 0.17} \\
                 Dodgeball &          3.88 $\pm$ 0.37 &          2.51 $\pm$ 0.32 & \textbf{0.89 $\pm$ 0.34} &                   4.05 $\pm$ 0.19 &                   3.54 $\pm$ 0.41 \\
                 Starpilot & \textbf{1.10 $\pm$ 0.06} &          2.07 $\pm$ 0.14 &          1.38 $\pm$ 0.11 &                   1.44 $\pm$ 0.13 &          \textbf{1.20 $\pm$ 0.08} \\
                    Chaser & \textbf{1.37 $\pm$ 0.12} &          3.18 $\pm$ 0.26 & \textbf{1.36 $\pm$ 0.14} &                   1.98 $\pm$ 0.34 &          \textbf{1.43 $\pm$ 0.25} \\
                 Bossfight & \textbf{1.33 $\pm$ 0.12} &          4.19 $\pm$ 0.35 & \textbf{1.17 $\pm$ 0.12} &          \textbf{1.15 $\pm$ 0.17} &          \textbf{1.16 $\pm$ 0.33} \\
\midrule
Average Mutual Information &          2.61 $\pm$ 0.05 &          3.31 $\pm$ 0.03 & \textbf{2.00 $\pm$ 0.07} &                   2.84 $\pm$ 0.08 &                   2.60 $\pm$ 0.02 \\
\bottomrule
\end{tabular}
\end{sc}
\end{small}
\end{center}
\vskip -0.1in
\end{table}

\begin{table}[!htb]
\caption{Level classifier accuracies measured under different adaptive sampling strategies. We report aggregated accuracies across 5 training runs, each initialised with a different seed. To compute this quantity for each run and environment, we first fit a linear classifier $p_\vtheta$ to predict level identities from the agent's penultimate layer outputs, using rollouts from levels sampled uniformly from $L$. We then measure the classifier accuracy, using a different set of rollouts also sampled uniformly from $L$. In the last row we compute the mean accuracy across environments for each run, and we report the mean and standard deviation of that quantity across all runs. Bolded methods are not significantly different from the method with lowest mean ($p<0.05$), unless all are, in which case none are bolded.}
\label{tb:procgen_accuracy}
\vskip 0.15in
\begin{center}
\begin{small}
\begin{sc}
\begin{tabular}{lrrrrr}
\toprule
                Environment &                  $S=S^V$ &       $P_S=\mathcal{U}$ &      $S=S^{\mathrm{MI}}$ & $S=S^V, P_{S^\prime}=\mathcal{U}$ & $S=S^V, S^\prime=S^{\mathrm{MI}}$ \\
\midrule
                    Bigfish & \textbf{17.2 $\pm$ 10.1} &          66.5 $\pm$ 5.5 &  \textbf{12.8 $\pm$ 4.5} &          \textbf{24.7 $\pm$ 10.3} &           \textbf{14.3 $\pm$ 9.9} \\
                      Heist &           83.1 $\pm$ 9.5 &          77.8 $\pm$ 3.8 & \textbf{38.1 $\pm$ 20.7} &                    77.5 $\pm$ 5.6 &                    72.5 $\pm$ 9.1 \\
                    Climber &          53.4 $\pm$ 15.0 &          82.9 $\pm$ 3.3 & \textbf{17.4 $\pm$ 10.5} &                    73.2 $\pm$ 3.6 &                    64.0 $\pm$ 5.6 \\
                  Caveflyer &           34.9 $\pm$ 5.4 &          62.5 $\pm$ 3.9 &  \textbf{21.4 $\pm$ 1.5} &                    48.5 $\pm$ 6.0 &          \textbf{33.1 $\pm$ 11.5} \\
                     Jumper &          62.0 $\pm$ 10.4 &          64.4 $\pm$ 6.3 & \textbf{28.0 $\pm$ 17.8} &                    64.1 $\pm$ 5.6 &                   59.4 $\pm$ 10.0 \\
                   Fruitbot &   \textbf{2.7 $\pm$ 4.1} &          18.0 $\pm$ 1.3 &   \textbf{5.3 $\pm$ 6.9} &            \textbf{2.6 $\pm$ 2.0} &            \textbf{2.3 $\pm$ 2.8} \\
                    Plunder &  \textbf{13.5 $\pm$ 9.8} &          25.3 $\pm$ 8.8 &   \textbf{2.6 $\pm$ 3.4} &                    16.1 $\pm$ 6.8 &           \textbf{15.5 $\pm$ 9.9} \\
                    Coinrun &  \textbf{14.1 $\pm$ 0.8} &          22.3 $\pm$ 1.9 &  \textbf{10.5 $\pm$ 4.7} &           \textbf{11.6 $\pm$ 4.0} &           \textbf{13.4 $\pm$ 4.2} \\
                      Ninja &           22.8 $\pm$ 7.5 &          33.9 $\pm$ 2.4 &   \textbf{5.6 $\pm$ 7.2} &                    27.8 $\pm$ 2.9 &                    18.4 $\pm$ 4.9 \\
                     Leaper &           10.2 $\pm$ 5.6 &          11.6 $\pm$ 2.1 &            8.5 $\pm$ 4.0 &                    13.0 $\pm$ 1.0 &                    12.9 $\pm$ 3.2 \\
                       Maze &           69.7 $\pm$ 2.6 & \textbf{65.5 $\pm$ 1.3} & \textbf{69.2 $\pm$ 10.2} &           \textbf{63.9 $\pm$ 3.8} &           \textbf{67.7 $\pm$ 7.8} \\
                      Miner &           85.5 $\pm$ 1.1 &          92.0 $\pm$ 0.7 &  \textbf{67.7 $\pm$ 8.2} &                    80.4 $\pm$ 4.3 &           \textbf{78.1 $\pm$ 4.5} \\
                  Dodgeball &           63.1 $\pm$ 0.8 &          45.4 $\pm$ 7.3 &   \textbf{5.7 $\pm$ 5.8} &                    68.5 $\pm$ 7.5 &                    53.5 $\pm$ 7.4 \\
                  Starpilot &            8.1 $\pm$ 3.1 &          14.7 $\pm$ 3.3 &   \textbf{3.2 $\pm$ 2.4} &                     9.0 $\pm$ 2.7 &                     6.8 $\pm$ 1.1 \\
                     Chaser &  \textbf{13.0 $\pm$ 3.3} &          40.4 $\pm$ 4.7 &   \textbf{7.8 $\pm$ 4.0} &                   27.7 $\pm$ 12.3 &           \textbf{16.2 $\pm$ 6.1} \\
                  Bossfight &   \textbf{6.9 $\pm$ 6.1} &         60.2 $\pm$ 14.3 &   \textbf{3.1 $\pm$ 4.1} &           \textbf{10.4 $\pm$ 6.2} &           \textbf{10.5 $\pm$ 8.5} \\
\midrule
Average Classifier Accuracy &           35.0 $\pm$ 1.7 &          49.0 $\pm$ 0.9 &  \textbf{19.2 $\pm$ 2.2} &                    38.7 $\pm$ 1.3 &                    33.7 $\pm$ 1.6 \\
\bottomrule
\end{tabular}
\end{sc}
\end{small}
\end{center}
\vskip -0.1in
\end{table}

\begin{figure}[!htb]
\centering
    \begin{subfigure}{0.49\linewidth}
            \includegraphics[width=1\linewidth]{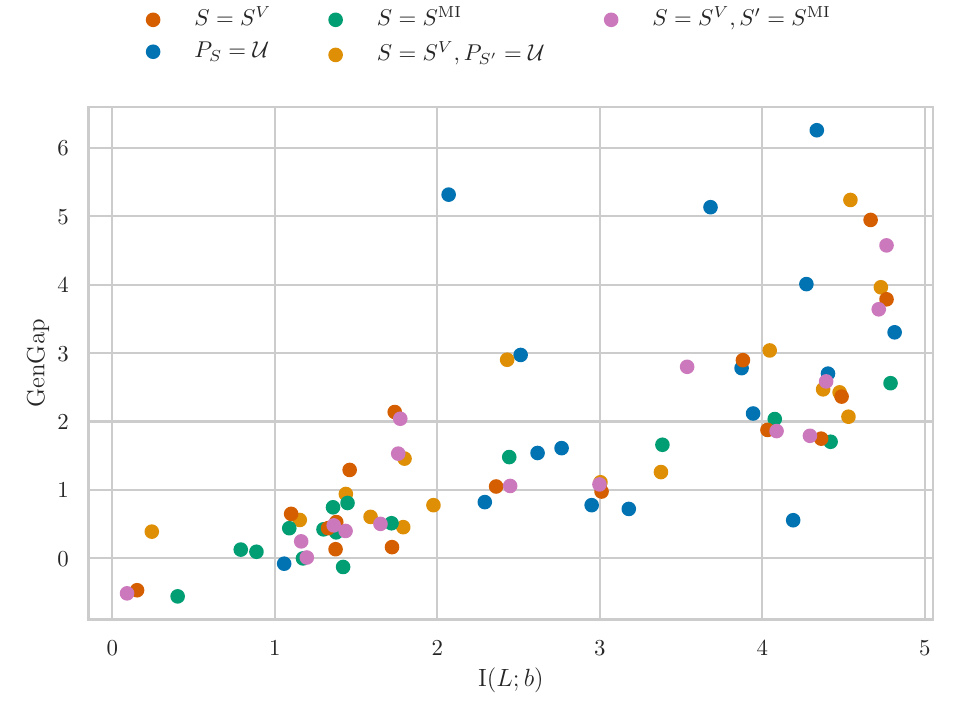}
    \end{subfigure}%
    \begin{subfigure}{0.49\linewidth}
            \includegraphics[width=1\linewidth]{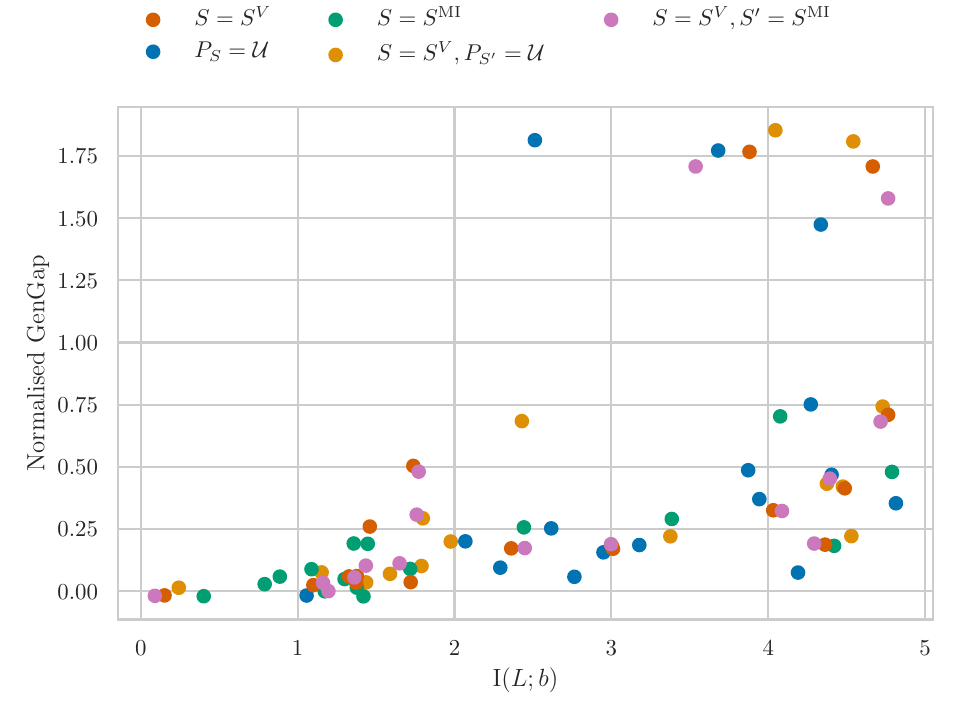}
    \end{subfigure}\\%
    \begin{subfigure}{0.49\linewidth}
            \includegraphics[width=1\linewidth]{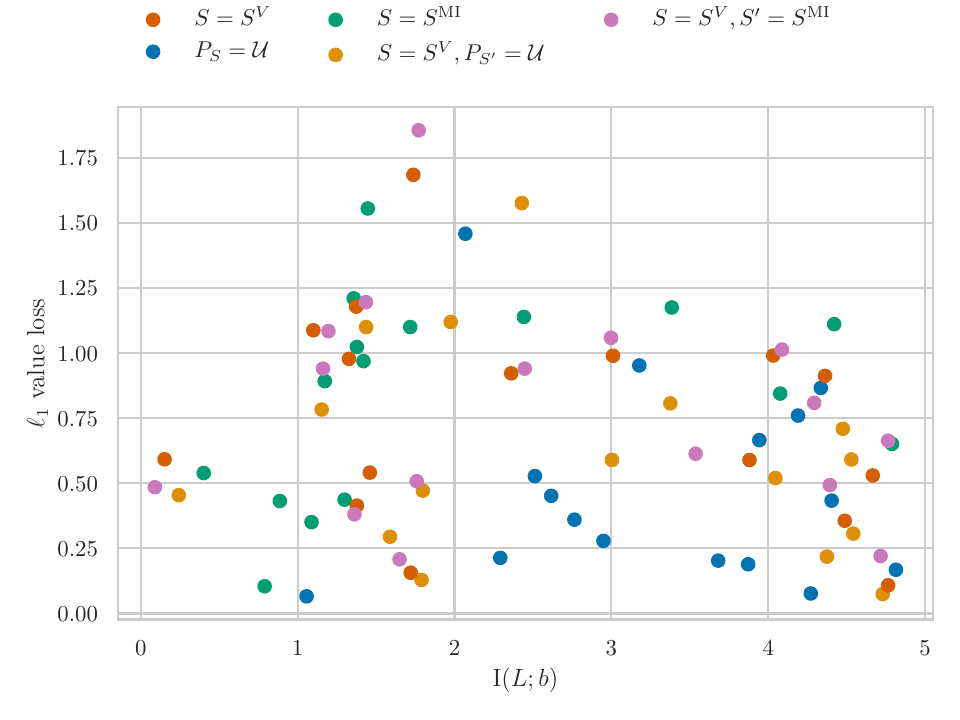}
    \end{subfigure}%
    \caption{Scatter plot displaying how $\mut{L}{b}$ compares to the $\text{GenGap}$ (top left), to the normalised $\text{GenGap}$ (top right), and to the $\ell_1$ average value loss (bottom) across all methods and Procgen games, at the end of training. Each plotted point represents the average of 5 seeds in a particular game.}\label{fig:correlation_analysis}
\end{figure}

To better understand the interaction between the mutual information, the value loss and the generalisation gap, we plot our estimate for $\mut{L}{b}$ at the end of training against the $\text{GenGap}$ and the $\ell_1$-value loss for all methods tested and across Procgen games in \Cref{fig:correlation_analysis}. We find a positive correlation (correlation coefficient $\rho=0.60$) and rank correlation (Kendall rank correlation coefficient $\xi=0.50, p<1e-50$) between $\mut{L}{b}$ and the $\text{GenGap}$. We find similar correlation ($\rho=0.50$) and rank correlation ($\xi=0.49, p<1e-47$) between $\mut{L}{b}$ and the normalised $\text{GenGap}$. We also observe a weaker but statistically significant negative correlation ($\rho=-0.18$) and negative rank correlation ($\xi=-0.11, p<0.001$) between $\mut{L}{b}$ and the $\ell_1$-value loss. 

We report the $\mut{L}{b}$ averaged across seeds for all games and methods tested in \Cref{tb:procgen_mi}. In order to obtain a more intuitive quantification of $\mut{L}{b}$, we also report the classification accuracy of the linear classifier $p_\vtheta$ in \Cref{tb:procgen_accuracy}, as these two quantities are proportional to one-another. Out of 200 training levels, the classifier correctly predicts the current level $49\%$ of the times under uniform sampling, $35\%$ under $(S=S^V)$ and $19\%$ under $S^{\text{MI}}$. While adaptive sampling strategies are able to significantly reduce $\mut{L}{b}$, the mean classifier accuracy is still $70$ times random guessing for $(S=S^V)$ and $38$ times random guessing for $(S=S^{\text{MI}})$.

\subsection{A qualitative analysis of when adaptive sampling strategies may or may not be effective in reducing the $\text{GenGap}$}\label{app:procgen_level_analysis}

\begin{figure}[!htb]
\centering
    \begin{subfigure}{0.49\linewidth}
            \includegraphics[width=1\linewidth]{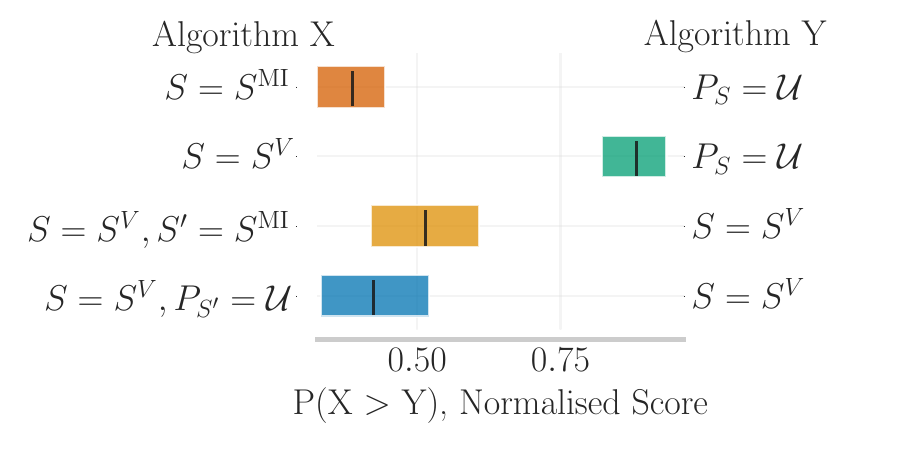}
    \end{subfigure}%
    \begin{subfigure}{0.49\linewidth}
            \includegraphics[width=1\linewidth]{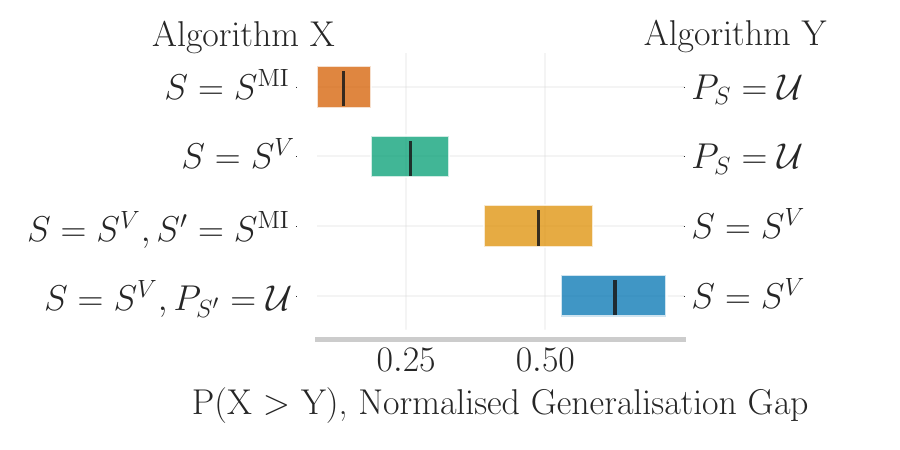}
    \end{subfigure}%
    \caption{Probability of algorithm $X$ incurring a higher normalised test score (left) and $\text{GenGap}$ (right) than algorithm $Y$. Colored boxes indicate the 95\% confidence interval.}\label{fig:p_improvement_procgen}
\end{figure}

\begin{figure}[!htb]
\centering
    \begin{subfigure}{0.3\linewidth}
            \includegraphics[width=1\linewidth]{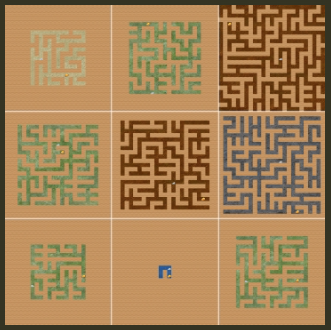}
    \end{subfigure}%
    \begin{subfigure}{0.3\linewidth}
        \includegraphics[width=1\linewidth]{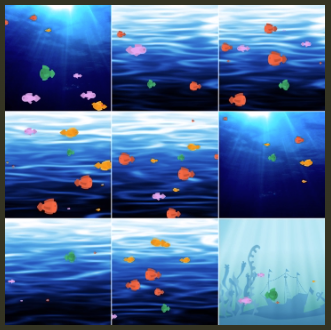}
    \end{subfigure}%
    \caption{Agent observations sampled from 9 levels from the Maze (left) and Bigfish (right) games of the Procgen benchmark.}\label{fig:procgen_levels}
\end{figure}

We report in \Cref{fig:p_improvement_procgen} that, when compared to uniform sampling across procgen games, adaptive sampling strategies are significantly more likely to reduce the $\text{GenGap}$. Strategies employing ($S=S^V$) as their primary scoring function are also more likely to improve their test set scores. However, we measure significant variability across procgen games  for $\mut{L}{b}$ in \Cref{tb:procgen_mi} (and by extension for the $\text{GenGap}$) for the different strategies tested. To better understand why, we compare the measured accuracy with a qualitative analysis of the observations and levels encountered in the ``Maze'' and ``Bigfish'' games (see \Cref{fig:procgen_levels} for renderings of different levels from each game). In Maze, the classifier accuracy remains over $60\%$ ($120\times$ random) for all methods tested and the reduction in $\text{GenGap}$ is insignificant. On the other hand, in Bigfish all adaptive sampling strategies tested lead to a significant reduction in classifier accuracy when compared to uniform sampling, dropping from $65\%$ to between $12\%$ and $25\%$ (depending on the strategy), and correspond to a significant drop in the $\text{GenGap}$ and improvement in test scores. 

In Maze, the observation space lets the agent observe the full layout at every timestep. The maze layout is unique to each level and should be easily identifiable by the agent's ResNet architecture. Intuitively, we can hypothesise that adaptive sampling strategies will not be effective if all the levels are easily identifiable by the model, which appears to be the case in Maze. In these cases, other data regularisation techniques, such as augmenting the observations, could be more effective, and in fact \cite{PLR} report that Maze is one of the games where combining PLR with UCB-DrAC \cite{raileanu2021UCB-DrAC}, a data augmentation method, leads to a significant improvement in test scores.

On the other hand, we observe that many of the Bigfish levels yield similar observations. Indeed, both the features relevant to the task (the fish) and irrelevant (the background) are similar in many of the training levels. Furthermore, there's significant variation in the observations encountered during an episode, as fish constantly appear and leave the screen. Yet, some levels (top left, middle and bottom right) are easily identifiable thanks to their background, and we can expect them to be more prone to overfitting. By using the background to identify the current level the agent is able to infer when and where fish will appear on screen. This is an effective strategy to solve the training level in question and to accurately predict its value function, however applying this strategy to unseen levels at test time will fail, as the background and the fish locations are not correlated across levels. Adaptive sampling strategies minimising $\mut{L}{b}$ de-prioritise problematic levels when the agent's representation starts overfitting, essentially performing data regularisation via a form of rejection sampling.

\subsection{Procgen extended results}\label{app:procgen_extended_results}

\begin{table}[H]
\caption{Test scores of a PPO agent trained under different adaptive sampling strategies. We report aggregated scores across 5 training runs, each initialised with a different seed. For each run the test score is obtained by evaluating the final policy’s average score on 1000 episodes, each episode sampling a different level not in the train set. Following \cite{raileanu2021UCB-DrAC}, normalised test scores per run are computed by dividing its test score per run for each environment by the corresponding average test score of the uniform-sampling strategy over all runs. In the last row we compute the mean normalised score across environments for each run, and we report the mean and standard deviation of that quantity across all runs. Bolded methods are not significantly different from the method with highest mean ($p<0.05$), unless all are, in which case none are bolded.}
\label{tb:procgen_test_scores}
\vskip 0.15in
\begin{center}
\begin{small}
\begin{sc}
\begin{tabular}{lrrrrr}
\toprule
                Environment &                  $S=S^V$ &       $P_S=\mathcal{U}$ &    $S=S^{\mathrm{MI}}$ & $S=S^V, P_{S^\prime}=\mathcal{U}$ & $S=S^V, S^\prime=S^{\mathrm{MI}}$ \\
\midrule
                    Bigfish &  \textbf{11.5 $\pm$ 2.1} &           4.2 $\pm$ 1.2 &          7.3 $\pm$ 0.6 &           \textbf{10.2 $\pm$ 2.0} &           \textbf{12.4 $\pm$ 3.4} \\
                      Heist &            3.2 $\pm$ 0.7 &           2.9 $\pm$ 0.2 &          2.7 $\pm$ 0.5 &                     2.9 $\pm$ 0.8 &                     3.0 $\pm$ 0.4 \\
                    Climber &   \textbf{6.9 $\pm$ 0.3} &           5.8 $\pm$ 0.3 &          4.9 $\pm$ 0.6 &            \textbf{7.0 $\pm$ 0.4} &            \textbf{7.0 $\pm$ 0.2} \\
                  Caveflyer &   \textbf{6.3 $\pm$ 0.1} &           5.7 $\pm$ 0.1 &          5.8 $\pm$ 0.4 &            \textbf{6.5 $\pm$ 0.3} &            \textbf{6.2 $\pm$ 0.2} \\
                     Jumper &   \textbf{6.0 $\pm$ 0.1} &           5.7 $\pm$ 0.2 &          5.8 $\pm$ 0.1 &            \textbf{6.0 $\pm$ 0.0} &            \textbf{5.9 $\pm$ 0.1} \\
                   Fruitbot &  \textbf{28.4 $\pm$ 0.4} & \textbf{27.7 $\pm$ 0.8} &         23.6 $\pm$ 1.4 &           \textbf{28.2 $\pm$ 0.5} &           \textbf{28.6 $\pm$ 0.8} \\
                    Plunder &   \textbf{7.3 $\pm$ 1.0} &           5.0 $\pm$ 0.3 &          5.6 $\pm$ 1.5 &            \textbf{7.7 $\pm$ 1.4} &            \textbf{8.5 $\pm$ 0.8} \\
                    Coinrun &   \textbf{8.9 $\pm$ 0.2} &           8.7 $\pm$ 0.2 & \textbf{9.1 $\pm$ 0.1} &            \textbf{9.0 $\pm$ 0.1} &            \textbf{9.1 $\pm$ 0.1} \\
                      Ninja &   \textbf{7.1 $\pm$ 0.4} &           6.1 $\pm$ 0.2 &          5.0 $\pm$ 0.6 &            \textbf{7.1 $\pm$ 0.2} &            \textbf{7.2 $\pm$ 0.3} \\
                     Leaper &   \textbf{7.1 $\pm$ 0.4} &           4.5 $\pm$ 0.1 &          2.7 $\pm$ 0.1 &            \textbf{6.8 $\pm$ 1.5} &            \textbf{7.2 $\pm$ 1.8} \\
                       Maze &   \textbf{5.6 $\pm$ 0.4} &  \textbf{5.3 $\pm$ 0.3} &          5.0 $\pm$ 0.1 &            \textbf{5.4 $\pm$ 0.4} &            \textbf{5.6 $\pm$ 0.5} \\
                      Miner &   \textbf{9.6 $\pm$ 0.3} &  \textbf{9.3 $\pm$ 0.3} &          8.6 $\pm$ 0.5 &            \textbf{9.6 $\pm$ 0.3} &            \textbf{9.7 $\pm$ 0.3} \\
                  Dodgeball &   \textbf{2.3 $\pm$ 0.3} &           1.6 $\pm$ 0.2 &          0.9 $\pm$ 0.1 &            \textbf{2.1 $\pm$ 0.3} &            \textbf{2.2 $\pm$ 0.4} \\
                  Starpilot &  \textbf{26.9 $\pm$ 1.3} & \textbf{26.5 $\pm$ 2.0} &         18.9 $\pm$ 0.9 &           \textbf{25.2 $\pm$ 2.3} &           \textbf{26.7 $\pm$ 2.4} \\
                     Chaser &   \textbf{6.6 $\pm$ 0.9} &           3.9 $\pm$ 1.1 & \textbf{5.2 $\pm$ 0.8} &            \textbf{6.0 $\pm$ 1.1} &            \textbf{5.9 $\pm$ 1.5} \\
                  Bossfight &   \textbf{9.0 $\pm$ 0.3} &           7.4 $\pm$ 0.3 &          7.8 $\pm$ 0.4 &                     8.4 $\pm$ 0.3 &            \textbf{8.6 $\pm$ 0.5} \\
\midrule
Normalised Test Scores (\%) & \textbf{130.3 $\pm$ 5.3} &         100.0 $\pm$ 1.4 &         97.0 $\pm$ 4.0 &          \textbf{125.4 $\pm$ 5.3} &          \textbf{131.3 $\pm$ 8.4} \\
\bottomrule
\end{tabular}
\end{sc}
\end{small}
\end{center}
\vskip -0.1in
\end{table}

\begin{table}[H]
\caption{Train scores of a PPO agent trained under different adaptive sampling strategies. We report aggregated scores across 5 training runs, each initialised with a different seed. For each run the train score is obtained by evaluating the final policy’s average score on 1000 episodes, each episode sampling a different level from the train set. Following \cite{raileanu2021UCB-DrAC}, normalised train scores per run are computed by dividing its train score per run for each environment by the corresponding average test score of the uniform-sampling strategy over all runs. In the last row we compute the mean normalised score across environments for each run, and we report the mean and standard deviation of that quantity across all runs. Bolded methods are not significantly different from the method with highest mean ($p<0.05$), unless all are, in which case none are bolded.}
\label{tb:procgen_train_scores}
\vskip 0.15in
\begin{center}
\begin{small}
\begin{sc}
\begin{tabular}{lrrrrr}
\toprule
                 Environment &                  $S=S^V$ &       $P_S=\mathcal{U}$ &    $S=S^{\mathrm{MI}}$ & $S=S^V, P_{S^\prime}=\mathcal{U}$ & $S=S^V, S^\prime=S^{\mathrm{MI}}$ \\
\midrule
                     Bigfish &  \textbf{13.6 $\pm$ 0.9} & \textbf{10.5 $\pm$ 1.8} &          8.2 $\pm$ 1.3 &           \textbf{13.1 $\pm$ 1.4} &           \textbf{14.4 $\pm$ 3.0} \\
                       Heist &   \textbf{8.2 $\pm$ 0.5} &  \textbf{8.0 $\pm$ 0.6} &          4.7 $\pm$ 0.7 &            \textbf{8.2 $\pm$ 0.7} &            \textbf{7.6 $\pm$ 0.5} \\
                     Climber &            8.8 $\pm$ 0.2 &           8.5 $\pm$ 0.3 &          6.4 $\pm$ 0.8 &            \textbf{9.4 $\pm$ 0.3} &                     8.9 $\pm$ 0.2 \\
                   Caveflyer &            7.3 $\pm$ 0.2 &  \textbf{7.8 $\pm$ 0.3} &          6.3 $\pm$ 0.1 &            \textbf{7.7 $\pm$ 0.2} &                     7.3 $\pm$ 0.1 \\
                      Jumper &   \textbf{8.4 $\pm$ 0.2} &  \textbf{8.5 $\pm$ 0.1} &          7.4 $\pm$ 0.3 &            \textbf{8.5 $\pm$ 0.1} &            \textbf{8.5 $\pm$ 0.2} \\
                    Fruitbot &           27.9 $\pm$ 0.4 & \textbf{29.3 $\pm$ 0.3} &         23.1 $\pm$ 0.7 &                    28.6 $\pm$ 0.3 &                    28.1 $\pm$ 0.3 \\
                     Plunder &   \textbf{8.6 $\pm$ 1.0} &           5.7 $\pm$ 0.4 &          6.1 $\pm$ 1.3 &            \textbf{9.2 $\pm$ 1.7} &           \textbf{10.1 $\pm$ 0.9} \\
                     Coinrun &            9.5 $\pm$ 0.1 &           9.5 $\pm$ 0.1 &          9.5 $\pm$ 0.1 &                     9.6 $\pm$ 0.0 &                     9.6 $\pm$ 0.1 \\
                       Ninja &   \textbf{8.1 $\pm$ 0.2} &           7.6 $\pm$ 0.2 &          4.9 $\pm$ 0.3 &            \textbf{8.2 $\pm$ 0.1} &            \textbf{8.3 $\pm$ 0.2} \\
                      Leaper &   \textbf{7.2 $\pm$ 0.1} &           4.4 $\pm$ 0.1 &          2.8 $\pm$ 0.1 &            \textbf{7.3 $\pm$ 1.6} &            \textbf{7.7 $\pm$ 1.9} \\
                        Maze &   \textbf{9.4 $\pm$ 0.1} &  \textbf{9.3 $\pm$ 0.1} &          7.5 $\pm$ 0.3 &            \textbf{9.4 $\pm$ 0.1} &            \textbf{9.3 $\pm$ 0.1} \\
                       Miner &           11.3 $\pm$ 0.3 & \textbf{12.6 $\pm$ 0.1} &         10.3 $\pm$ 0.4 &                    11.6 $\pm$ 0.2 &                    11.5 $\pm$ 0.2 \\
                   Dodgeball &   \textbf{5.2 $\pm$ 0.4} &  \textbf{4.6 $\pm$ 0.6} &          1.0 $\pm$ 0.3 &            \textbf{5.2 $\pm$ 0.6} &            \textbf{5.0 $\pm$ 0.6} \\
                   Starpilot &           27.6 $\pm$ 1.8 & \textbf{31.8 $\pm$ 1.5} &         19.3 $\pm$ 0.9 &                    26.2 $\pm$ 1.8 &                    26.7 $\pm$ 1.9 \\
                      Chaser &   \textbf{6.8 $\pm$ 0.9} &           4.6 $\pm$ 1.3 & \textbf{6.0 $\pm$ 1.2} &            \textbf{6.8 $\pm$ 1.0} &            \textbf{6.3 $\pm$ 1.1} \\
                   Bossfight &   \textbf{9.4 $\pm$ 0.2} &           8.0 $\pm$ 0.4 &          7.8 $\pm$ 0.1 &                     8.9 $\pm$ 0.3 &                     8.9 $\pm$ 0.3 \\
\midrule
Normalised Train Scores (\%) & \textbf{170.4 $\pm$ 2.0} &         153.1 $\pm$ 2.1 &        113.1 $\pm$ 5.0 &          \textbf{171.4 $\pm$ 5.9} &          \textbf{171.1 $\pm$ 9.4} \\
\bottomrule
\end{tabular}
\end{sc}
\end{small}
\end{center}
\vskip -0.1in
\end{table}

\begin{figure}[H]
            \includegraphics[width=1\linewidth]{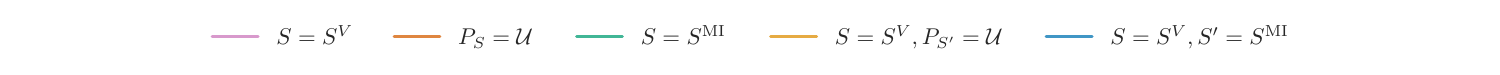}
    \begin{subfigure}{0.33\linewidth}
            \includegraphics[width=1\linewidth]{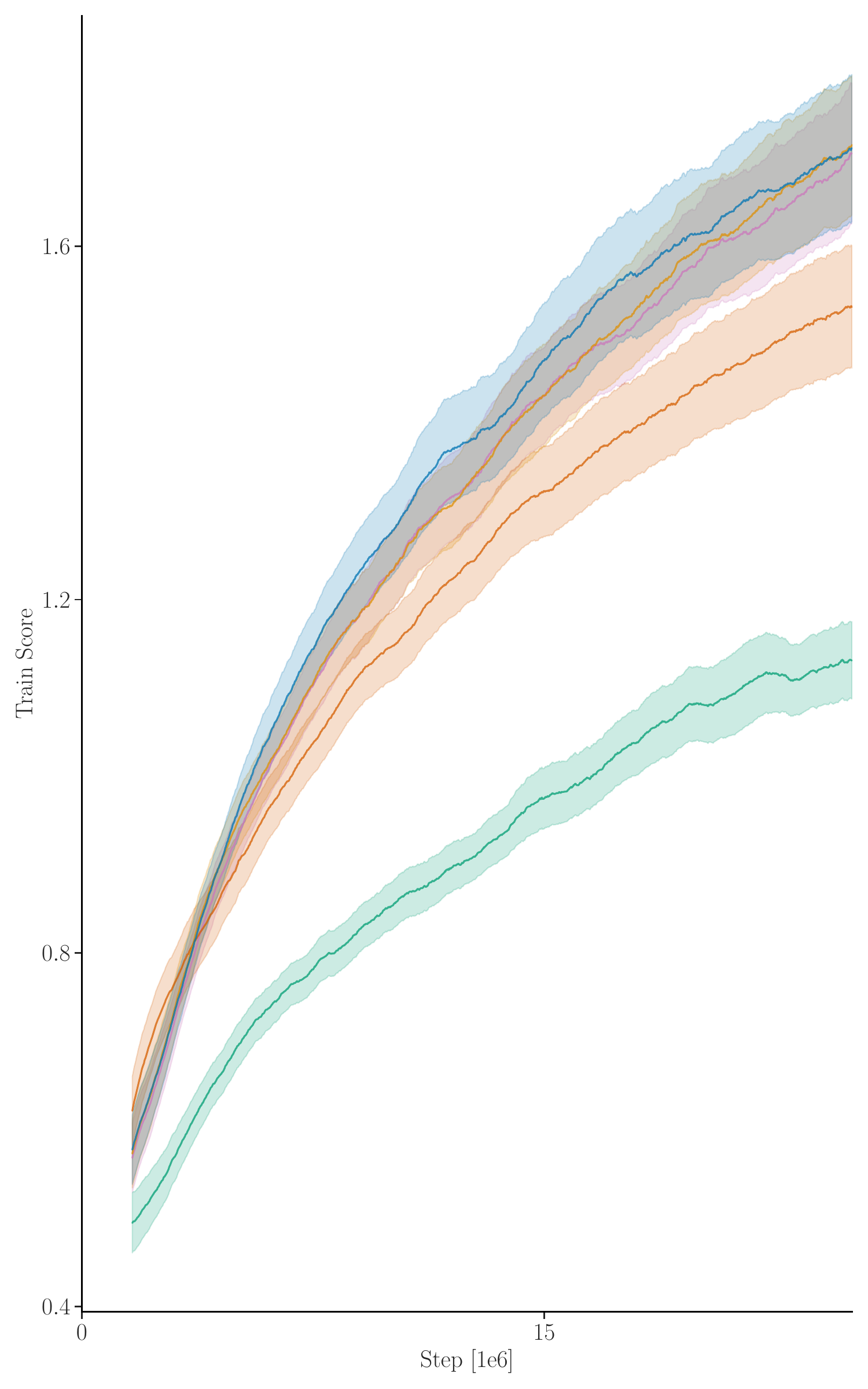}
            \caption{Train Score}\label{subfig:procgen_agg_train}
    \end{subfigure}%
        \begin{subfigure}{0.33\linewidth}
            \includegraphics[width=1\linewidth]{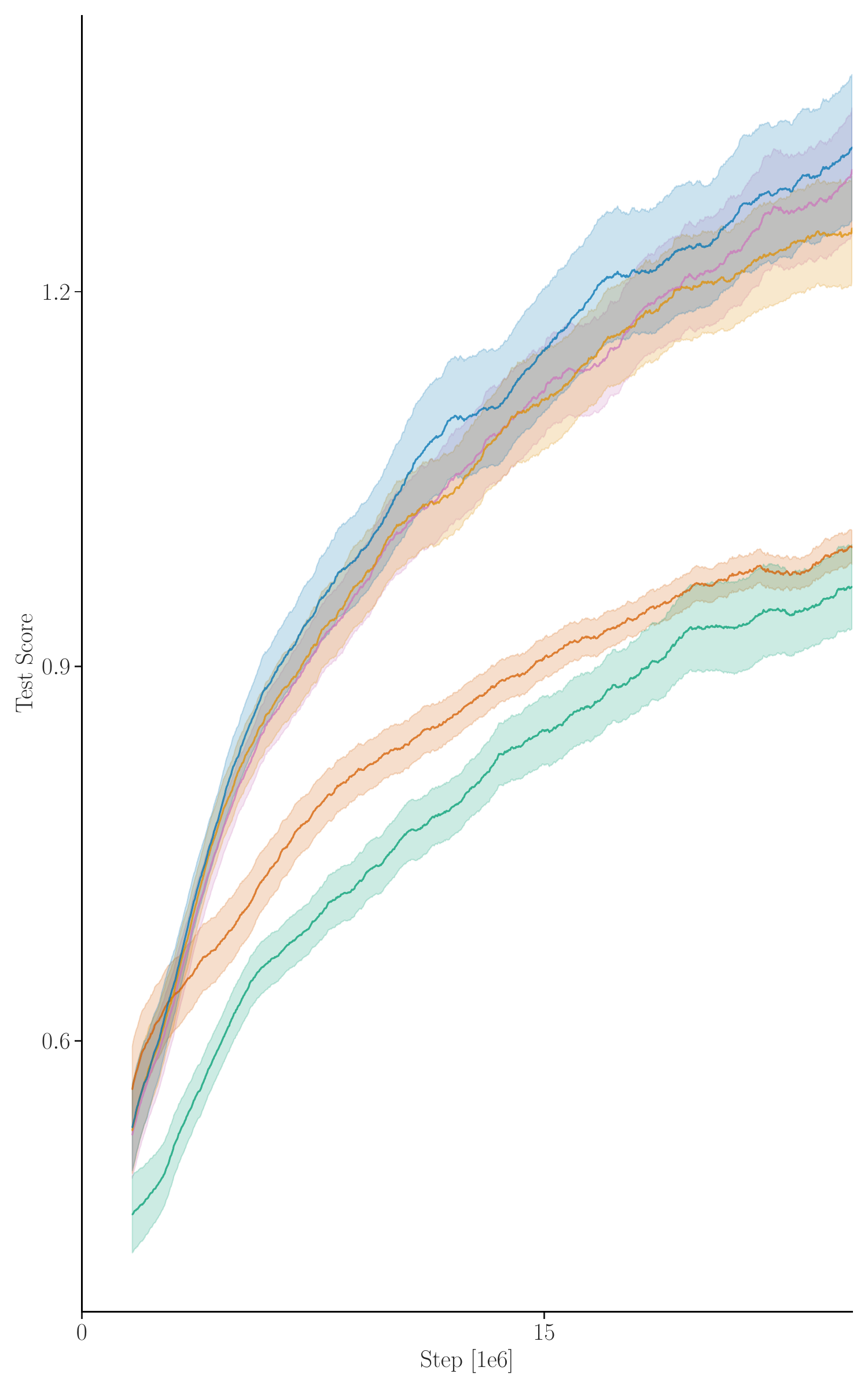}
            \caption{Test Score}\label{subfig:procgen_agg_test}
    \end{subfigure}%
    \begin{subfigure}{0.33\linewidth}
            \includegraphics[width=1\linewidth]{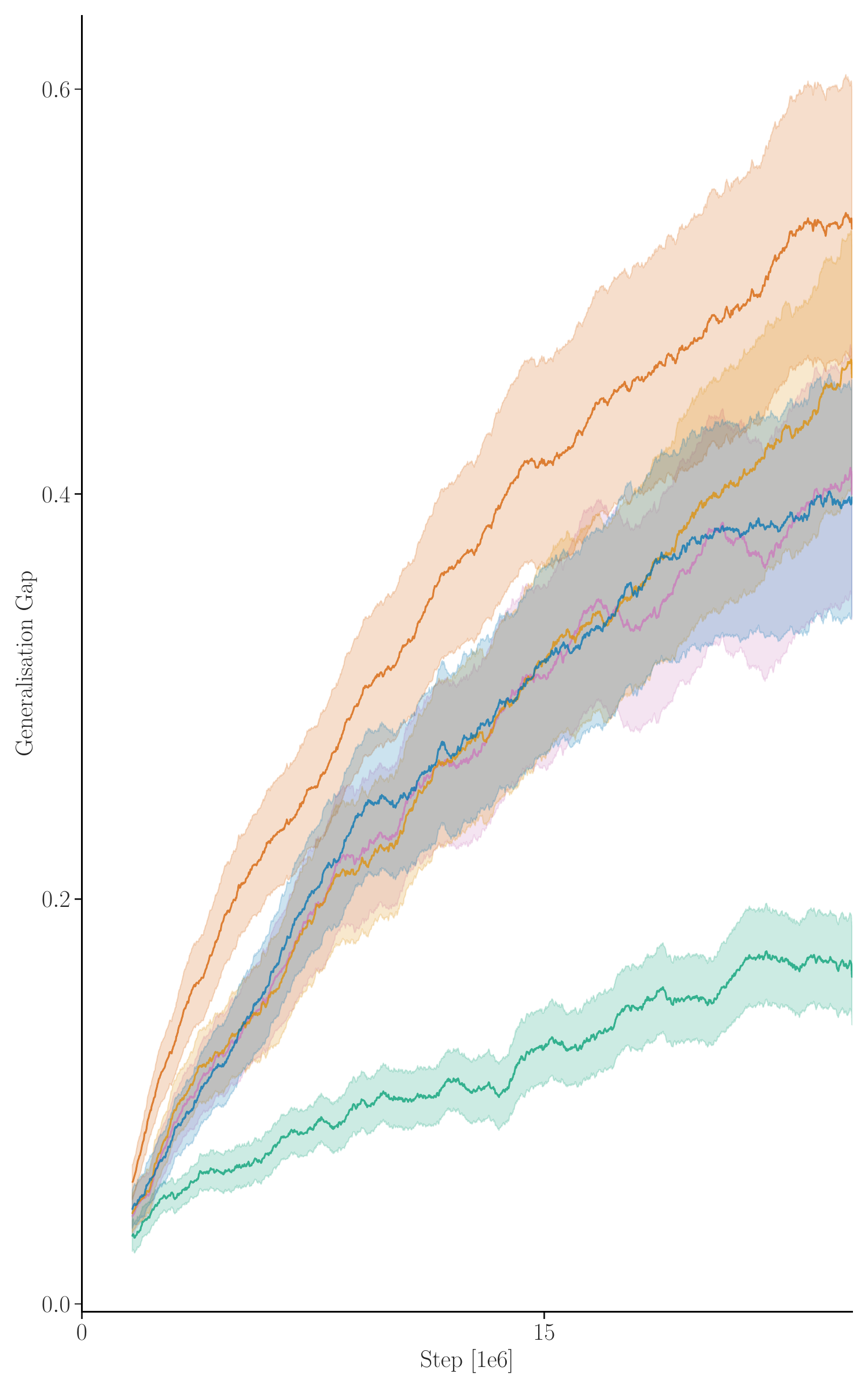}
            \caption{Generalisation Gap}\label{subfig:procgen_agg_gengap}
    \end{subfigure}
    \caption{Procgen normalised scores across environments and generalisation gap over the course of training.
    }
    \label{fig:procgen_agg_train_test_curves}
\end{figure}

\begin{figure}[H]
    \centering
    \includegraphics[width=1\linewidth]{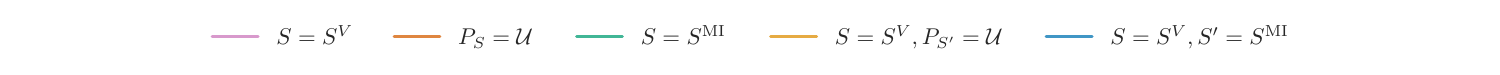}
    \includegraphics[width=1\linewidth]{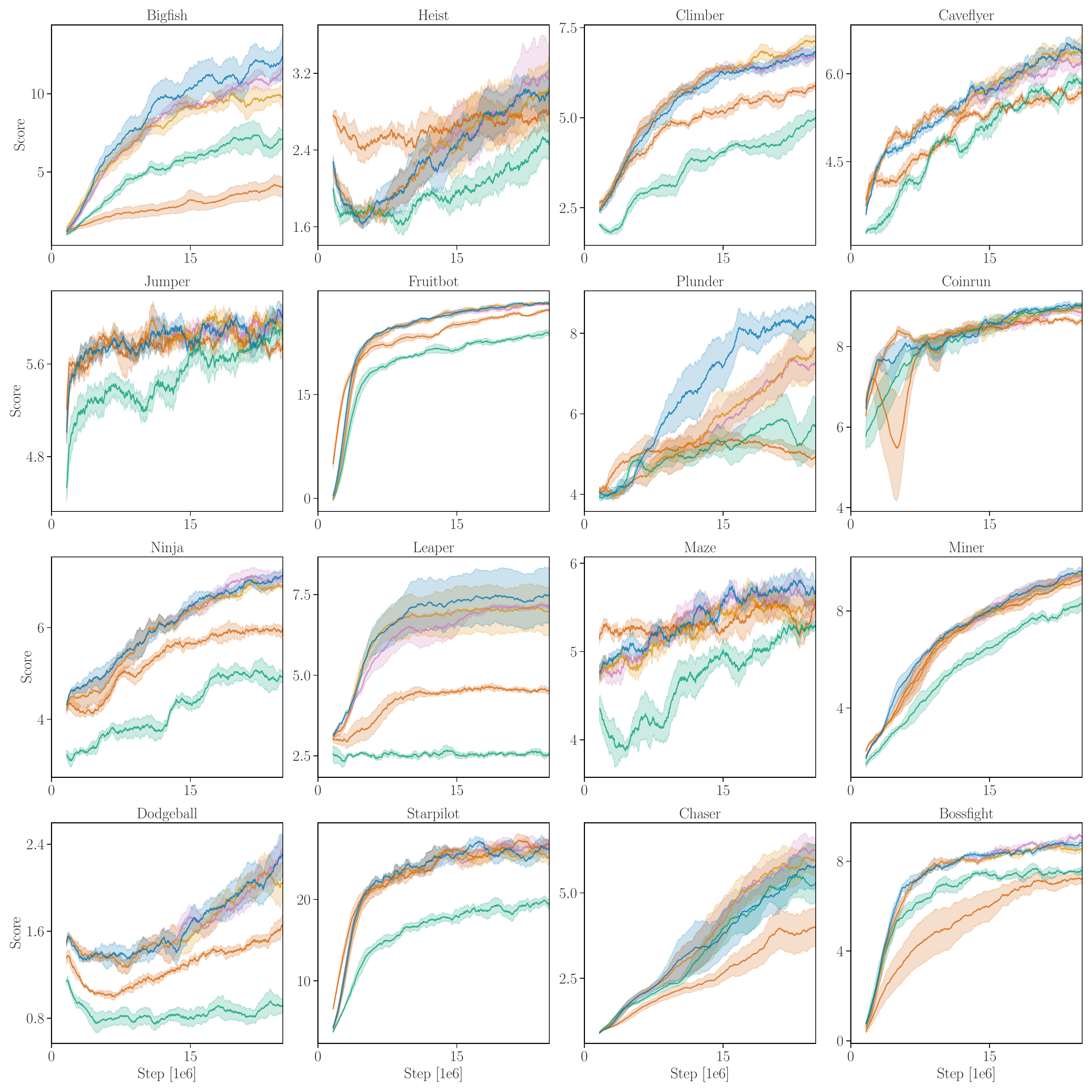}
    \caption{Procgen per game test set scores over the course of training.
    }
    \label{fig:procgen_perenv_test_curves}
\end{figure}

\begin{figure}[H]
    \centering
    \includegraphics[width=1\linewidth]{app_content/procgen/trainset_curves_per_env_legend.pdf}
    \includegraphics[width=1\linewidth]{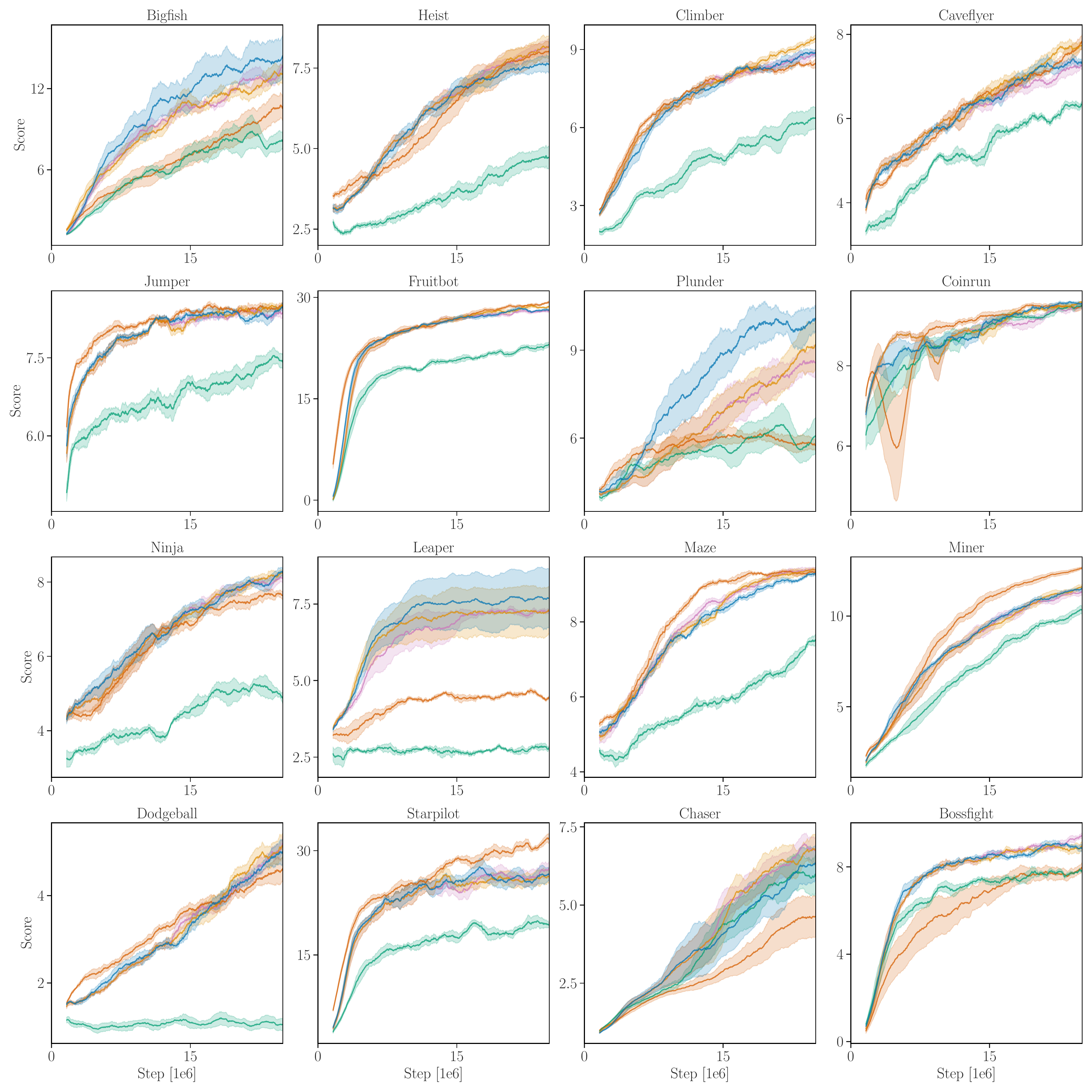}
    \caption{Procgen per game train set scores over the course of training.
    }
    \label{fig:procgen_perenv_train_curves}
\end{figure}

\section{Minigrid additional experimental results}\label{app:results_minigrid}

\subsection{Quantifying the distributional shift in minigrid}\label{app:dist-shift}

\begin{figure}[!htb]
    \centering
    \includegraphics[width=1\linewidth]{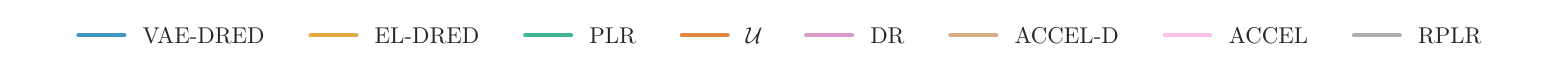}
    \begin{subfigure}{0.49\linewidth}
            \includegraphics[width=1\linewidth]{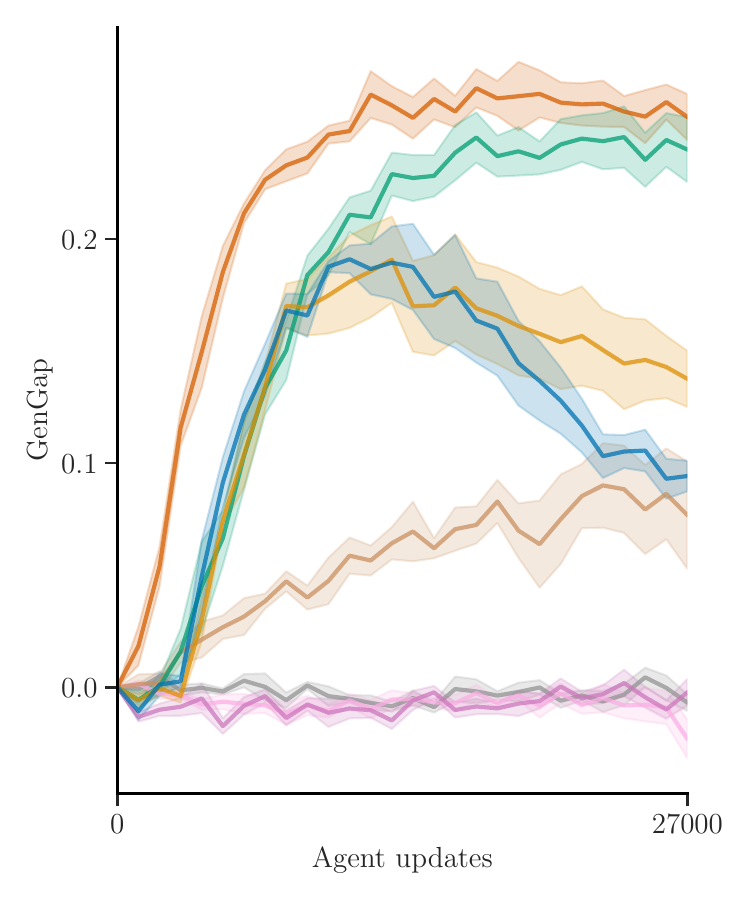}
            \caption{}\label{subfig:gengap}
    \end{subfigure}%
        \begin{subfigure}{0.49\linewidth}
            \includegraphics[width=1\linewidth]{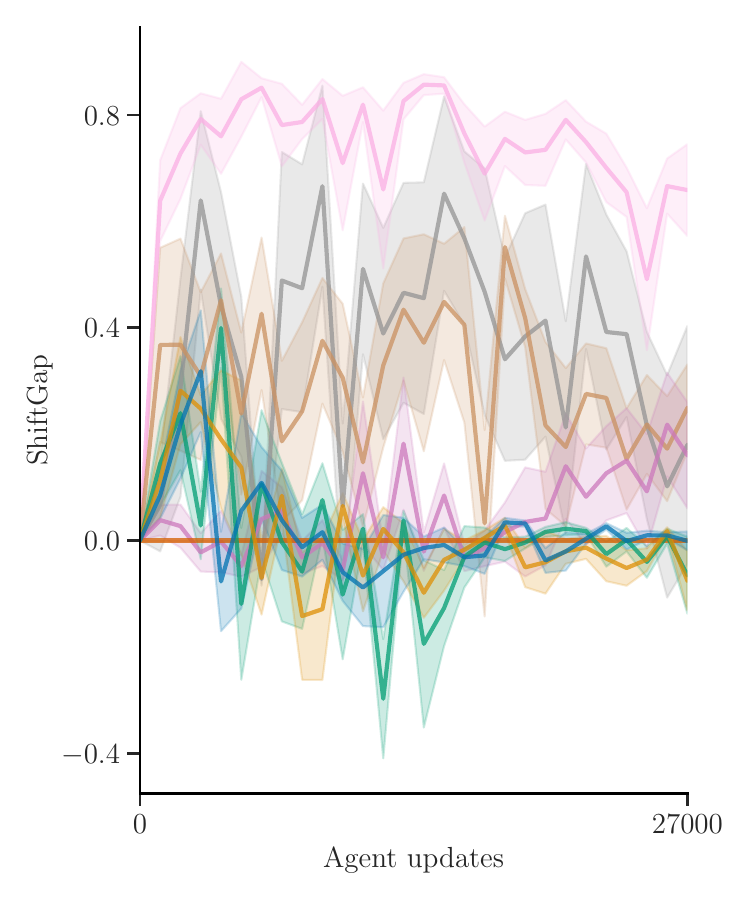}
            \caption{}\label{subfig:overgengap}
    \end{subfigure}%
    \caption{Evolution of (a) the $\text{GenGap}$ (\Cref{eq:GenGap}) and (b) the $\text{ShiftGap}$ (\Cref{eq:ShiftGap}) over the course of training. Fixed set sampling strategies experience higher $\text{GenGap}$ and low $\text{ShiftGap}$, while UED methods follow the opposite trend. Both DRED methods maintain near-zero $\text{ShiftGap}$ while achieving a significantly smaller $\text{GenGap}$ than fixed set strategies. DR spends the majority of training with a low $\text{GenGap}$ and $\text{ShiftGap}$, but a sharp increase towards the end of training brings its $\text{ShiftGap}$ to similar levels as RPLR and ACCEL-D.
    }
    \label{fig:gen_gaps}
\end{figure}

\begin{figure}[H]
            \includegraphics[width=1\linewidth]{app_content/all-legend.pdf}
    \begin{subfigure}{0.49\linewidth}
            \includegraphics[width=1\linewidth]{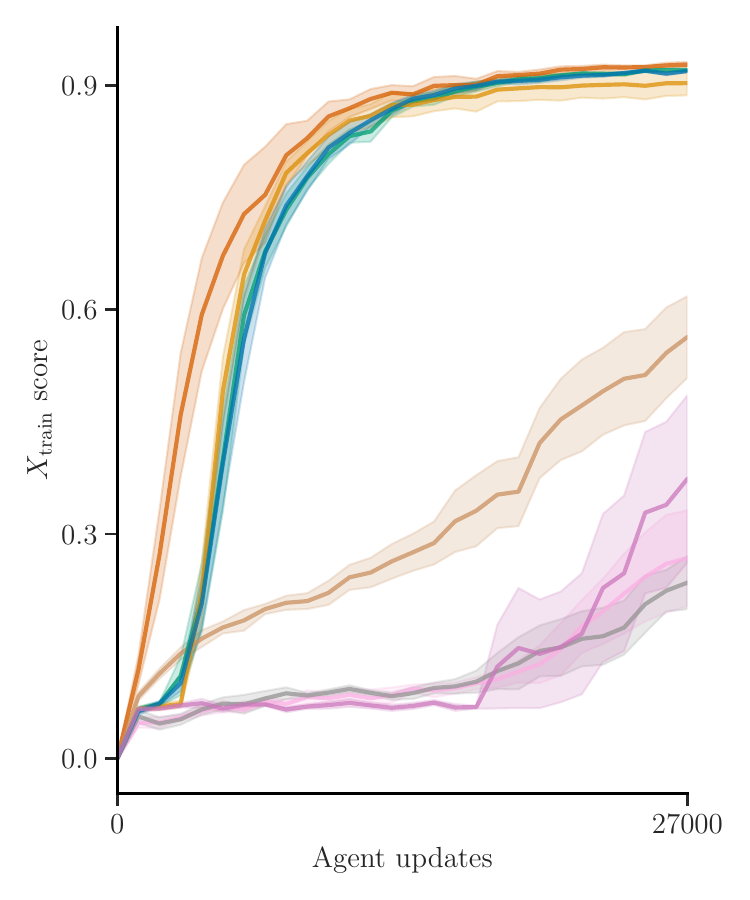}
    \end{subfigure}%
        \begin{subfigure}{0.49\linewidth}
            \includegraphics[width=1\linewidth]{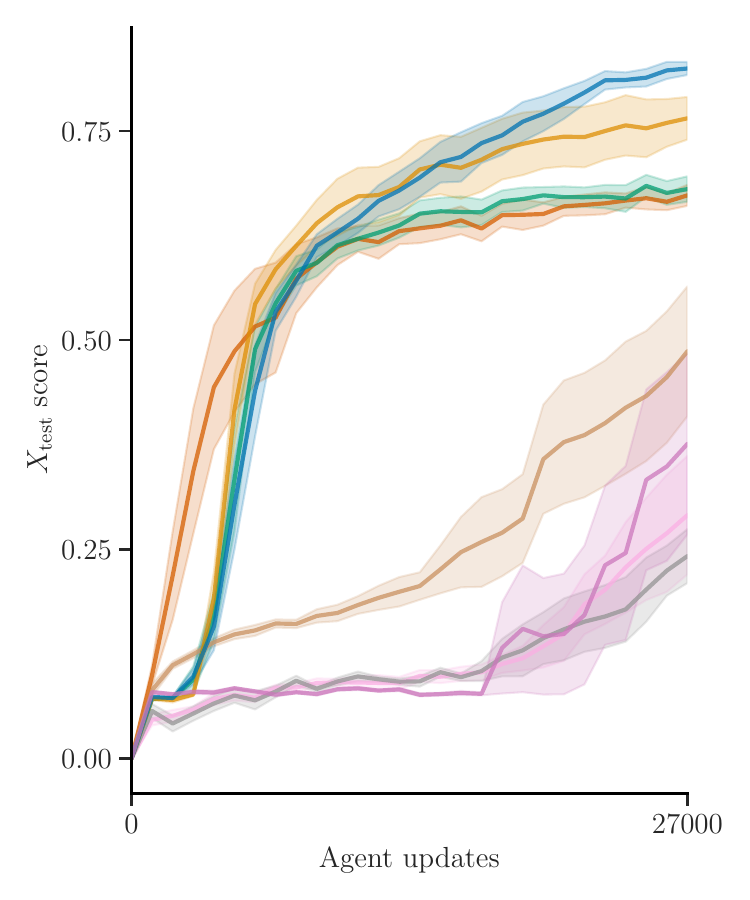}
    \end{subfigure}%
    \caption{Minigrid scores for $X_\text{train}$ and $X_\text{test}$ over the course of training. Shaded area represents the standard error across 5 seeds.
    }
    \label{fig:minigrid_train_curves}
\end{figure}

In \Cref{fig:gen_gaps}, we report the $\text{GenGap}$ and the $\text{ShiftGap}$ in the Minigrid experiments. The $\text{GenGap}$ is not helpful in interpreting the poor test set performance of UED methods. In fact, we report in \Cref{fig:minigrid_train_curves} that UED methods tend to perform well when evaluated on levels sampled from $P_\Lambda$, while performing poorly on the train and test sets. This results in the $\text{GenGap}$ remaining close to zero throughout training. 

We would also expect the $\text{GenGap}$ and the train and test scores to be near-zero for a random policy, that is if the agent was not learning anything at all. However ACCEL and RPLR experience a different failure mode: the UED-trained agents are learning, as demonstrated by their scores on their respective training distributions, but they are learning to solve a different CMDP. In contrast, the $\text{ShiftGap}$ is effective at distinguishing between the agent not learning at all and it learning an out-of-context policy.

While the $\text{ShiftGap}$ allows us to quantify how the distributional shift impacts the agent performance, we desire to quantify the distributional shift directly, and within a feature space consistent with the CMDP semantics. We first compute the distribution $c(t,d|i_{\vx})$, the probability of tile type $t$ occurring at shortest path $d$ from the goal location, for each level $i_{\vx} \in \Lambda$. $c_p$ is the marginal $c_p = \mathbb{E}_{i_{\vx} \sim p(i_{\vx})}[c(t,d|i_{\vx})]$ and we measure the Jensen-Shannon Divergence $\text{JSD}(c_p||c_q)$, with $p=P_\Lambda$ and $q=\mathcal{U}(X_\text{train})$.

\begin{figure}[!htb]
    \centering
    \includegraphics[width=1\linewidth]{app_content/all-legend.pdf}
    \begin{subfigure}{0.49\linewidth}
            \includegraphics[width=1\linewidth]{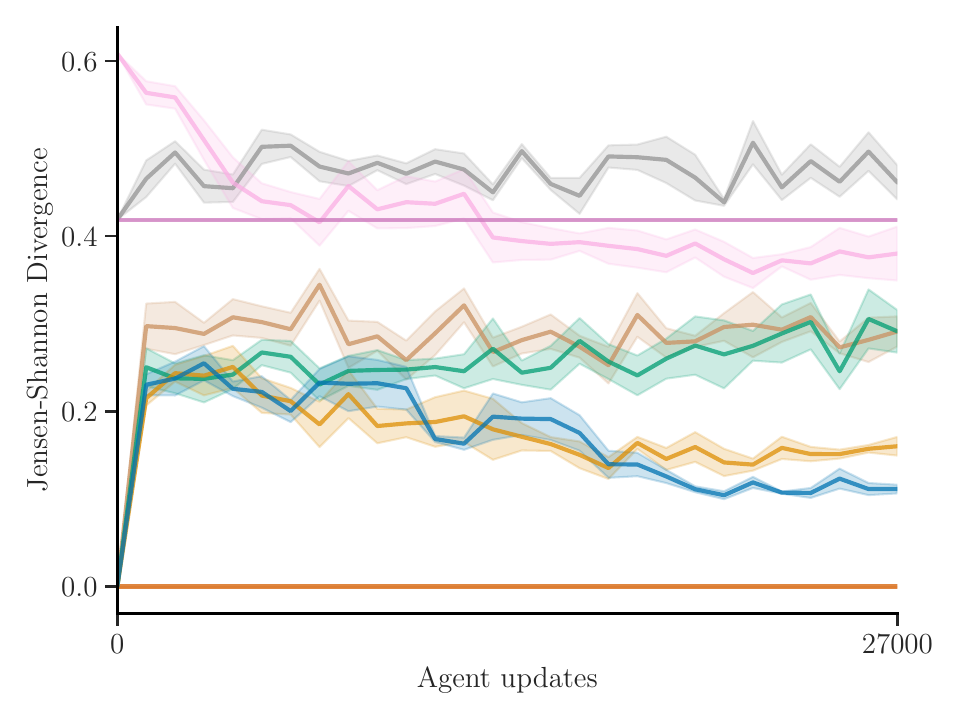}
            \caption{}\label{subfig:jsd}
    \end{subfigure}%
        \begin{subfigure}{0.49\linewidth}
            \includegraphics[width=1\linewidth]{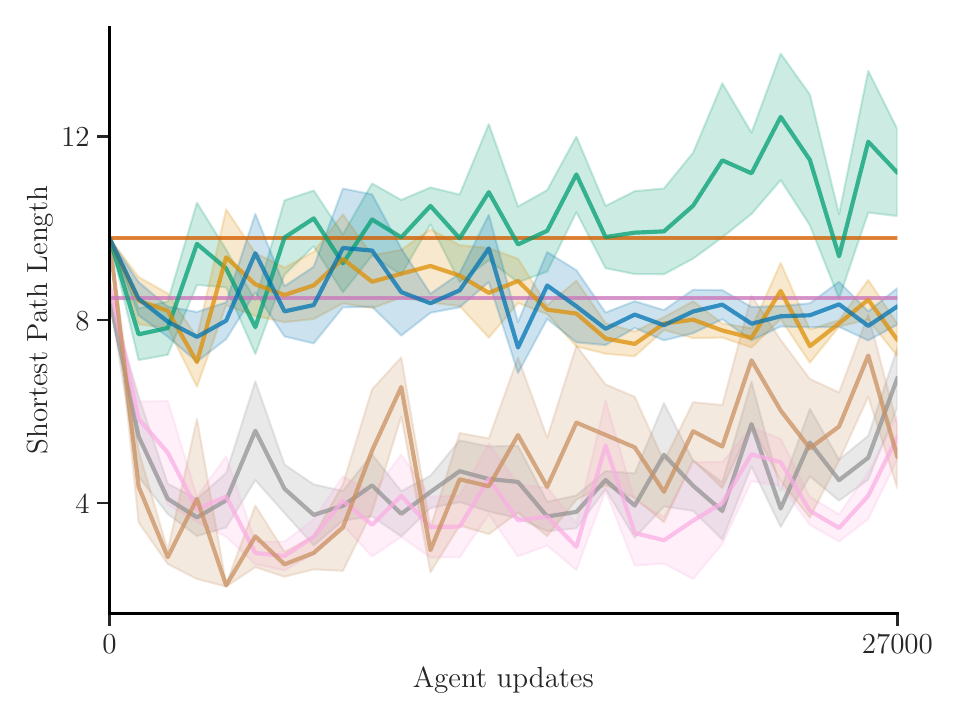}
            \caption{}
    \end{subfigure}\\
        \begin{subfigure}{0.49\linewidth}
            \includegraphics[width=1\linewidth]{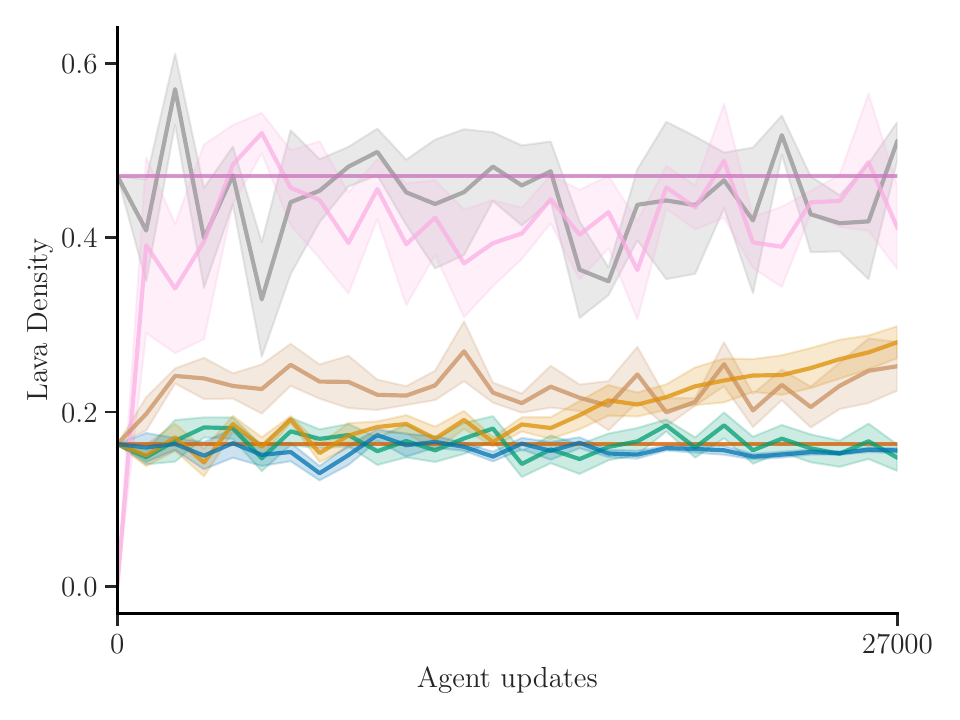}
            \caption{}
    \end{subfigure}%
            \begin{subfigure}{0.49\linewidth}
            \includegraphics[width=1\linewidth]{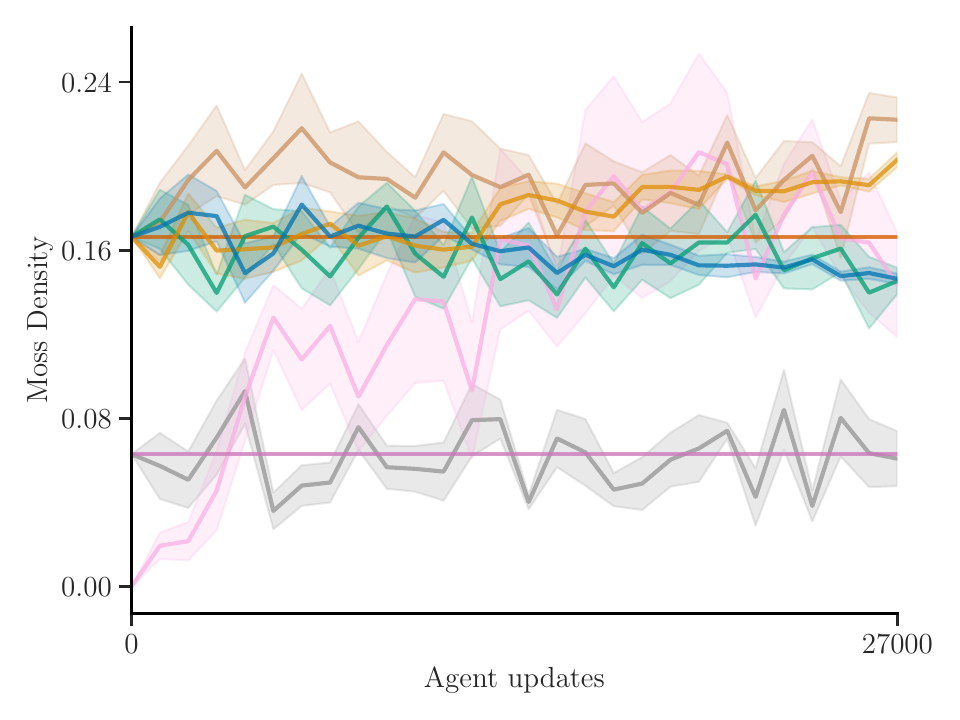}
            \caption{}
    \end{subfigure}%
    \caption{In sampled levels over the course of training, evolution of (a) the Jensen-Shannon divergence, (b) the shortest path length between the start and goal location, (c) the lava tile density (over non-navigable tiles) and (d) the moss tile density (over navigable tiles).}
    \label{fig:buffermetrics}
\end{figure}

We report how the JSD evolves over the course of training for different methods in \Cref{subfig:jsd}. We observe that distributional shift occurs early on during training and remains relatively stable afterwards in all methods. Surprisingly, both DRED methods demonstrate a smaller JSD than PLR, even though PLR only samples $X_\text{train}$ levels. JSD and ShiftGap are positively correlated for all methods except DR and PLR, which both present high JSD but low ShiftGap. VAE-DRED is the only generative method to maintain a low JSD and ShiftGap throughout training. Interestingly, and with the exception of $\mathcal{U}$, methods rank nearly identically to their test set performance ranking when sorted according to their JSD (lowest first). This relationship appears to hold throughout training, as can be observed by comparing the JSD with $X_\text{test}$ scores in \Cref{fig:minigrid_train_curves}.%

In \Cref{fig:buffermetrics}, we report additional metrics on the levels sampled by each method. Over the course of training, only VAE-DRED and PLR stay relatively consistent with $X_\text{train}$ across the three metrics considered.

\subsection{Extended Minigrid results}\label{app:minigrid_extended_results}

\begin{figure}[H]
\centering%
    \begin{subfigure}[]{.08\linewidth}
    \centering
    \includegraphics[width=1\linewidth]{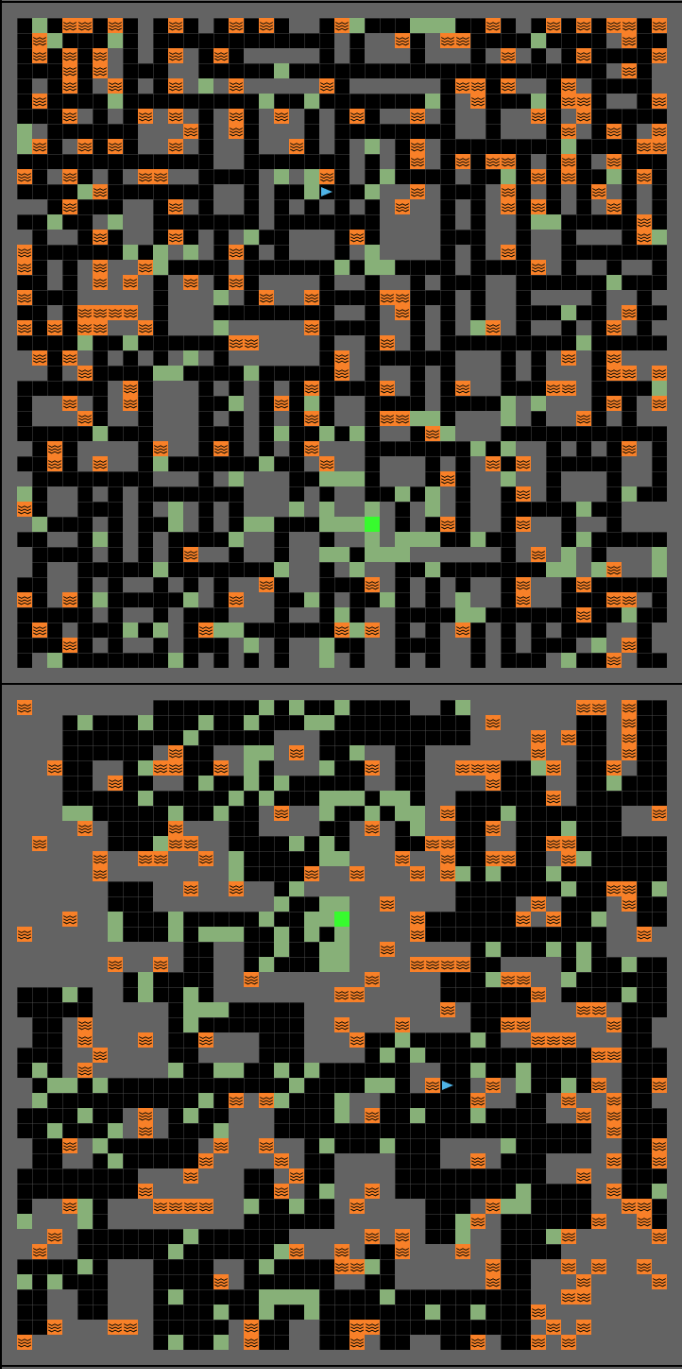}
  \end{subfigure}%
   \begin{subfigure}[]{.72\linewidth}
    \centering
    \includegraphics[width=1\linewidth]{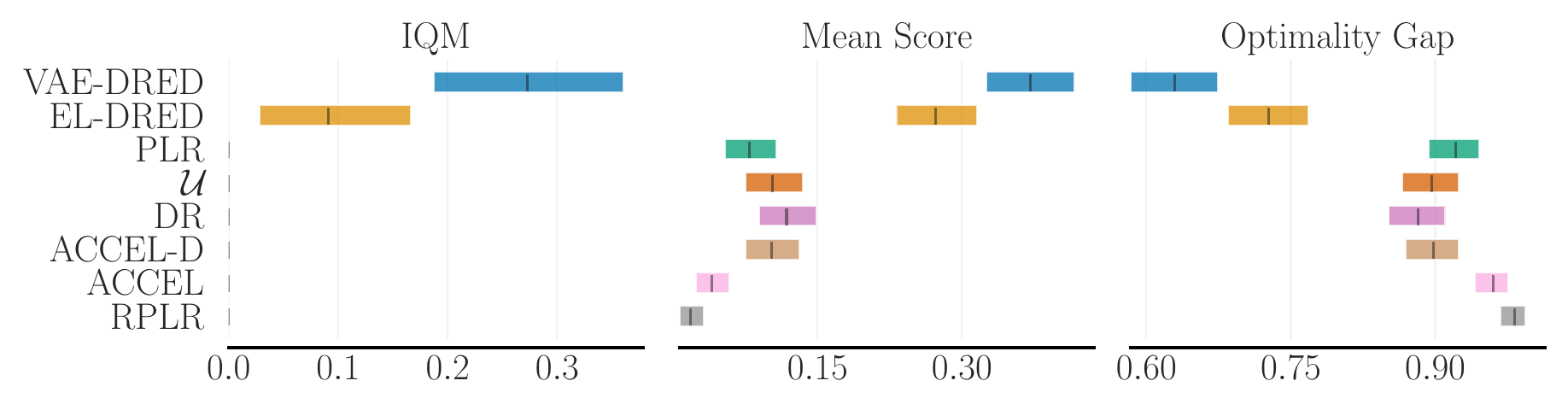}
  \end{subfigure}%
    \begin{subfigure}[]{.2\linewidth}
    \centering
    \includegraphics[width=1\linewidth]{content/solved_rates/solved_rate_Hardcoreall.pdf}
    \end{subfigure}%
\caption{Aggregate final performance and mean solved rate on 216 ``Hardcore'' levels, with a 45x45 layout size, making Hardcore layouts 9 times larger than $X_\text{train}$ layouts. Example layouts from each evaluation set are plotted on the left. The coloured boxes indicate a 99\% confidence interval and the black horizontal bars indicate standard error across 5 training seeds. We refer the reader to \Cref{app:cmdp_levelset} for additional details on our evaluation sets.}
\label{fig:zsg_hard_set_and_edge_cases}
\end{figure}

\begin{table}[H]
\caption{Mean scores achieved in the Minigrid setting for each level evaluation dataset. We report the mean and standard deviation across 5 training runs, each corresponding to a different random seed. For each run the mean score is obtained by evaluating the final policy’s once in each level of the corresponding level dataset, and averaging. Bolded methods are not significantly different from the method with highest mean ($p<0.05$), unless all are, in which case none are bolded.}
\label{tb:minigrid_scores}
\vskip 0.15in
\begin{center}
\begin{small}
\begin{sc}
\tabcolsep=0.11cm
\begin{tabular}{lrrrrrrrr}
\toprule
      Level\\ datasets &                 VAE-DRED &                  EL-DRED &                      PLR &            $\mathcal{U}$ &              DR &         ACCEL-D &           ACCEL &            RPLR \\
\midrule
$X_{\mathrm{train}}$ & \textbf{0.92 $\pm$ 0.00} & \textbf{0.90 $\pm$ 0.03} & \textbf{0.92 $\pm$ 0.00} & \textbf{0.93 $\pm$ 0.01} & 0.37 $\pm$ 0.22 & 0.56 $\pm$ 0.11 & 0.27 $\pm$ 0.13 & 0.24 $\pm$ 0.07 \\
 $X_{\mathrm{test}}$ & \textbf{0.82 $\pm$ 0.02} & \textbf{0.77 $\pm$ 0.05} &          0.68 $\pm$ 0.03 &          0.67 $\pm$ 0.03 & 0.38 $\pm$ 0.22 & 0.49 $\pm$ 0.16 & 0.29 $\pm$ 0.14 & 0.24 $\pm$ 0.06 \\
              Edge C & \textbf{0.74 $\pm$ 0.05} & \textbf{0.67 $\pm$ 0.11} &          0.35 $\pm$ 0.11 &          0.32 $\pm$ 0.04 & 0.37 $\pm$ 0.21 & 0.34 $\pm$ 0.17 & 0.25 $\pm$ 0.14 & 0.17 $\pm$ 0.05 \\
            Hardcore & \textbf{0.37 $\pm$ 0.06} & \textbf{0.27 $\pm$ 0.07} &          0.08 $\pm$ 0.04 &          0.10 $\pm$ 0.04 & 0.12 $\pm$ 0.09 & 0.10 $\pm$ 0.11 & 0.04 $\pm$ 0.04 & 0.02 $\pm$ 0.02 \\
\bottomrule
\end{tabular}
\end{sc}
\end{small}
\end{center}
\vskip -0.1in
\end{table}

\begin{table}[H]
\caption{Solved rates achieved in the Minigrid setting for each level evaluation dataset. We report the mean and standard deviation across 5 training runs, each corresponding to a different random seed. For each run the solved rate is obtained by counting how many levels were solved (i.e. the agent reached the goal) after evaluating the final policy’s once in each level of the corresponding level dataset. Bolded methods are not significantly different from the method with highest mean ($p<0.05$), unless all are, in which case none are bolded.}
\label{tb:minigrid_solved_rate}
\vskip 0.15in
\begin{center}
\begin{small}
\begin{sc}
\tabcolsep=0.11cm
\begin{tabular}{lrrrrrrrr}
\toprule
      Level\\ datasets &                 VAE-DRED &                  EL-DRED &                      PLR &            $\mathcal{U}$ &              DR &         ACCEL-D &           ACCEL &            RPLR \\
\midrule
$X_{\mathrm{train}}$ & \textbf{1.00 $\pm$ 0.00} & \textbf{0.98 $\pm$ 0.03} & \textbf{1.00 $\pm$ 0.00} & \textbf{0.99 $\pm$ 0.00} & 0.48 $\pm$ 0.27 & 0.66 $\pm$ 0.15 & 0.34 $\pm$ 0.18 & 0.28 $\pm$ 0.09 \\
 $X_{\mathrm{test}}$ & \textbf{0.95 $\pm$ 0.01} & \textbf{0.89 $\pm$ 0.06} &          0.79 $\pm$ 0.03 &          0.77 $\pm$ 0.02 & 0.49 $\pm$ 0.26 & 0.57 $\pm$ 0.20 & 0.36 $\pm$ 0.19 & 0.29 $\pm$ 0.08 \\
              Edge C & \textbf{0.87 $\pm$ 0.06} & \textbf{0.80 $\pm$ 0.14} &          0.45 $\pm$ 0.12 &          0.40 $\pm$ 0.05 & 0.46 $\pm$ 0.25 & 0.41 $\pm$ 0.22 & 0.31 $\pm$ 0.19 & 0.19 $\pm$ 0.06 \\
            Hardcore & \textbf{0.46 $\pm$ 0.07} & \textbf{0.34 $\pm$ 0.10} &          0.10 $\pm$ 0.05 &          0.13 $\pm$ 0.05 & 0.16 $\pm$ 0.13 & 0.13 $\pm$ 0.14 & 0.06 $\pm$ 0.07 & 0.02 $\pm$ 0.03 \\
\bottomrule
\end{tabular}
\end{sc}
\end{small}
\end{center}
\vskip -0.1in
\end{table}

\begin{table}[H]
\caption{$\text{GenGap}$ and $\text{ShiftGap}$ at the end of training for each method tested. We report their mean and standard deviation across 5 training runs, each corresponding to a different random seed. Bolded methods are not significantly different from the method with lowest mean ($p<0.05$), unless all are, in which case none are bolded.}
\label{tb:minigrid_gen_metrics}
\vskip 0.15in
\begin{center}
\begin{small}
\begin{sc}
\tabcolsep=0.11cm
\begin{tabular}{lrrrrrrrr}
\toprule
           Metric &                  VAE-DRED &                   EL-DRED &                       PLR &            $\mathcal{U}$ &                        DR &         ACCEL-D &                     ACCEL &                      RPLR \\
\midrule
  $\text{GenGap}$ &           0.09 $\pm$ 0.01 &           0.14 $\pm$ 0.03 &           0.24 $\pm$ 0.03 &          0.25 $\pm$ 0.02 & \textbf{-0.00 $\pm$ 0.01} & 0.08 $\pm$ 0.05 & \textbf{-0.02 $\pm$ 0.02} & \textbf{-0.01 $\pm$ 0.01} \\
$\text{ShiftGap}$ & \textbf{0.00 $\pm$ 0.04} & \textbf{-0.08 $\pm$ 0.11} & \textbf{-0.07 $\pm$ 0.14} & n/a (0) &  \textbf{0.16 $\pm$ 0.20} & 0.25 $\pm$ 0.16 &           0.66 $\pm$ 0.17 &  \textbf{0.18 $\pm$ 0.45} \\
     $\text{JSD}$ &           \textbf{0.11 $\pm$ 0.01} &           0.16 $\pm$ 0.02 &           0.29 $\pm$ 0.05 & n/a (0) &           0.42 $\pm$ 0.00 & 0.29 $\pm$ 0.03 &           0.38 $\pm$ 0.06 &           0.46 $\pm$ 0.04 \\
\bottomrule
\end{tabular}
\end{sc}
\end{small}
\end{center}
\vskip -0.1in
\end{table}

\section{CMDP specification and generation}\label{app:cmdp_levelset}

In Minigrid \cite{gym_minigrid}, the agent receives as an observation a partial view of its surroundings (in our experiments it is set to two tiles to each side of the agent and four tiles in front) and a one-hot vector representing the agent's heading. The action space consists of 7 discrete actions but, in our setting, only the actions moving the agent forward and rotating it to the left or right have an effect. The episode starts with the agent at its start tile and facing its starting orientiation. The episode terminates successfully when the agent reaches the goal tile. It receives a reward between 0 and 1 based on the number of timesteps it took to get there. The episode will terminate without a reward if the agent steps on a lava tile, or when the maximum number of timesteps is reached.

\begin{figure}[!htb]
    \centering
            \includegraphics[width=.4\linewidth]{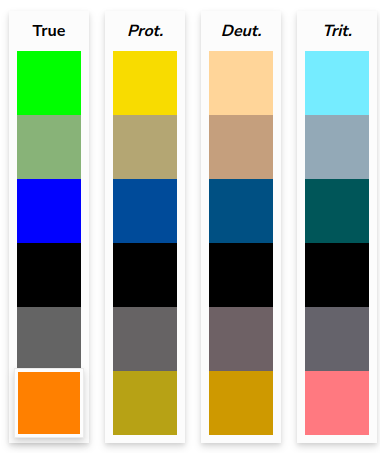}
    \caption{Color palette used for rendering minigrid layouts in this paper and their equivalent for Protanopia (Prot.), Deuteranopia (Deut.) and Tritanopia (Trit.) color blindness. We refer to each row in the main text as, in order: green (goal tiles), pale green (moss tiles), blue (agent), black (empty/floor tiles), grey (wall tiles) and orange (lava tiles).
    }
    \label{fig:color_palette}
\end{figure}

Levels are parameterised as 2D grids representing the overall layout, with each tile type represented by an unique ID. Tiles can be classified as navigable (for example, moss or empty tiles) or non-navigable (for example, walls and lava, as stepping into lava terminates the episode). To be valid, a level must possess exactly one goal and start tile, and to be solvable there must exist a navigable path between the start and the goal location. We provide the color palette of tiles used in \Cref{fig:color_palette}.

\begin{figure}[!htb]

\sbox\twosubbox{%
  \resizebox{\dimexpr.7\textwidth-1em}{!}{%
    \includegraphics[height=3cm]{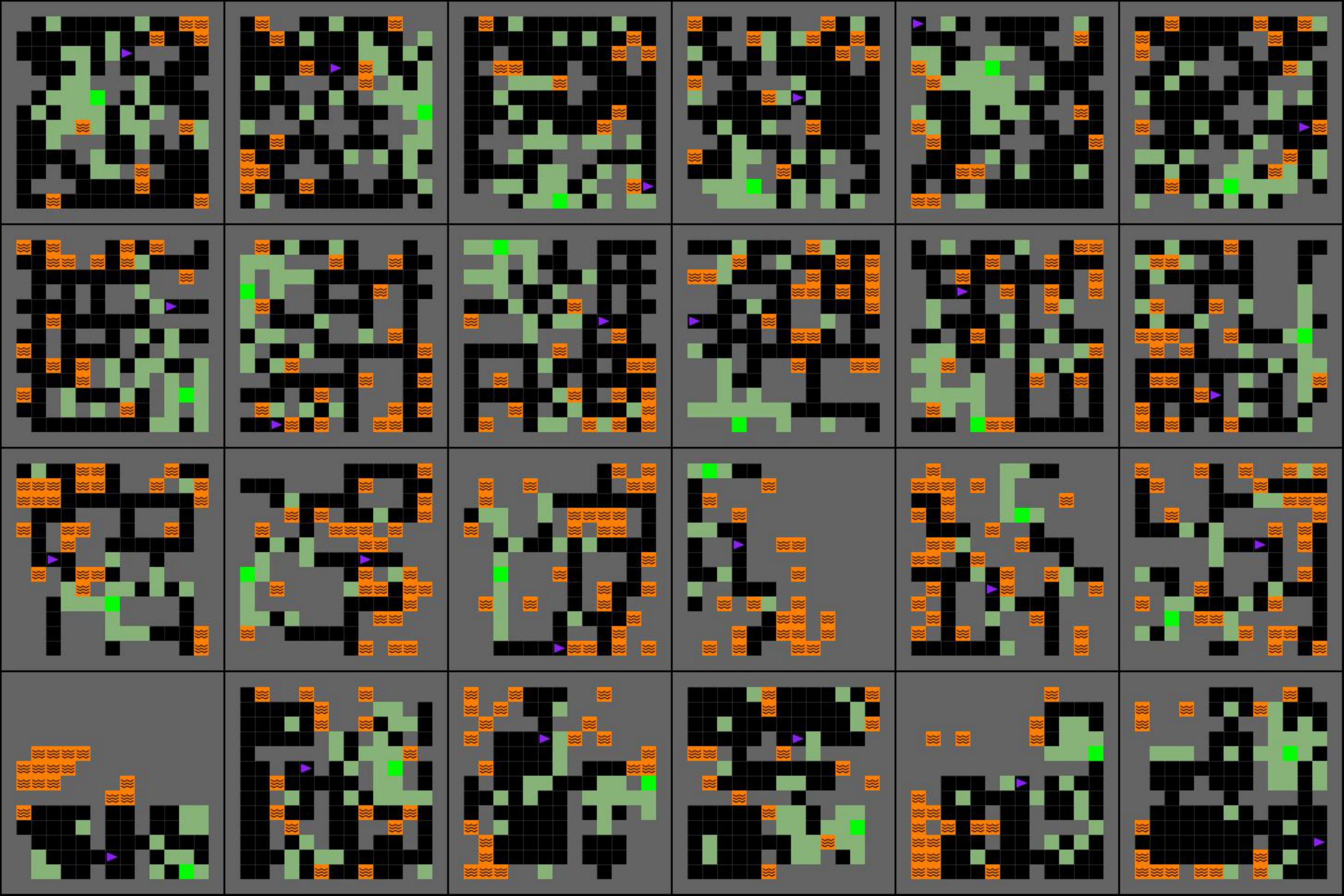}%
    \includegraphics[height=3cm]{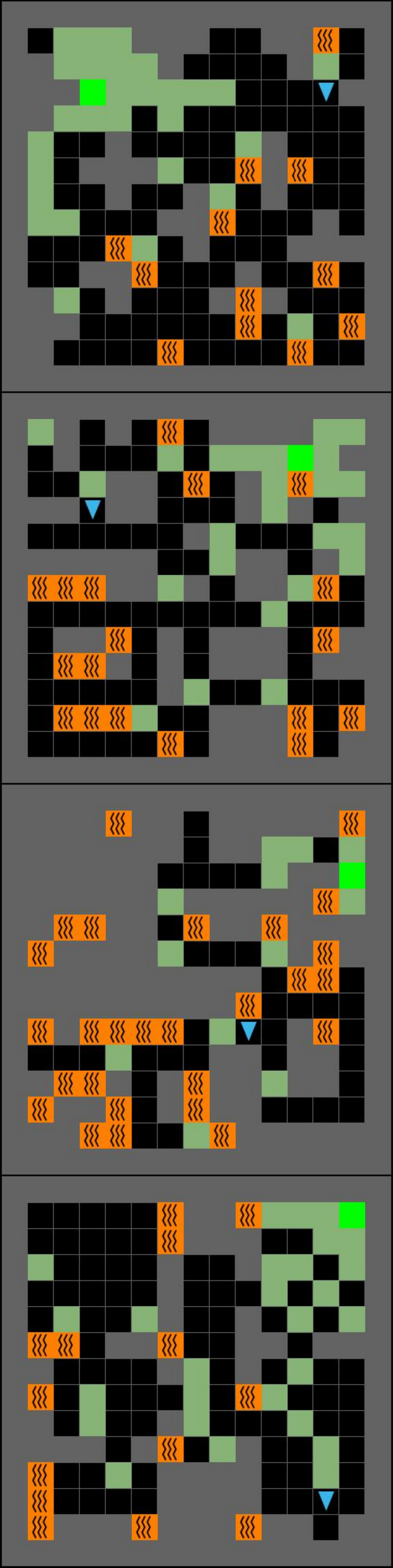}%
  }%
}
\setlength{\twosubht}{\ht\twosubbox}

\centering

\subcaptionbox{Example training levels from $X_\text{train}$\label{f}}{%
  \includegraphics[height=\twosubht]{app_content/layouts/cave_escape_dataset.pdf}%
}\quad
\subcaptionbox{Held-out levels\label{s1}}{%
  \includegraphics[height=\twosubht]{app_content/layouts/cave_escape_16k_4ts_id_rotated.pdf}%
}

\caption{Sample levels from $X_\text{train}$ and from held-out test set $X_\text{test}$. Wall tiles are rendered in gray, empty tiles in black, moss tiles in green and the goal tile in lime green. The agent is rendered as a blue or purple triangle, and is depicted at its start location. Each row corresponds to levels generated with a specific wave function collapse base pattern. Four different base patterns are used to generate $X_\text{train}$ and $X_\text{test}$ layouts. As in Procgen games, we find that the PPO agent exhibits a significant $\text{GenGap}$ on held-out levels when being restricted to train on 200-500 levels. We therefore have $X_\text{train}$ contain 512 levels and we have $X_\text{test}$ contain 2048 levels for more accurate agent evaluation.}\label{fig:layouts_dataset}

\end{figure}

We provide example levels of the CMDP in \Cref{fig:layouts_dataset}. The CMDP level space corresponds to the subset of solvable levels in which moss and lava node placement is respectively positively and negatively correlated with their shortest path distance to the goal.\footnote{To measure the shortest path distance to goal of a non-navigable node, we first find the navigable node that is closest from it and measure its shortest path distance to the goal. We then add to it the number of tile separating this navigable node to the non-navigable node of interest. If there are multiple equally close navigable nodes, we select the navigable node with the smallest shortest path distance to the goal.} Under partial observability, the optimal policy for this CMDP would leverage moss and lava locations as contextual cues, seeking regions with high moss density and avoiding regions with high lava density. 

\begin{figure}[!htb]

\sbox\twosubbox{%
  \resizebox{\dimexpr.9\textwidth-1em}{!}{%
    \includegraphics[height=3cm]{app_content/layouts/cave_escape_dataset.pdf}%
    \includegraphics[height=3cm]{app_content/layouts/cave_escape_16k_4ts_id_rotated.pdf}%
  }%
}
\setlength{\twosubht}{\ht\twosubbox}

\centering

\subcaptionbox{Example levels from the first set of edge cases}{%
  \includegraphics[height=\twosubht]{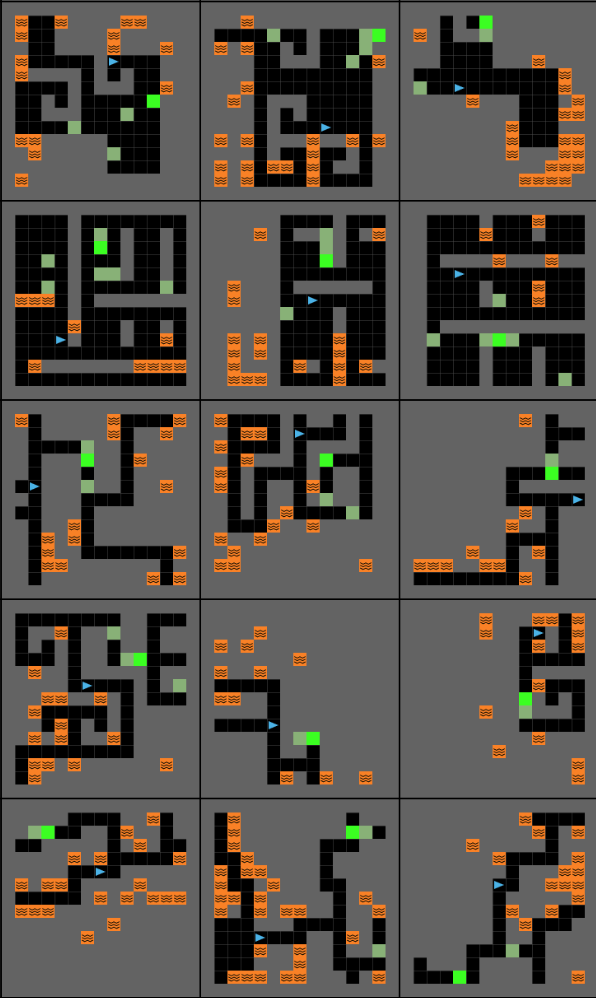}%
}\quad
\subcaptionbox{Example levels from the second set of edge cases\label{s2}}{%
  \includegraphics[height=\twosubht]{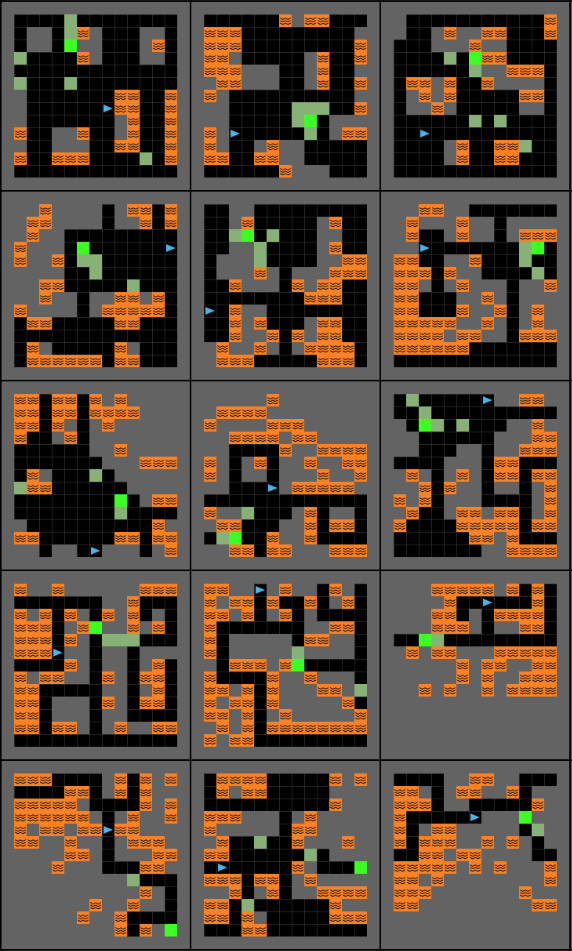}%
}

\caption{We generate 2 separate sets of edge cases, using 14 different base patterns not used to generate $X_\text{train}$. The moss density of the first set (a) is 2 to 5 times as small as levels found within $X_\text{train}$, making finding the goal using CMDP contextual cues more challenging. (b) is the same as (a) but with a lava tile density 2 to 5 times higher than $X_\text{train}$ levels, which makes avoiding lava more challenging. Both sets contain the same number of levels (224), and they are combined when evaluating the agent on edge cases.}\label{fig:layout_edge_cases}

\end{figure}

\subsection{Generating highly structured levels}

We use the wave function collapse (WFC) procedural generation algorithm \citep{WFC} to obtain highly structured but still diverse gridworld layouts. WFC gradually collapses a superposition of all possible level parameters into a layout respecting the constraints defined by an input pattern. By doing so, it is possible to generate a vast number of tasks from a small number of starting patterns. Given a suitable base pattern, WFC provides a high degree of structure and complexity in generated layouts, and it guarantees that both task structure and diversity scale with the gridworld dimensions. Our implementation supports generating layouts from 22 %
different base patterns, and supports the specification of custom user-defined patterns. 

After generating a layout using WFC, we convert the navigable nodes of a layout into a graph, choose its largest connected component as the layout and convert any unreachable nodes to non-navigable nodes. We place the goal location at random and place the start at a node located at the median geodesic distance from the goal in the navigation graph. By doing so we ensure that the complexity of generated layouts is relatively consistent given a specific grid size and base pattern. Finally, we sample tiles according to parameterisable distributions defined over the navigable and non-navigable node sets. 

In our experiments, the tile set consists of the $\{$ moss, empty, start, goal $\}$ tiles as the navigable set and the $\{$ wall, lava$\}$ tiles as the non-navigable set. We parameterise tile distributions such that moss tiles are more likely to be sampled on navigable nodes close to the goal, while lava tiles are more likely on non-navigable nodes far away from the goal.

\subsection{Controlling level complexity}

We provide two options to vary the complexity of the level distribution. The first is to change the layout size. Due to partial observability this significantly increases their complexity, and makes levels from the ``Hardcore'' set depicted in \Cref{fig:zsg_hard} challenging to solve for Human players. The second option is to change the sampling probability of moss and lava tiles. Reducing the fraction of moss to navigable tiles diminishes their usefulness as context cues. On the other hand, increasing the density of lava tiles increases the risk associated with selecting the wrong action during play. In our experiments we assess the agent's performance on edge-cases by defining level sets with a larger layout size (\Cref{fig:zsg_hard}), or with different moss and lava tile distributions (\Cref{fig:layout_edge_cases}).  

\section{Implementation details}\label{app:imp_details}

\subsection{Procgen}\label{app:procgen}
The Procgen Benchmark is a set of 16 diverse PCG environments that echoes the gameplay variety seen in the ALE benchmark \cite{ALE}. The game levels, determined by a random seed, can differ in visual design, navigational structure, and the starting locations of entities. All Procgen environments use a common discrete 15-dimensional action space and generate $64 \times 64 \times 3$ RGB observations. A detailed description of each of the 16 environments is provided in \citet{procgen}. RL algorithms such as PPO reveal significant differences between test and training performance in all games, making Procgen a valuable tool for evaluating generalisation performance. 

We conduct our experiment on the easy setting of Procgen, which employs 200 training levels and a budget of 25M training steps, and evaluate the agent's ZSG performance on the full range of levels, excluding the training levels.

We employ the same ResNet policy architecture and PPO hyperparameters (identical for all games) as \cite{procgen}, which we reference in \Cref{table:hyperparams}. To compute $\mut{L}{b}$ and the scoring strategy for $(S^\text{MI})$ in our experiments, we model $p_\vtheta(\lvl|b(o_t))$ as a linear classifier predicting the level identity from the output of the last shared network layer between the actor and critic. $p_\vtheta$ is trained using trajectory rollouts collected by sampling uniformly from $L$. We ensure the training processes of the agent and the classifier remain independent from one-another by employing a separate optimiser, and by stopping the gradients from propagating through the agent's network.

\subsection{Minigrid RL agent} \label{app:impl_minigrid}

We use the recurrent PPO agent and hyperparameters employed in \cite{ACCEL} for all our experiments. The actor and critic share their initial layers. The first initial layer consists of a convolutional layer with 16 output channels and kernel size 3 processes the agent's view and a fully connected layer that processes its directional information. Their output is concatenated and fed to an LSTM layer with hidden size 256. The actor and critic heads each consist of two fully connected layers of size 32, the actor outputs a categorical distribution over action probabilities while the critic outputs a scalar. Weights are optimised using Adam and we employ the same hyperparameters in all experiments, reported in \Cref{table:hyperparams}. Trajectories are collected via 36 worker threads, with each experiment conducted using a single GPU and 10 CPUs.

Following \cite{ACCEL}, DR and RPLR use domain randomisation as their standard level generation process, in which the start and goal locations, alongside a random number between 0 and 60 moss, wall or lava tiles are randomly placed. The level editing process of ACCEL and EL-DRED remains unchanged from \cite{ACCEL}, consisting of five steps. The first three steps may change a randomly selected tile to any of its counterparts, whereas the last two are reserved to replacing the start and goal locations if they had been removed in prior steps.

We train three different seeds for each baseline. We use the hyperparameters reported in \cite{ACCEL} for the DR, RPLR and ACCEL methods and the hyperparameters reported in \cite{PLR} for PLR, as an extensive hyperparameter search was conducted in a similarly sized Minigrid environment for each method. VAE-DRED employs the same hyperparameters as PLR for its level buffer, with some additional secondary sampling strategy hyperparameters introduced by VAE-DRED. We did not perform an hyperparameter search for VAE-DRED as we found that the initial values worked adequately. We report all hyperparameters in \Cref{table:hyperparams}.

\subsection{VAE architecture and pre-training procedure}\label{app:vae_imp}

We employ the $\beta-$VAE formulation proposed in \cite{betaVAE}, and we parametrise the encoder as a Graph Convolutional Network (GCN), a generalisation of the Convolutional Neural Network (CNN) \citep{CNNimagenet} to non Euclidian spaces. Our choice of a GCN architecture is not motivated by a desire of maximising the VAE's performance (in fact we expect a CNN or MLP encoder to work just as well in Minigrid). It is instead intended as a proof of concept for an architecture that could transfer to more complex simulators. Since the level parameter space $\sX$ is simulator-specific, employing a graph as an input modality for our encoder makes our model architecture easier to transfer to different simulators and domains. Using a GCN, some of the inductive biases that would be internal in a traditional architecture can be defined through an external wrapper that encodes the environment parameter $\vx$ into the graph $\mathcal{G}_\vx$. 

In Minigrid, we represent the gridworld levels as a grid graphs, each cell being an individual node. This effectively makes the GCN equivalent to a traditional CNN in this scenario. However a GCN provides ways of providing additional domain specific biases that a CNN lacks. For example, a domain expert with the notion that goal, lava and moss tiles are somehow correlated can add edges between these tiles in the graph.

We select the GIN architecture \citep{GIN} for the GCN, which we connect to an MLP network that outputs latent distribution parameters $(\vmu_\vz, \vsigma_\vz)$. The decoder is a fully connected network with three heads. The \textit{layout} head outputs the parameters of categorical distributions for each grid cell, predicting the tile identity between [Empty, Moss, Lava, Wall]. The \textit{start} and \textit{goal} heads output the parameters of categorical distributions predicting the identity of the start and goal locations across grid cells, ensuring a single goal or start node get sampled in any given level.

We pre-train the VAE for 200 epochs on $X_\text{train}$, using cross-validation for hyperparameter tuning, each run taking about 25 minutes on a GPU-equipped laptop. During training, we formulate the reconstruction loss as a weighted sum of the cross-entropy loss for each head.\footnote{To compute the cross-entropy loss of the layout head, we replace the start and goal nodes in the reconstruction targets by a uniform distribution across \{moss, empty\}.}%
At deployment, we guarantee \textit{valid} layouts  get generated (i.e. layouts containing a unique start and goal location, but not necessarily solvable) by generating level parameters sequentially. We first sample the layout, then we sample the start location, masking any non-navigable tile generated in the last step. We then add the start tile to our mask before sampling the goal location. In this way, we guarantee valid start and goal locations that will not override one-another. Note that our generative model may still generate \textit{unsolvable} layouts, which do not have a passable path between start and goal locations, and therefore it must have learnt to generate solvable layouts in order to be useful. 

\begin{figure}[!htb]

\sbox\twosubbox{%
  \resizebox{\dimexpr.9\textwidth-1em}{!}{%
    \includegraphics[height=3cm]{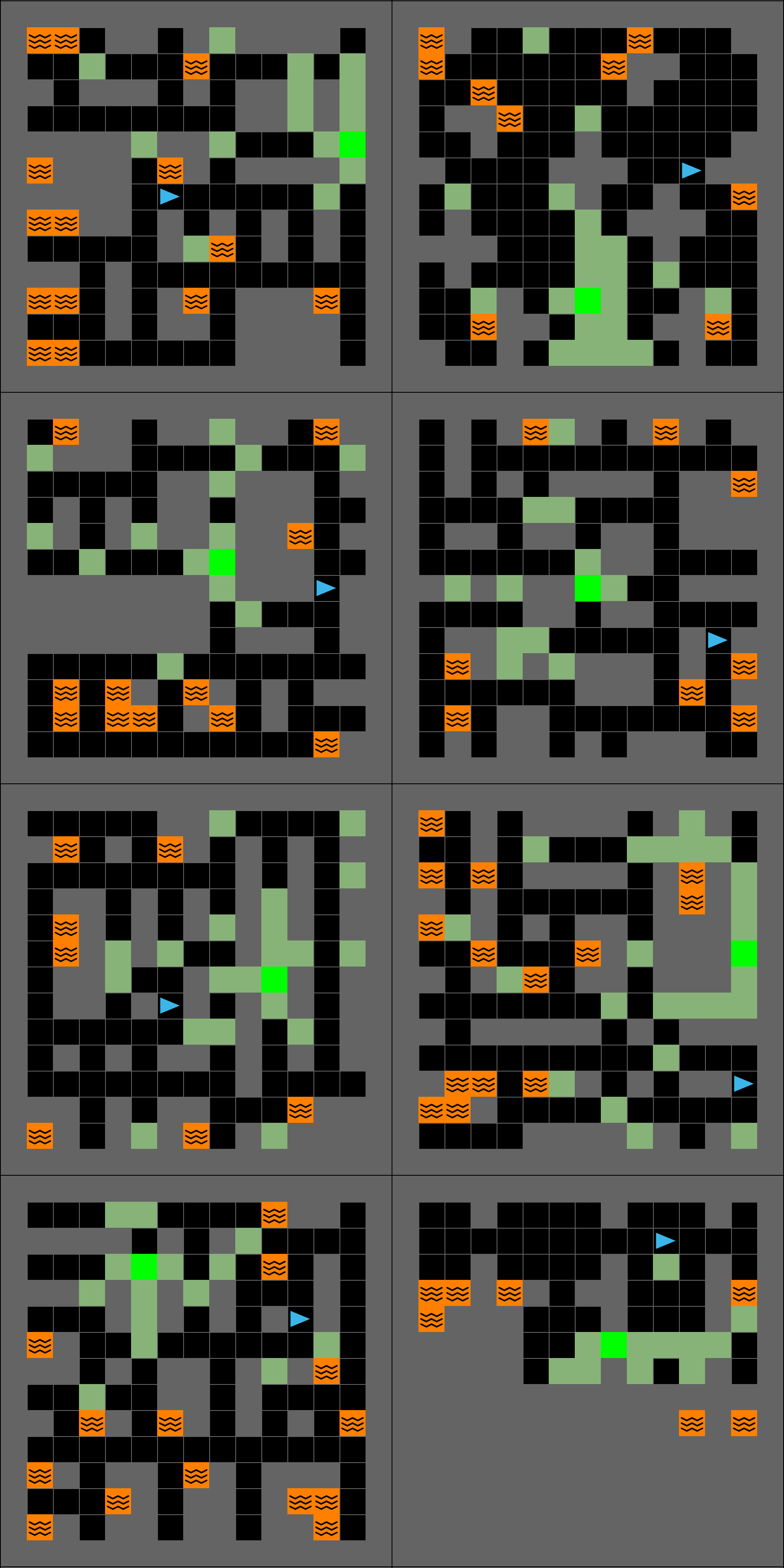}%
    \includegraphics[height=3cm]{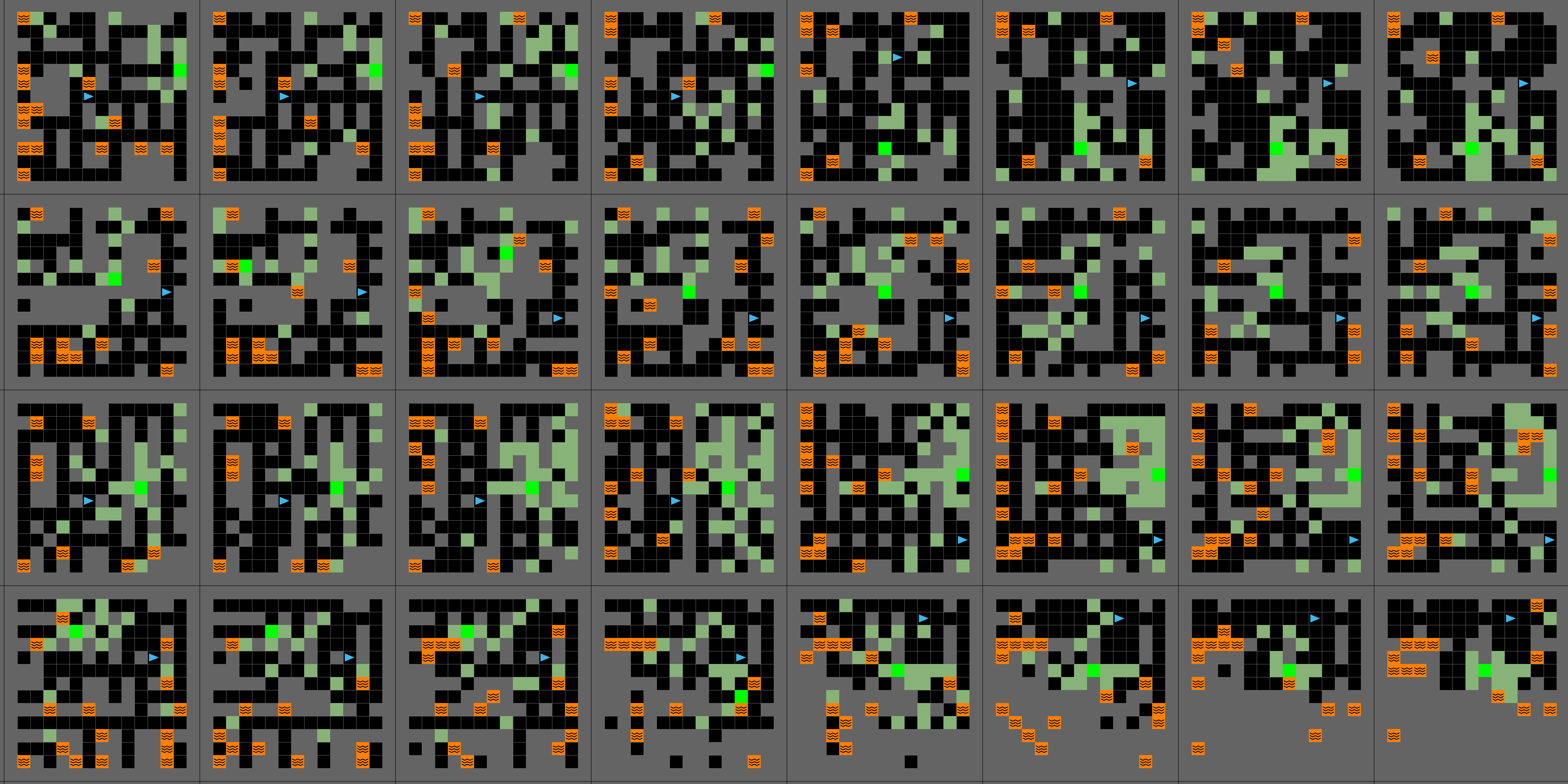}%
  }%
}
\setlength{\twosubht}{\ht\twosubbox}

\centering

\subcaptionbox{Parent levels sampled from $X_\text{train}$}{%
  \includegraphics[height=\twosubht]{app_content/layouts/base_levels_crop.png}%
}\quad
\subcaptionbox{Levels generated by the VAE}{%
  \includegraphics[height=\twosubht]{app_content/layouts/interp_levels_crop.png}%
}

\caption{Sample of levels generated by interpolating within the latent space of the VAE. On the right, each row represents an interpolation between the latent embeddings of a pair of base levels (on the left). The layouts generated are not always solvable, but, when they are solvable, they tend to retain semantics consistent with the CMDP.}\label{fig:interpolated_levels}

\end{figure}

We do not explicitly encourage the VAE to generate solvable levels, but we find that models with high ELBO (\Cref{eq:ELBO}) on the validation set tend to also have a high generated layout solvability rate. Layouts reconstructed from $X_\text{train}$ have over 80\% solvability rate, while layouts generated via latent space interpolations have over 70\% solvability rate. In practice, maximising the ELBO results in generated layouts sharing contextual semantics with $X_\text{train}$ levels. This can be observed qualitatively in the VAE-generated layouts included in \Cref{fig:interpolated_levels}, and quantitatively in \Cref{fig:gen_gaps,fig:buffermetrics}, where we report that contextually important semantics are transferred to the agent's training distribution.

To tune the VAE we conduct a random sweep over architectural parameters (number of layers, layer sizes), the $\beta$ coefficient, individual decoder head reconstruction coefficients and the learning rate over a total budget of 100 runs. We select the configuration achieving the highest ELBO on the validation set. These hyperparameters are reported in %
\Cref{table:hyperparams_vae}.

\begin{table}[!htb]
\caption{Hyperparameters used for Minigrid experiments. Hyperparameters shared between methods are only reported if they change from the method above.}
\label{table:hyperparams}
\begin{center}
\scalebox{0.88}{
\begin{tabular}{lcccc}
\toprule
\textbf{Parameter} & Procgen & MiniGrid\\
\midrule
\emph{PPO} & \\
$\gamma$ &0.999 & 0.995 \\
$\lambda_{\text{GAE}}$ & 0.95 & 0.95  \\
PPO rollout length & 256 & 256 \\
PPO epochs & 3 & 5 \\
PPO minibatches per epoch & 8 & 1 \\
PPO clip range & 0.2 & 0.2 \\
PPO number of workers & 64 & 32 \\
Adam learning rate & 5e-6 & 1e-4 \\
Adam $\epsilon$ & 1e-5 & 1e-5  \\
PPO max gradient norm & 0.5  & 0.5  \\
PPO value clipping & yes  & yes \\
return normalisation & yes & no \\
value loss coefficient & 0.5 & 0.5 \\

\addlinespace[10pt]
\emph{PLR} & & & \\
Scoring function & $\ell_1$-value loss  & $\ell_1$-value loss \\
Replay rate, $p$ & 1.0 & 1.0 \\
Buffer size, $K$ & 200 & 512 \\
Prioritisation, & rank & rank \\
Temperature, & 0.1  & 0.1 \\
Staleness coefficient, $\rho$ & 0.1  & 0.3 \\

\addlinespace[10pt]
\emph{RPLR} & & & \\
Scoring function, & & positive value loss \\
Replay rate, $p$ &  & 0.5 \\
Buffer size, $K$ &  & 4000 \\

\addlinespace[10pt]
\emph{ACCEL} & & \\
Edit rate, $q$ & & 1.0 \\
Replay rate, $p$ & & 0.8 \\
Buffer size, $K$ & & 4000 \\
Edit method, & & random  \\
Levels edited, & & easy \\

\addlinespace[10pt]
\emph{VAE-DRED} & & & \\
Replay rate, $p$ & & 1.0 \\
Scoring function support, & & dataset \\
Staleness support, &  & dataset \\
Secondary Scoring function, & & $\ell_1$-value loss \\
Secondary Scoring function support, & & buffer \\
Secondary Temperature, & & 1.0 \\
Mixing coefficient, $\eta$ & & linearly increased from 0 to 1 \\

\bottomrule 
\end{tabular}
}
\end{center}
\end{table}

\begin{table}[!htb]
\caption{Hyperparameters used for pre-training the VAE.}
\label{table:hyperparams_vae}
\begin{center}
\scalebox{0.88}{
\begin{tabular}{lccc}
\toprule
\textbf{Parameter} & \\
\midrule
\emph{VAE} & \\
$\beta$ & 0.0448 \\
layout head reconstruction coefficient & 0.04 \\
start and goal heads reconstruction coefficients & 0.013 \\
number of variational samples & 1 \\
Adam learning rate & 4e-4 \\
Latent space dimension & 1024 \\
number of encoder GCN layers & 4 \\
encoder GCN layer dimension & 12 \\
number of encoder MLP layers (including bottleneck layer) & 2 \\
encoder MLP layers dimension & 2048 \\
encoder bottleneck layer dimension & 256 \\
number of decoder layers & 3 \\
decoder layers dimension & 256 \\

\bottomrule 
\end{tabular}
}
\end{center}
\end{table}

